%% file: acl_latex.tex
\documentclass[11pt]{article}

\usepackage[final]{acl}

\usepackage{times}
\usepackage{latexsym}
\usepackage[most]{tcolorbox}
\usepackage{tabularx}
\usepackage{xcolor}
\usepackage{enumitem}
\usepackage{tcolorbox}
\usepackage[most]{tcolorbox}
\usepackage{graphicx}
\usepackage{booktabs}
\usepackage{arydshln}
\definecolor{lightgray}{gray}{0.9}
\definecolor{stepgray}{gray}{0.8}
\definecolor{PineGreen}{HTML}{007C4A} % HTML格式定义

\makeatletter
\renewcommand\tableofcontents{%
  \@starttoc{toc}%
}

\let\origaddcontentsline\addcontentsline
\newcommand{\DisableTOC}{\let\addcontentsline\@gobblethree}
\newcommand{\EnableTOC}{\let\addcontentsline\origaddcontentsline}

\newcommand\blfootnote[1]{%
  \begingroup
  \renewcommand\thefootnote{}%
  \footnotetext{#1}%
  \addtocounter{footnote}{-1}%
  \endgroup
}

\makeatother

\usepackage[T1]{fontenc}
\usepackage[utf8]{inputenc}

\usepackage{microtype}

\usepackage{inconsolata}

\usepackage{graphicx}

\title{ShotFinder: Imagination-Driven Open-Domain Video Shot Retrieval via Web Search}

\author{
\normalfont\mdseries
Tao Yu$^{1,2 *\spadesuit}$,
Haopeng Jin$^{1*\dagger}$,
Hao Wang$^{1*\dagger}$,
Shenghua Chai$^{1\dagger}$,
Yujia Yang$^{2}$,
Junhao Gong$^{4}$
\\
\normalfont\mdseries
Jiaming Guo$^{1\dagger}$,
Minghui Zhang$^{1\dagger}$,
Xinlong Chen$^{1,2}$,
Zhenghao Zhang$^{2}$,
Yuxuan Zhou$^{5}$,
Yufei Xiong$^{1\dagger}$
\\
\normalfont\mdseries
Shanbin Zhang$^{1,2}$,
Jiabing Yang$^{1,2}$,
YiFan Zhang$^{1,2\ddagger}$,
Hongzhu Yi$^{2\ddagger}$,
Xinming Wang$^{1,2}$
\\
\normalfont\mdseries
Cheng Zhong$^{3}$,
Xiao Ma$^{3}$,
Zhang Zhang$^{1,2}$,
Yan Huang$^{1,2\ddagger}$,
Liang Wang$^{1,2}$
\\
$^{1}$CASIA
\quad
$^{2}$UCAS
\quad
$^{3}$Lenovo
\quad
$^{4}$Peking University
\quad
$^{5}$Tsinghua University
\\
yutao2025@ia.ac.cn
\\
\url{https://github.com/yutao1024/ShotFinder}
}

\begin{document}

\maketitle

\blfootnote{%
\parbox[t]{\columnwidth}{%
\footnotesize
$^{*}$ Equal contribution.\\
$^{\dagger}$ Work done during an internship at CASIA.\\
$^{\ddagger}$ Corresponding author.\\
$^{\spadesuit}$ Project leader.
}}

\DisableTOC

\begin{abstract}

In recent years, large language models (LLMs) have made rapid progress in information retrieval, yet existing research has mainly focused on text or static multimodal settings. Open-domain video shot retrieval, which involves richer temporal structure and more complex semantics, still lacks systematic benchmarks and analysis. To fill this gap, we introduce ShotFinder, a benchmark that formalizes editing requirements as keyframe-oriented shot descriptions and introduces five types of controllable single-factor constraints: Temporal order, Color, Visual style, Audio, and Resolution. We curate 1,210 high-quality samples from YouTube across 20 thematic categories, using large models for generation with human verification. Based on the benchmark, we propose ShotFinder, a text-driven three-stage retrieval and localization pipeline: (1) query expansion via video imagination, (2) candidate video retrieval with a search engine, and (3) description-guided shot localization. Experiments on multiple closed-source and open-source models reveal a significant gap to human performance, with clear imbalance across constraints: temporal localization is relatively tractable, while color and visual style remain major challenges. These results reveal that open-domain video shot retrieval is still a critical capability that multimodal large models have yet to overcome.

\end{abstract}

\section{Introduction}

% 近年来，大模型技术的快速发展显著提升了模型在信息检索任务中的能力。文本大模型已能够通过推理与工具调用的方式与搜索引擎交互，从而完成复杂的文本信息检索任务，例如 Search-R1 和 WebDancer 等工作。与此同时，多模态大模型也开始展现出通过推理与工具调用进行跨模态信息检索的潜力，如 WebWatcher 等方法。然而，这类研究主要集中于文本或静态多模态信息检索，对于更具时序结构和语义复杂性的视频检索问题仍缺乏系统性的探索。

% 在视频剪辑实践中，视频被划分为多个基本单位——镜头。镜头通常指的是视频中的一段连续画面，直到发生切换或过渡，这些切换通常伴随着场景或内容的显著变化。镜头是视频的基本组成部分，它不仅构成了视频的时间轴，还在视频内容表达上起到至关重要的作用。剪辑师通常需要大量的视频浏览经验，才能高效地从海量素材中检索到所需的镜头。当剪辑师希望获取某一具体镜头时，通常需要事先观看并记住视频内容及其关键位置，这使得视频剪辑在很大程度上依赖个人经验积累，增加了从业者的入门门槛。

% 在此背景下，一个自然的问题是：现有大模型是否具备通过推理与工具调用，与搜索引擎交互以完成视频镜头级检索与定位的能力？ 回答这一问题不仅具有重要的学术意义，也直接关系到真实视频剪辑场景中的应用潜力。

% 为此，我们提出一个面向真实剪辑检索场景的视频镜头检索 benchmark —— ShotFinder。该 benchmark 将剪辑师在视频素材检索过程中的核心需求，系统刻画为镜头描述和时序、色彩、风格、声音及分辨率五种单一因素约束，在保持任务可控性的同时，系统分析不同描述因素对视频检索与定位的影响。为降低视频内容分布带来的偏置，我们采用约束感知的主题分配策略，并通过大模型生成结合人工校验的方式构建了一个包含 1200 个视频镜头的高质量数据集。评测阶段采用基于关键画面的大模型辅助评估，以更有效刻画复杂描述条件下的检索性能差异。

% 在此 benchmark 上，我们进一步提出了一种面向开放域视频的文本描述驱动镜头检索与定位方法。给定镜头描述及可选的附加约束，该方法采用分阶段处理流程：首先通过基于视频联想的查询扩展，将镜头级描述提升为视频级搜索语义，以提高检索召回率；随后从网络视频平台获取候选视频集合；最后在候选视频中基于描述执行时间定位，识别与输入语义一致的目标镜头。该方法将镜头级理解、开放域视频检索与描述引导的精细定位有机结合，为利用大模型进行视频镜头级检索提供了一种可行且有效的解决方案。

% 基于 ShotFinder 的实验结果，我们对人类表现以及多种闭源与开源模型进行了全面评估，主要结论如下：

% 人类评估显著优于所有模型。 人类在各类别上均取得最高性能，平均准确率达到 88.5，远高于所有模型，表明 ShotFinder 对现有模型仍具有较高挑战性。

% 闭源模型整体性能更优，但与人类水平仍存在显著差距。 在闭源模型中，GPT-5.2 表现最佳，平均准确率为 26.9，在时间定位和风格理解方面具有明显优势，但在颜色相关任务上的表现仍然有限。

% 开源模型在部分子任务上表现出一定竞争力。 Qwen3 系列在开源模型中表现突出，尤其在时间定位和颜色理解任务上取得较好结果，但在风格理解方面仍落后于闭源模型。

% 不同任务类别之间存在明显的性能不均衡。 时间定位是整体表现最好的任务类别，而颜色与风格理解仍然是当前模型面临的主要挑战。

% 模型规模并不能保证在所有任务上的一致性能提升。更大的模型规模并不一定带来全面的性能提升。例如，尽管模型规模较小，Qwen3-Omni-30B-A3B 在时间定位任务上的表现仍优于 Gemini-3-Pro。这表明模型架构设计与多模态对齐策略在性能提升中起着超越模型规模本身的重要作用。

In recent years, the rapid advances in LLMs \citep{yang2025qwen3technicalreport, yu2025aligningmultimodalllmhuman, zhang2025mmrlhfstepforwardmultimodal,zhou2025does,chou2026improving, openai2024gpt4, google2023gemini, anthropic2024claude3} have substantially improved their capability in information retrieval. Text-based LLMs can already interact with search engines via reasoning and tool use to accomplish complex text retrieval tasks \citep{jin2025searchr1trainingllmsreason, wu2025webdancer, wu2025webwalker, tao2025webshaper, yu2025browseragentbuildingwebagents}. Meanwhile, multimodal large language models (MLLMs) have begun to show the potential to perform cross-modal retrieval through reasoning and tool invocation, exemplified by methods \citep{geng2025webwatcher,yu2026omnideepsearchbenchmarkaudiodrivenomnimodal}. However, existing research has largely focused on text or static multimodal retrieval, while systematic investigation of video retrieval remains limited, despite its richer temporal structure and more complex semantics.

In practical video editing workflows, videos are typically divided into multiple basic units—shots \citep{10.1145/3591106.3592247}. A shot generally refers to a continuous sequence of frames between one cut or transition and the next. As the fundamental building blocks of a video, shots not only constitute the temporal axis of the video but also serve as important carriers of content expression \citep{lu2025skaldlearningbasedshotassembly}. When video editors attempt to obtain a specific shot, they often need to preview the footage in advance and remember the timestamps, which makes video editing largely dependent on personal experience and raises the entry barrier for newcomers.

Against this backdrop, a natural question arises: can current MLLM interact with search engines through reasoning and tool use to accomplish shot-level video retrieval and temporal localization? Answering this question is not only of academic importance but also directly relates to the practical potential of real-world video editing applications.

\begin{figure*}[t]
 
\centering
\includegraphics[width=0.9\linewidth]{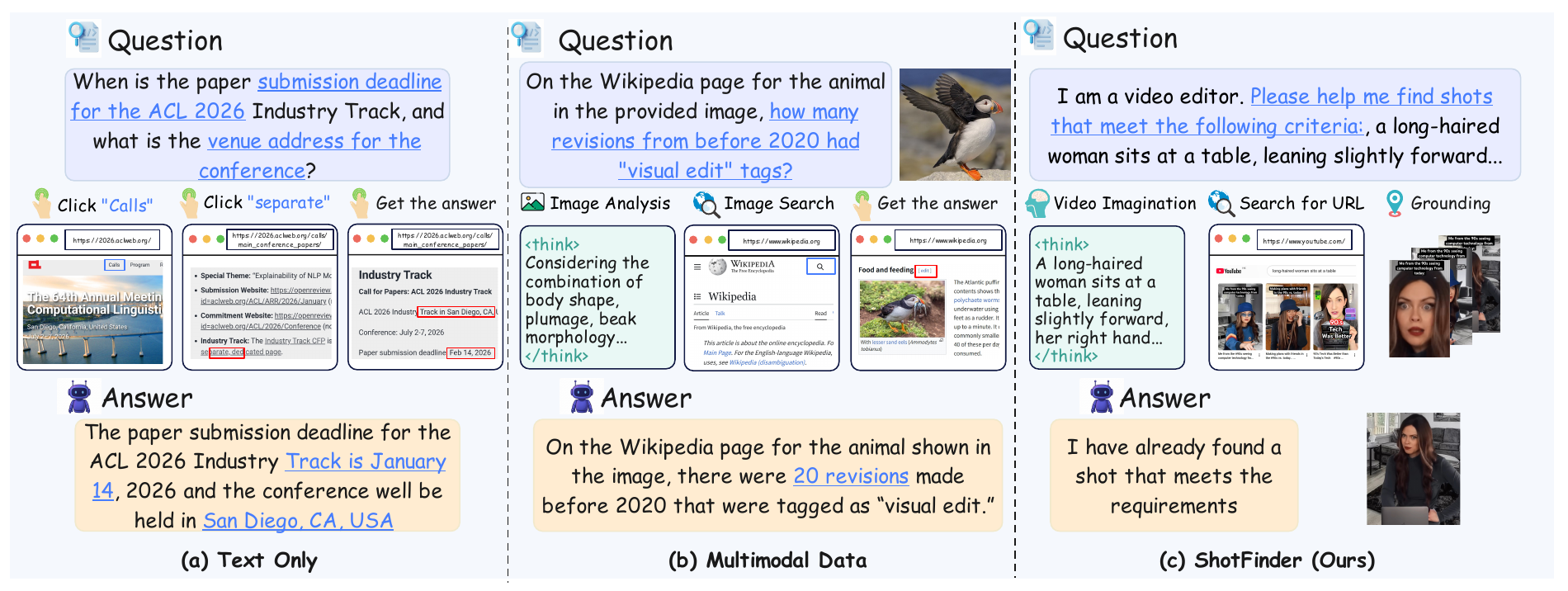}        %这个是在LaTeX文件夹中的相对路径
\caption{Comparison of ShotFinder benchmark with (a) text benchmarks and (b) other multimodal benchmarks. ShotFinder targets open-domain video retrieval. It requires the model to perform "Video Imagination" to bridge the gap between shot descriptions and full video search, followed by searching for URLs and grounding shots.}
\label{1}
\vspace{-10pt}
\end{figure*}

To this end, we propose ShotFinder, a shot retrieval benchmark tailored to realistic video editing workflows. ShotFinder systematically formulates the core requirements of video editors in footage retrieval as shot descriptions and five single-factor constraints in terms of temporal, color, style, audio, and resolution \citep{pardo2021learningcutwatchingmovies, bonneel2013example, tian2018audiovisualeventlocalizationunconstrained, Tu_2021}. While keeping the task controllable, ShotFinder enables analysis of how different descriptive factors affect video retrieval and temporal localization. To mitigate biases introduced by content distribution, we adopt a constraint-aware topic assignment strategy and construct a dataset containing 1,210 high-quality video shots through LLM-based generation with human verification. For evaluation, we employ a keyframe-based LLM-assisted assessment method, which more effectively captures performance differences among methods under complex conditions. The difference between ShotFinder and previous benchmarks is shown in Figure \ref{1}.

Building upon the benchmark, we further propose ShotFinder, a text-description-driven method for open-domain video shot retrieval and localization. Given a shot description and optional additional constraints, the method follows a staged processing pipeline. First, it performs query expansion based on video imagination, lifting shot-level descriptions to video-level search semantics. It then interacts with search engines to collect a set of candidate videos from online video platforms. Finally, it conducts description-guided shot localization to identify the target shot that matches the input semantics. By organically integrating shot-level understanding, open-domain video retrieval, and description-driven fine-grained localization.

% Based on experiments on ShotFinder benchmark, we conduct a comprehensive evaluation of human performance as well as a variety of closed-source and open-source models. The key findings are summarized as follows: (1) \textbf{Closed-source models perform better overall but still fall far short of human-level performance.} Among the closed-source models, GPT-5.2 achieves the highest average accuracy, showing clear advantages in \textbf{Temporal} and \textbf{Style}, while still being limited in \textbf{Color}. (2) \textbf{Open-source models are competitive on certain subtasks.} The Qwen3 series achieve strong results, especially in \textbf{Temporal} and \textbf{Color}, but still lags behind closed-source models in \textbf{Style}. (3) \textbf{There is a clear imbalance across task categories.} \textbf{Temporal} is the best-performing category overall, whereas \textbf{Color} and \textbf{Style} remain the major challenges for current models. (4) \textbf{Model scale does not guarantee consistent gains across all tasks.} Qwen3-Omni-30B-A3B performs close to GPT-5.2 on \textbf{Temporal}, suggesting that architectural design and multimodal alignment strategies play a critical role beyond model scale alone.

On the ShotFinder benchmark, we evaluate human performance and multiple closed-source and open-source models. The results show that all models remain far below human performance. Closed-source models perform best overall, with GPT-5.2 achieving the highest average accuracy, while Qwen3 models show competitive results on \textbf{Temporal} and \textbf{Color}. Across tasks, \textbf{Temporal} is relatively easier, whereas \textbf{Color} and \textbf{Style} remain challenging. Moreover, model scale does not always yield consistent gains, suggesting the importance of architecture and multimodal alignment~\citep{yu2026scoperoutercostawareopensetvlm,wang2026beyond}.

In summary, the main contributions are as follows: we introduce ShotFinder, the first benchmark for open-domain video shot retrieval, and propose an imagination-driven retrieval method. Experiments on the ShotFinder benchmark reveal the key challenges faced by current models in this field and point to directions for improvement.

% \begin{itemize}
    % \item \textbf{Human evaluation significantly outperforms all models.} Humans achieve the best performance across all categories, far exceeding all models, indicating that ShotFinder remains highly challenging for current models.
    
%  \textbf{Closed-source models perform better overall but still fall far short of human-level performance.} Among the closed-source models, GPT-5.2 achieves the highest average accuracy, showing clear advantages in \textbf{Temporal} and \textbf{Style}, while still being limited in \textbf{Color}.
    
%     \item \textbf{Open-source models are competitive on certain subtasks.} The Qwen3 series achieve strong results, especially in \textbf{Temporal} and \textbf{Color}, but still lags behind closed-source models in \textbf{Style}.
    
%     \item \textbf{There is a clear imbalance across task categories.} \textbf{Temporal} is the best-performing category overall, whereas \textbf{Color} and \textbf{Style} remain the major challenges for current models.
    
%     \item \textbf{Model scale does not guarantee consistent gains across all tasks.} Qwen3-Omni-30B-A3B outperforms Gemini-3-Pro on \textbf{Temporal}, suggesting that architectural design and multimodal alignment strategies play a critical role beyond model scale alone.
% \end{itemize}

\section{Related Works}

\textbf{Web-Based Information Retrieval} Web-Based Information Retrieval involves utilizing web resources such as search engines, web pages, and external tools to autonomously acquire and synthesize information~\citep{yu2026seeingbelievingbenchmark,wang2025hitchhiker}. Systems like Search-R1 combine search engine queries with reasoning abilities, continuously refining answers through dynamic search interactions. WebDancer adopts a curriculum-driven strategy, guiding the agent through web-based tasks to perform multi-step reasoning and collect information from multiple sources~\citep{yu2026allinoneagentbenchmarkingrolespecialized,zhou2025jpro}. WebWatcher further extends this by integrating tools such as text and image search, code interpretation, and webpage browsing, enabling the agent to handle more complex multi-step queries~\citep{yu2026paperxunifiedframeworkmultimodal}. Despite these advancements, research in this area remains primarily focused on text or static multimodal information retrieval, with limited systematic exploration of video shot retrieval tasks that involve temporal structure and semantic complexity~\citep{yu2026closedpoolvideoretrievalbenchmark}. To address this gap, we propose a benchmark for video shot retrieval in real-world editing scenarios, which includes 6 tasks and covers 20 topics.    

%基于网络的信息检索涉及利用搜索引擎、网页和外部工具等网络资源来自主获取和综合信息。像 Search-R1 这样的系统结合了搜索引擎查询与推理能力，通过动态的搜索交互不断优化回答。WebDancer 采用课程驱动策略，指导代理通过网络任务执行多轮推理，从多个来源收集信息。WebWatcher 在此基础上进一步扩展，整合了文本和图像搜索、代码解释和网页浏览等工具，使代理能够解决更复杂的多步骤查询。尽管取得了这些进展，这类研究现在仍主要集中于文本或静态多模态信息检索，对于更具时序结构和语义复杂性的视频镜头检索问题仍缺乏系统性的探索。为了解决这个问题，我们提出一个面向真实剪辑检索场景的视频镜头检索 benchmark，包括6个任务，涵盖20个主题。

\begin{figure*}[t]
 
\centering
\includegraphics[width=\linewidth]{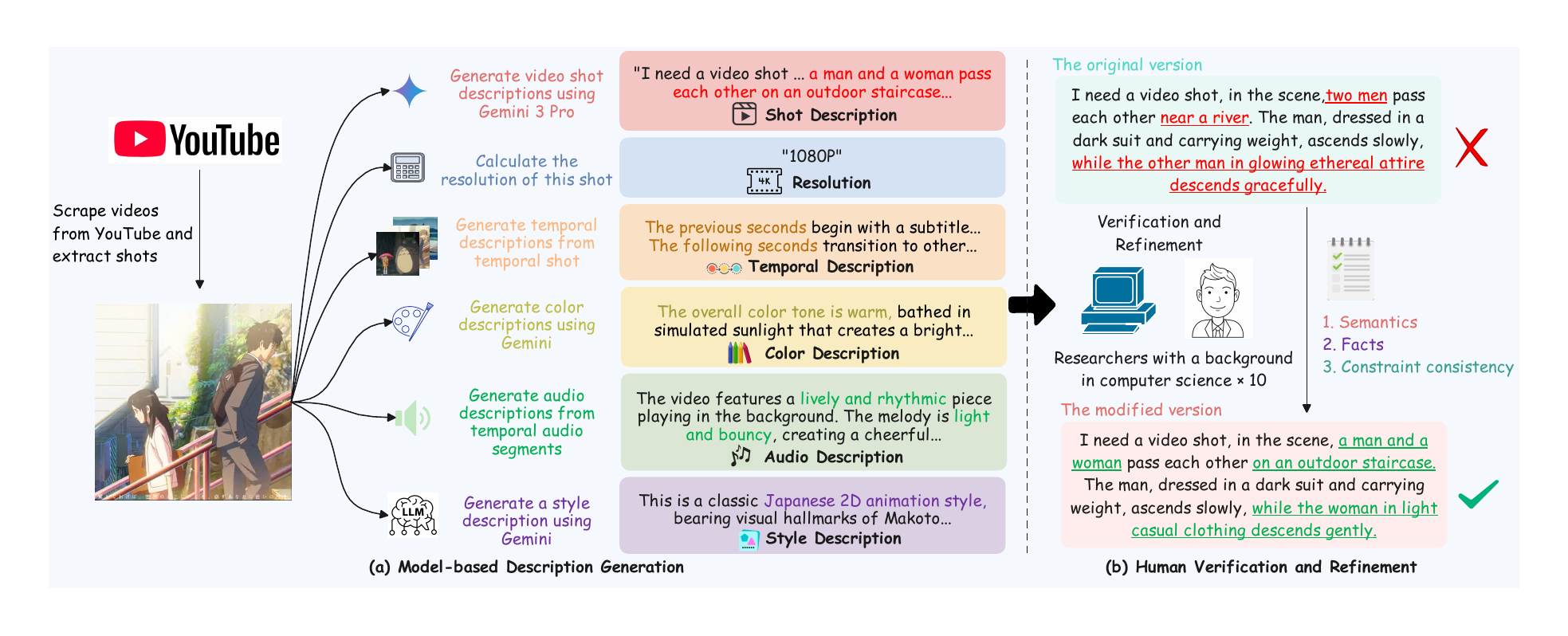}        %这个是在LaTeX文件夹中的相对路径
\caption{ShotFinder construction pipeline. (a) \textbf{Model-based Description Generation}, where Gemini generates shot descriptions with five specific constraints (Resolution, Temporal, Color, Style, Audio); and (b) \textbf{Human Verification and Refinement}, where experts refine the data to ensure semantic accuracy and constraint consistency.}
\label{2}
\vspace{-10pt}
\end{figure*}

\textbf{Video Temporal Grounding} Video Temporal Grounding (VTG) aims to localize specific moments in videos from natural language queries. Early methods such as Moment-DETR \citep{lei2021qvhighlightsdetectingmomentshighlights} and CONE \citep{hou2023coneefficientcoarsetofinealignment} rely on specialized architectures for short and long videos, respectively. Recently, MLLMs have become dominant: VTimeLLM \citep{huang2023vtimellmempowerllmgrasp} enables fine-grained temporal reasoning, TimeChat \citep{ren2024timechattimesensitivemultimodallarge} introduces timestamp-aware encoding for long videos, and Grounded-VideoLLM \citep{wang2025groundedvideollmsharpeningfinegrainedtemporal} improves temporal grounding through segment-level perception. However, existing VTG methods assume the target video is already available, limiting their use when videos must first be retrieved from external sources. Our work addresses this gap by coupling open-domain video retrieval with temporal grounding.

%目前，短视频和长视频的时序定位（VTG）任务已有一定进展。对于短视频，像Moment-DETR等方法通过变换器架构有效进行时刻预测，能够精准定位广告中的产品展示或体育赛事中的关键动作。非DETR方法，如UMT和Mr.BLIP，通过多模态线索和大型语言模型结合视觉与文本信息，进一步提升了短视频时刻预测的准确性。对于长视频，Temporal-Transformer等方法采用粗到细的对齐和分层管道技术，有效解决了长视频中时刻稀疏的问题，能够定位电影或电视剧中的关键情节。然而，这些方法通常假设视频已经存在，并且仅在已知视频的前提下执行任务，当视频尚未获得时，便无法发挥作用，限制了其应用范围。为此，我们提出了一种新的方法，将视频时序定位（VTG）与视频检索（Search）相结合，不仅解决了检索相关视频的问题，还将时序定位扩展到一个全新领域，允许模型在检索到的视频中进行时序定位，从而推动视频检索与时序定位任务的深度融合。

\section{ShotFinder Benchmark}

\subsection{Task Definition}

% 给定一个视频镜头关键画面的详细描述，可能带有时序、色彩、风格、分辨率等信息，任务目标是在互联网上找到符合描述的视频镜头，该任务面临以下的独特挑战：
% 1. 信息的不对等：输入的描述是视频镜头的关键画面，而搜索得到的是视频本身，相当于在拿视频1/10到1/100的信息量来寻找视频。
% 2. 信息的提取：输入的描述非常详细，不可能直接用于搜索，这就要求模型具有合理的提取关键词的能力，尽可能的将描述中的信息浓缩
% 3. 视频的信息冗余：与仅关注文本或图像的搜索不同，视频这个模态具有更多的信息量，对模型的上下文窗口的长度要求很高。

% 本文提出一种开放域视频镜头检索任务：给定目标视频镜头的关键画面所对应的细粒度自然语言描述（可包含主体/动作/场景等内容信息，并可能附带时序、色彩、风格、声音、分辨率等约束），要求系统在互联网中检索并返回与描述语义一致且满足约束的目标视频镜头。该任务面临以下的独特挑战：
% 1. 信息不对等（粒度错配）：输入是镜头关键画面的局部信息，而检索对象是完整视频，等价于用视频极少部分（约 1/10∼1/100 的信息量）去匹配整段视频，有可能出现视频整体主题相近但缺少描述要求的关键瞬间。
% 2. 信息提取与压缩：描述通常非常细、很长，无法直接作为检索查询；模型需要从中提取可检索的关键实体/属性/动作，并区分强约束与弱描述，将其压缩为有效的检索表示（关键词或向量）。
% 3. 视频冗余与长上下文压力：视频包含大量无关帧与多段事件，既要从冗余内容中定位匹配瞬间，又要建模时序线索；这对帧采样、跨帧聚合以及模型可处理的上下文长度提出更高要求。

This paper proposes an open-domain shot retrieval task: given a fine-grained natural language description corresponding to the key frame of the target shot (which may include subject/action/scene information, and possibly constraints such as temporal sequence, color, style, audio, resolution, etc.), the system is required to search the internet and return video shots that are semantically consistent with the description and satisfy the given constraints. This task faces the following unique challenges:

1. \textbf{Information imbalance}. The input consists of partial information of the key frame of the shot, while the search target is the entire video. This is equivalent to using a small portion of the video’s information to match the full video. As a result, it may happen that the overall theme of the video is close, but the key moment described in the query is missing.

2. \textbf{Information extraction and compression}. The descriptions are usually very detailed and long, making them unsuitable for direct retrieval queries. The model needs to extract searchable key entities/attributes/actions from the description, distinguish between strong constraints and weak descriptions, and compress them into effective retrieval representations (e.g. keywords).

3. \textbf{Video redundancy and long context pressure}. Videos contain a large number of irrelevant frames and multiple event segments. The system must both locate the matching moment amidst redundant content and model the temporal clues. This presents higher requirements for frame sampling, cross-frame aggregation, and the context length that the model can handle.

\subsection{Data Collection}

% 数据收集：
% 1. 分类：分为6类，镜头描述/时序描述/色彩描述/风格描述/声音描述/分辨率，在这里我们只讨论镜头描述/镜头描述+X的影响
% 2. 主题，在这里讲主题对应关系
% 3. 生成流程，两阶段，然后风格和色彩单独处理

\subsubsection{Task Design}

% 该分类体系源于真实剪辑检索场景：剪辑师在检索素材时，最基础的需求是“画面内容是否满足描述”，而更高层的需求则涉及“镜头是否具有特定叙事功能或视听呈现效果”。为此，我们将检索任务划分为基础剪辑与在基础剪辑之上引入额外约束的扩展设置。

% 1. 基础剪辑（Basic Editing）

% 基础剪辑指检索目标仅要求候选镜头满足关键画面内容的描述，即描述主要由人物/物体属性、动作行为、场景元素等组成，判定依据是画面内容是否一致。例如：“一个年长男性拥有银灰色中分长发，肤色较深，神情慈祥而专注，身穿深色立领正装，正全神贯注地发表讲话。” 在该设置中，只要候选镜头的关键画面能够在语义层面满足上述内容，即视为匹配。

% 2. 基础剪辑 + 单一因素约束（Basic + One Constraint）

% 在真实剪辑中，许多“可用镜头”的选择不仅取决于画面里“有什么”，还取决于镜头在叙事中的位置、整体氛围、视听风格或技术规格等。为了在研究中可控地分析这些因素对检索的影响，我们并未直接覆盖所有复杂组合的故事性约束，而是采用“在基础剪辑上仅额外加入一种因素”的设计，将扩展任务限定为以下五种单因素设置：

% 基础剪辑 + 前后文（Temporal/Context Constraint）：要求镜头满足与前后镜头或动作阶段相关的条件（如“发言前起身”“发言后鼓掌”等）。

% 基础剪辑 + 色彩（Color Constraint）：要求画面整体色调、冷暖倾向等符合描述，以刻画不同色彩带来的氛围差异。

% 基础剪辑 + 风格（Style/Medium Constraint）：要求真人实拍、3D/2D 动画等视觉风格符合描述，以反映媒介与风格对观感的影响。

% 基础剪辑 + 声音（Audio Constraint）：要求存在特定声音元素或声画关系（如说话声、背景音乐、环境音等），以引入跨模态约束。

% 基础剪辑 + 分辨率（Resolution Constraint）：要求清晰度、画质或画幅等技术属性符合描述，以反映素材可用性层面的限制。

% 因此，我们的研究共包含 6 种任务设置：
% 基础剪辑，以及 基础剪辑分别 +（前后文 / 色彩 / 风格 / 声音 / 分辨率）。
% 在本文中，我们不进一步考虑多因素组合（如“基础剪辑+前后文+色彩”或更多因素叠加）的复杂情形。

% 为什么只研究“单因素扩展”，不研究多因素组合？（这里加粗）

% 我们做出上述限定主要基于方法论与实验可解释性的考虑：

% 可控性与归因性：若直接引入多因素组合，性能变化难以归因于某一个因素；单因素设计能够更清晰地回答“某一因素引入后，检索难点在哪里、性能下降多少”。

% 评测可重复与边界清晰：单因素约束的匹配标准更容易严格定义与标注，从而降低主观性，提高可复现性。

% 复杂度分层的第一步：多因素故事性剪辑属于更高阶、更接近真实制作的综合约束问题。本文将其视为后续工作方向，当前先通过单因素拆解建立清晰的能力剖面，为未来组合约束的研究提供基线与方法。

This classification system is derived from real-world editing retrieval scenarios: when editors search for footage, the basic need is "whether the content of the shot meets the description," while higher-level needs involve "whether the shot has a specific narrative function or audiovisual presentation effect." To address this, we divide the retrieval task into shot descriptions and factor constraints, with the latter introducing additional constraints on top of the shot descriptions.

\textbf{Shot Description} Shot Description refers to retrieval tasks where the candidate shot is only required to match the description of the shot. The description mainly includes attributes of people/objects, actions, and scene elements, and the judgment is based on whether the visual content is consistent. For example: "An older male with silver-gray, center-parted long hair, darker skin tone, wearing a dark high-collared suit, giving a speech with full attention." In this setting, as long as the key frame of the candidate shot semantically matches the above content, it is considered a match.

\textbf{Shot Description + Single Constraint} In real-world editing, the selection of "usable footage" depends not only on "what's in the shot" but also on the shot's position in the narrative, overall atmosphere, or audiovisual style. To controllably analyze the impact of these factors on retrieval, we do not directly cover all the complex combinations of narrative constraints. Instead, we adopt a design where "only one additional factor is added to shot description." This extends the task to the following five single-factor setups:

These constraints include \textbf{Temporal}, which require the shot to align with conditions related to preceding or following shots or action phases. We also consider \textbf{Color}, where the overall tone and warmth/coolness of the shot must match the description, capturing the atmospheric differences that color can convey. Additionally, \textbf{Style} are included, where the visual style, such as live-action footage or 3D/2D animation, must align with the description to reflect the impact of style on perception. \textbf{Audio} are also introduced, requiring the presence of specific sound elements or audiovisual relationships, such as speech, background music, or environmental sounds, to introduce cross-modal constraints. Finally, \textbf{Resolution} are applied, ensuring that technical attributes like clarity meet the requirements outlined in the description, reflecting limitations on the usability of the footage.

Therefore, our research includes 6 task settings.
Shot Description, and Shot Description with the additional constraints of (\textbf{Temporal}, \textbf{Color}, \textbf{Style}, \textbf{Audio}, \textbf{Resolution}).
% In this paper, we do not further consider the complex cases of multi-factor combinations (e.g., "Shot Description + Temporal + Color" or other combinations of factors).

% \textbf{Why focus only on "Single Constraint" and not multi-factor combinations?} The primary limitation of our study is based on methodological and experimental interpretability considerations: 

% \textbf{Control and Attribution} If multi-factor combinations were directly introduced, performance changes would be difficult to attribute to any single constraint. A single-factor design can more clearly answer "What challenges does retrieval face after introducing a specific constraint, and how much does performance change?"

% \textbf{Repeatability of Evaluation and Clear Boundaries} The matching standards for single-factor constraints are easier to strictly define, thereby reducing subjectivity and improving reproducibility. 

% \textbf{Layered Complexity} Multi-constraint shot retrieval is a more complex constraint problem, which we regard as a future research direction. Currently, by breaking it down into single factors, we aim to establish a clear standard, providing a benchmark and methodology for future research.

\begin{figure*}[t]
 
\centering
\includegraphics[width=0.75\linewidth]{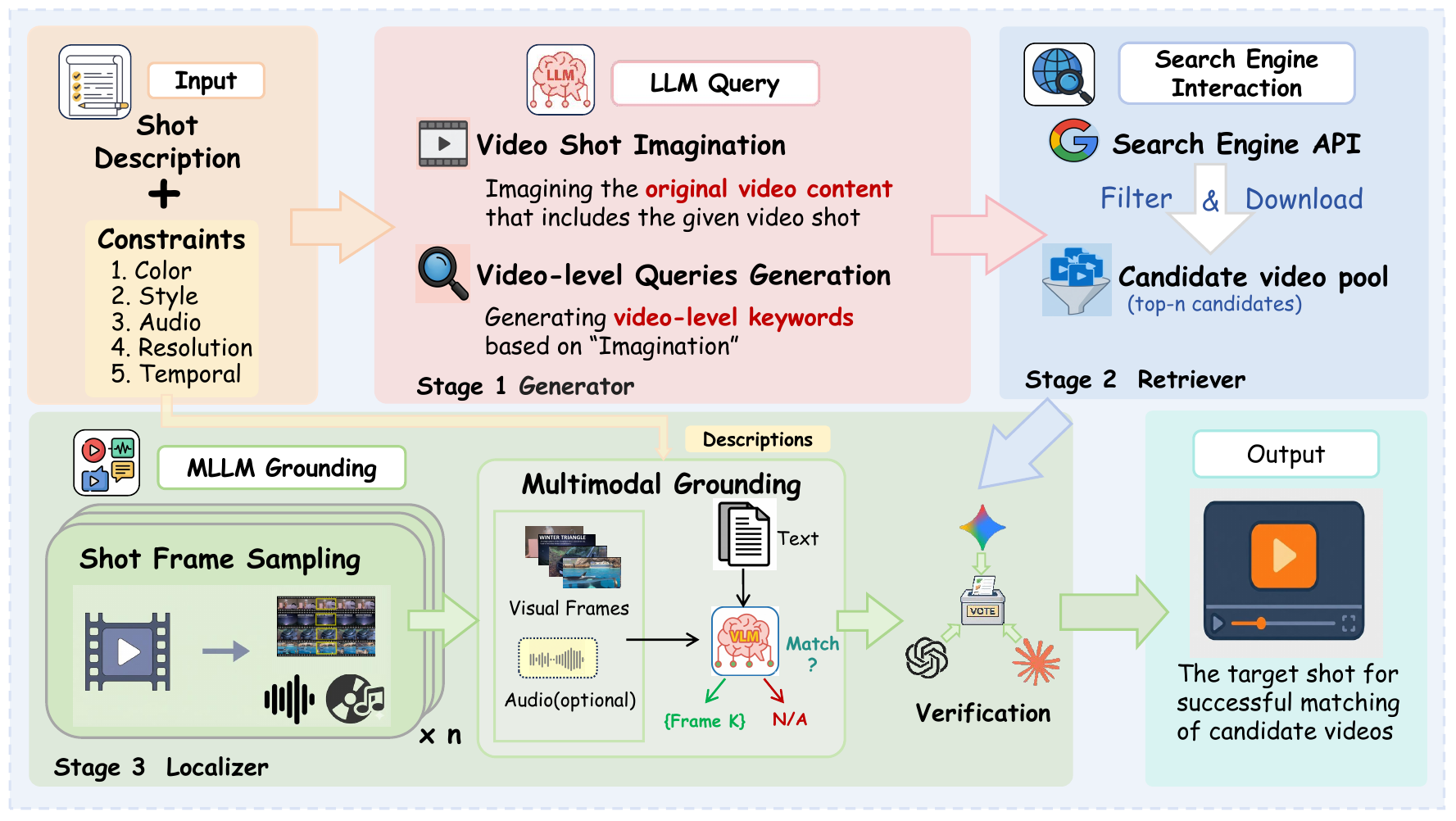}        %这个是在LaTeX文件夹中的相对路径
\caption{Illustration of the ShotFinder method pipeline. (1) \textbf{Generator}, utilizing "Video Imagination" to expand shot descriptions into effective search queries; (2) \textbf{Retriever}, fetching and filtering candidate videos from the web; and (3) \textbf{Localizer}, employing MLLMs with adaptive frame sampling to precisely locate the target shot.}
\label{3}
\vspace{-10pt}
\end{figure*}

\subsubsection{Topic Allocation Strategy}

% 不同因素约束对视频内容的适配性存在显著差异。例如，色彩描述更适用于具有明确视觉风格的视频，而声音描述则更依赖于音频线索。若在所有任务中强制使用统一的主题集合，可能引入大量与描述约束弱相关甚至不适配的样本，从而干扰对描述维度本身的分析。基于上述考虑，我们在每一种任务约束下分别选择与其语义高度相关的主题集合，并在该集合内部尽可能覆盖多样化的视频内容形态。例如，色彩描述任务侧重于时尚、艺术与旅行等具有显著视觉特征的主题，而声音描述任务则更多包含电视剧、纪录片与舞蹈等音频信息丰富的内容。同时，我们避免将某一主题与单一任务类型形成一一对应关系，而是通过多主题、多内容形态的设计，降低视频内容本身对评测结果的主导影响。详细分类方法请见附录。

% 对于具体分类，我们将时序描述对应为知识、健身、美食、体育、汽车、游戏、电影、电视剧这些主题，色彩描述对应为育儿、时尚、科技、艺术、纪录片、旅行这些主题，风格描述对应为综艺节目、动画、游戏、音乐这些主题，声音描述对应为电视剧、纪录片、舞蹈、动物这些主题，分辨率对应为时尚、电影、生活Vlog、健身这些主题。

% 不同因素约束在语义和感知层面对视频内容的适配性存在显著差异。例如，色彩相关描述通常依赖于明确的视觉外观与色彩分布，而声音相关描述则更侧重于音频线索及其语义信息。若在所有任务设置中强制采用统一的主题集合，容易引入大量与特定描述约束弱相关甚至不适配的样本，从而削弱对描述维度本身影响的分析，甚至引入额外噪声。

% 基于上述考虑，我们在数据构建阶段采用一种约束感知的主题分配策略（constraint-aware topic allocation）。具体而言，对于每一种因素约束，我们分别选择与其语义特性高度相关的主题集合，并在该集合内部尽可能覆盖多样化的视频内容形态，以避免单一内容类型对评测结果产生主导影响。同时，我们刻意避免将某一主题与单一任务类型形成一一对应关系，而是通过在同一约束下引入多个主题，实现任务约束与视频内容之间的弱耦合，从而降低视频内容本身带来的偏置。

% 在视频主题的选择上，我们参考了主流视频网站 Bilibili 的内容分类体系，最终选取了 20 种常见视频主题，涵盖综艺、动画、舞蹈、科技数码、美食、汽车、体育、生活 Vlog、电影、电视剧、纪录片、游戏、音乐、知识、时尚、动物、绘画、旅游、健身与亲子等内容类型。

% 在具体实现上，基础剪辑描述不施加额外因素约束，因此覆盖全部主题类别，以保证其内容分布的广泛性。对于引入因素约束的任务，我们根据约束类型选择相应的主题集合：时序描述主要对应知识、健身、美食、体育、汽车、游戏、电影和电视剧等具有明确事件结构或动作流程的主题；色彩描述侧重于育儿、时尚、科技、艺术、纪录片和旅行等具有显著视觉特征的主题；风格描述涵盖综艺节目、动画、游戏和音乐等风格属性突出的内容类型；音频描述主要涉及电视剧、纪录片、舞蹈和动物等音频信息丰富的主题；分辨率相关描述则对应时尚、电影、生活 Vlog 和健身等对视觉清晰度较为敏感的内容。更详细的分类规则见附录。

Different constraint factors exhibit significant differences in their semantic and perceptual adaptability to video content. For example, \textbf{Color} typically relies on clear visual appearance and color distribution, while \textbf{Audio} focuses more on audio cues and their semantic information. If a unified set of topics is forced across all tasks, it could introduce many samples that are weakly related or even mismatched with specific constraints, thereby weakening the analysis of the influence of the constraints and potentially introducing additional noise.

Based on these considerations, we adopt a constraint-aware topic allocation strategy. Specifically, for each type of constraint, we select a topic set that is highly relevant to its semantic characteristics, and within that set, we aim to cover as diverse a range of video content forms as possible. At the same time, we deliberately avoid establishing a one-to-one correspondence between topics and specific constraints, achieving weak coupling between constraints and video content, thereby reducing the bias introduced by the content itself.

In selecting video topics, we refer to the content classification of the mainstream video platform Youtube. We ultimately selected 20 common video topics, covering genres such as Variety Shows, Animation, Dance, Tech, Food, Automotive, Sports, Lifestyle Vlogs, Film, TV series, Documentary, Gaming, Music, Knowledge, Fashion, Animals, Visual Arts, Tourism, Fitness, and Parenting.

In practical implementation, Shot Description does not impose additional constraint factors, so they cover all topic categories to ensure the broadness of content distribution. For tasks with added constraint factors, we select the appropriate topic set based on the type of constraint: \textbf{Temporal} mainly corresponds to topics with clear event structures or action flows, such as knowledge, fitness, food, sports, automobiles, games, movies, and TV dramas; \textbf{Color} focuses on topics with significant visual features, such as parenting, fashion, technology, art, documentaries, and travel; \textbf{Style} covers content types with prominent style attributes, such as variety shows, animation, games, and music; \textbf{Audio} primarily involves topics with rich audio information, such as TV dramas, documentaries, dance, and animals; and \textbf{Resolution} corresponds to content types that are more sensitive to visual clarity, such as fashion, movies, lifestyle vlogs, and fitness. For detailed classification rules, see Appendix \ref{theme_classification}.

\subsubsection{Description Generation Pipeline}

\textbf{Model-based Description Generation} As shown in Figure \ref{2}(a), to efficiently construct a large-scale and structurally consistent description dataset, we crawl videos from YouTube and use large models for automated description generation. In this phase, we use Gemini-3-Pro to generate Shot Description, as well as various constraint descriptions (such as \textbf{Temporal}, \textbf{Color}, \textbf{Style}, \textbf{Audio}, and \textbf{Resolution}). The generated descriptions provide initial candidate texts for subsequent manual verification. The specific process details and generation templates are provided in Appendix \ref{data_shengcheng}.

% 为高效构建大规模且结构一致的描述数据，我们从Youtube爬取视频，采用大模型进行自动化描述生成。在该阶段，我们使用 Gemini 3 Pro 生成视频镜头的基础剪辑描述以及对应的多种约束描述（如时序、色彩、风格、声音与分辨率）。模型输入基于前述任务设计与主题分配策略构建，生成的描述覆盖视频镜头的核心内容及相应约束维度，从而为后续人工校验提供初始候选文本。具体的生成模板与流程细节详见附录。

\textbf{Human Verification and Refinement} As shown in Figure \ref{2}(b), to ensure the accuracy and consistency of the data, we introduce a human verification and refinement phase. Specifically, we recruited 10 researchers with backgrounds in computer science to review each of the model-generated descriptions manually. During the verification process, annotators were required to assess the correctness of the descriptions in terms of semantic, factual, and constraint consistency, and to revise any errors or imprecise statements they identified. Through this manual correction and revision process, we effectively eliminated the noise introduced during the automated generation phase, ultimately constructing a dataset of 1,210 high-quality samples.

% 为确保数据的准确性与一致性，我们进一步引入人工校验与修订阶段。具体而言，我们招募了 10 名具有计算机相关研究背景的研究人员，对模型生成的描述进行逐条核查。校验过程中，标注人员需结合视频内容判断描述在语义、事实与约束一致性方面的正确性，并对发现的错误或不严谨表述进行修订。通过该人工订对与修正流程，我们有效消除了自动生成过程中引入的噪声，最终构建得到包含 1210 个高质量样本的数据集，用于后续实验与评测。

\subsection{Dataset Statistics}

% 如图所示，benchmark中包含1210个数据，这些数据均匀覆盖了6个任务，覆盖了20个主题，数据来源视频的时间分布很广泛。

% 如图所示，基准数据集包含1210个样本，这些样本均匀分布在六个任务上，涵盖了20个不同的主题，确保了内容的多样性。用于构建该数据集的视频涵盖了广泛的时长，反映了各种视频类型和格式。这种分布为评估模型在不同领域、约束条件和内容形式下的表现提供了全面的基础。

As shown in the Figure \ref{data}, the benchmark dataset consists of 1,210 samples, evenly distributed across six tasks. These tasks cover a diverse range of 20 topics, ensuring broad content representation. The videos used for this dataset span a wide range of durations, reflecting a variety of video types. This distribution provides a comprehensive foundation for evaluating the model’s performance across different domains, constraints, and content forms.

\subsection{Evaluation Metrics}

% 我们首先尝试采用基于 CLIP 的相似度度量来评估模型检索结果与目标视频镜头之间的一致性。然而，在初步实验中我们发现，该类基于全局视觉–文本嵌入的度量方式难以有效刻画细粒度的描述约束，尤其是在涉及时序、风格或多属性组合描述的情况下，其评测结果与人工判断存在明显偏差。相关分析与对比结果已在消融实验部分中给出。

% 基于上述观察，我们进一步采用大模型辅助评测作为主要评估方式。具体而言，考虑到直接对视频镜头进行评测的成本与复杂度，我们使用视频镜头的关键画面（keyframes）作为视频内容的近似表示。对于每个检索结果，我们将其对应的关键画面、输入的文本描述以及 ground-truth 描述一并输入至 Gemini 模型，由其判断检索到的视频镜头是否在语义和约束层面符合给定描述。评测prompt详见附录。

% 该评测方式能够综合考虑视觉内容与文本描述之间的细粒度匹配关系，尤其适用于多约束条件下的视频搜索任务，从而为不同模型在复杂描述条件下的性能比较提供更具判别力的评价依据。

We first attempted to use CLIP-based \citep{radford2021learningtransferablevisualmodels} similarity metrics to evaluate the consistency between the model's retrieval results and the target shots. However, in preliminary experiments, we found that this type of global visual-text embedding metric struggled to effectively capture fine-grained description constraints. The evaluation results showed significant discrepancies with human judgment. Relevant analysis and comparison results are provided in Appendix \ref{human}.

% Based on these observations, we further adopted large-model-assisted evaluation as the primary evaluation method. Specifically, considering the cost and complexity of directly evaluating video shots, we use keyframes of the video shots as an approximate representation of the shot content, with the validity of the keyframes detailed in Section \ref{keyframe}. For each retrieval result, we input the corresponding keyframe, the input textual description, and the ground-truth frame into MLLM, which then determines whether the retrieved video shot meets the given description in terms of both semantics and constraints. The evaluation prompt is in Appendix \ref{pingce}. 

Based on these observations, we further adopt large-model-assisted evaluation as the primary evaluation method. Specifically, considering the cost and complexity of directly evaluating video shots, we use the keyframes of video shots as an approximate representation of the shot content, with the validity of the keyframes detailed in Appendix \ref{keyframe}. For each retrieval result, we input the corresponding keyframe, the input textual description, and the ground-truth frame into Gemini-2.5-Pro, GPT-5.2, and Claude-4.0-Sonnet, and let them vote to determine whether the retrieved video shot satisfies the given description in terms of both semantics and constraints. The evaluation prompt is provided in Appendix \ref{pingce}.

% This evaluation method can comprehensively consider the fine-grained matching relationship between visual content and textual descriptions, making it particularly suitable for video search tasks under multiple constraint conditions. It provides a more discriminative evaluation basis for comparing the performance of different models under complex description conditions.

\section{ShotFinder Method}

\subsection{Overview}

% ClipFinder 是一个基于文本描述的视频镜头检索与定位流程。给定镜头描述（及可选的附加约束），该方法依次执行查询扩展、候选视频检索以及基于描述的时间定位三个阶段。首先，Generator 模块通过视频联想机制生成视频级搜索查询，以提升检索召回率；随后，Retriever 模块从网络视频平台获取候选视频集合；最后，Localizer 模块在候选视频中基于描述对目标镜头进行时间定位。该分阶段设计将镜头级描述与开放域视频检索有效结合，实现了对复杂描述条件的稳健处理。

ShotFinder is a video shot retrieval and localization process based on textual descriptions. Given a shot description (and optional additional constraints), this method performs three stages: query expansion, candidate video retrieval, and description-based temporal localization, as shown in Figure \ref{3}. First, the Generator module generates video-level search queries through a video association mechanism. Next, the Retriever module fetches a candidate video set from search engines. Finally, the Localizer module performs temporal localization of the target shot based on the description. This staged design effectively combines shot-level descriptions with open-domain video retrieval.

\begin{table*}[t]
\centering
\resizebox{0.73\textwidth}{!}{
\begin{tabular}{lccccccc}
\toprule
 & \textbf{Shot} & \textbf{Temporal} & \textbf{Color} & \textbf{Style} & \textbf{Resolution} & \textbf{Audio} & \textbf{Avg.} \\
\midrule
Human & 85.1 & 91.6 & 91.4 & 83.3 & 91.7 & 87.5 & 88.5 \\
\midrule
\multicolumn{8}{l}{\textbf{Closed-source Models}} \\
\hdashline
Gemini-3-Pro & 22.5 & 31.0 & 15.7 & \underline{26.5} & \underline{21.0} & \textbf{30.0} & \underline{24.4} \\
Gemini-2.5-Pro & \underline{23.5} & 29.5 & 12.9 & 17.5 & 18.5 & \underline{26.5} & 21.3 \\
GPT-5.2 & \textbf{25.5} & \textbf{35.5} & 15.7 & \textbf{32.5} & \textbf{26.0} & -- & \textbf{26.9} \\
GPT-5-mini & 12.0 & 21.0 & 10.5 & 17.0 & 16.0 & -- & 15.2 \\
Claude-4.0-Sonnet & 18.0 & 30.0 & 8.1 & 15.5 & 19.5 & -- & 18.1 \\
\midrule
\multicolumn{8}{l}{\textbf{Open-source Models}} \\
\hdashline
Qwen3-Omni-30B-A3B & 19.0 & \underline{33.0} & 11.4 & 18.5 & 17.5 & 21.0 & 20.0 \\
Qwen3-VL-235B-A22B & 18.5 & 28.5 & \textbf{20.5} & 17.0 & 18.5 & -- & 20.6 \\
InternVL3.5-241B-A28B   & 20.0 & 20.8  & \underline{16.7} & 20.8 & 20.8 & --     & 19.8 \\
VideoLLaMA3-7B  & 15.0  & 12.5 & 12.5  & 16.7 & 16.7 & --     & 14.7 \\
\bottomrule
\end{tabular}}
\caption{Performance comparison of closed-source and open-source models on ShotFinder benchmark.}
\label{tab:main_results}
\vspace{-10pt}
\end{table*}

\subsection{Generator: Query Expansion via Video Imagination}

% 在给定视频镜头描述（及可选的附加约束描述）的情况下，一种直接的做法是要求大模型对输入文本进行概括或关键词提取，并将其作为搜索查询。然而，在初步实验中我们发现，该策略往往局限于镜头级的局部语义，难以有效覆盖真实视频标题、标签或描述中常见的高层次概念，从而导致检索召回率较低。

% 为缓解上述问题，我们提出一种基于**视频联想（video imagination）**的查询扩展策略。具体而言，我们引导模型首先推断该镜头最有可能出现于何种完整视频场景中，即让模型“想象”包含该镜头的原始视频内容（如视频主题、整体叙事或典型语境），再基于该想象生成用于检索的搜索关键词。与简单的文本概括不同，该过程显式地引入了视频级语义，从而使生成的查询更贴近真实在线视频平台中视频的组织与标注方式。

% 通过将镜头级描述映射到潜在的视频级语义空间，该策略能够有效提升搜索查询的覆盖性与表达能力。我们在消融实验中进一步对比了直接关键词生成与基于视频联想的查询扩展策略，实验结果验证了后者在检索效果上的显著优势。

Given a video shot description (and optional additional constraint descriptions), one straightforward approach is to ask a large model to summarize the input text or extract keywords, using them as search queries. However, in preliminary experiments, we found that this strategy tends to be limited to the local semantics of the shot, making it difficult to effectively cover high-level concepts commonly found in real video titles, tags, or descriptions, resulting in low retrieval recall.

To address this, we propose a query expansion strategy based on video imagination. Specifically, we guide the model to first infer what type of complete video the shot is most likely to appear in—i.e., we ask the model to "imagine" the original video content that includes the shot (such as the video’s theme), and then generate search keywords based on this imagination. Unlike text summarization, this process incorporates video-level semantics, making the query more aligned with how videos are organized and labeled on online platforms. The imagination prompt is in Appendix \ref{imagination}.

% By mapping shot-level descriptions to the potential video-level semantic space, this strategy effectively enhances the coverage and expressiveness of the search query.

\subsection{Retriever: Web Video Retrieval}

% 在生成搜索Expansion后，我们通过网络搜索接口执行视频检索，以构建候选视频集合。具体而言，我们使用生成的关键词与搜索引擎进行交互，获取与查询相关的公开视频链接，并从中筛选出 n 个可正常访问的 YouTube 视频 URL 作为候选结果。

% 为保证后续处理的稳定性与可复现性，我们仅保留能够成功访问且视频内容完整的链接，并将对应的视频文件下载至本地，用于后续基于描述的镜头定位。该检索过程旨在在保证较高召回率的同时，控制候选视频规模，从而为后续的精细化定位提供可管理的搜索空间。

After generating the search expansion, we perform video retrieval through a search engine to construct a candidate video set. Specifically, we use the generated keywords to interact with the search engine, retrieving publicly available video links relevant to the query, and then filter out 
$n$ accessible YouTube video URLs as candidate results. 

To ensure the stability of subsequent processing, we retain only the links that are successfully accessible and contain complete video content, and download the corresponding video files locally for later description-based shot localization.

\subsection{Localizer: Description-guided Temporal Localization}

% 对于每个候选视频，我们采用基于视频时长的自适应抽帧策略，对视频内容进行离散化表示。具体而言，不同时长的视频对应不同的抽帧频率，以在保证时间覆盖的同时控制输入规模。抽取的关键帧用于近似表示视频的时序内容，为后续基于描述的镜头定位提供视觉输入。

% 在定位阶段，我们将生成搜索关键词的描述与抽取的关键帧共同输入至大模型，并要求模型判断哪些帧在语义层面与给定描述相匹配。对于涉及音频约束的任务设置，我们额外提供视频的音频信息作为输入。若模型判断候选视频中不存在符合描述的镜头，则输出 N/A。该过程旨在在候选视频范围内实现基于文本描述的粗粒度时间定位。

% 最后，我们进一步将生成搜索关键词的描述、ground-truth 图片以及模型选出的候选帧输入至 Gemini 2.5 Pro，由其判断模型是否成功定位到符合描述的视频镜头。该双阶段定位与验证机制有效降低了单次模型判断带来的不确定性，为最终检索结果提供更可靠的判定依据。

% For each candidate video, we employ an adaptive frame sampling strategy based on video duration. Specifically, videos of different lengths correspond to different frame sampling strategies to ensure temporal coverage while controlling the input size. The extracted frames are used to approximate the content of the video, providing visual input for the subsequent description-based shot localization.

For each candidate video, we employ an adaptive shot sampling strategy based on its duration. Specifically, we use TransNetV2 \citep{soucek2020transnetv2effectivedeep} to segment the video into continuous shots and extract the midpoint frame from each shot as a representative keyframe. These keyframes serve as an approximate representation of the video content, providing visual input for subsequent description-based shot localization.

During the localization phase, we input the description from which the search queries are generated and the extracted frames into MLLM, and ask the model to determine which frame matches the given description. For tasks involving \textbf{Audio}, we additionally provide the video's audio as input. If the model determines that no shot in the candidate video matches the description, it outputs N/A. 
% This process aims to achieve temporal localization based on the text description within the candidate video set.

Finally, we input the description from which the search queries are generated, the ground-truth frame, and the frame selected by the model into three different MLLMs, which vote on whether the localized shot matches the description. This two-stage localization and verification mechanism effectively reduces the uncertainty introduced by a single model judgment.

\section{Experiments}

\subsection{Baselines and Settings}

% 为了评估不同模型的性能，我们采用了闭源模型和开源模型作为基线。闭源模型包括Gemini 3 Pro、Claude 4.0 Sonnet、Gemini 3 Flash、GPT-5.2和GPT-5-mini，而开源模型则包括Qwen3-VL-235B-A22B-Instruct、Qwen2.5-VL-72B-Instruct和Qwen3-Omni-30B-A3B-Instruct。在本实验中，我们统一使用Gemini 2.5 Pro进行评测。对于闭源模型，视频帧提取的设置如下：对于时长小于3分钟的视频，均匀抽取64帧；对于时长在3到10分钟之间的视频，均匀抽取128帧；对于时长超过10分钟的视频，均匀抽取192帧。而对于开源模型，所有视频均均匀抽取64帧。所有模型生成了两个搜索查询，每个查询提取两个URL。

To evaluate the performance of different models, we use both closed-source and open-source models as baselines. The closed-source models include Gemini-2.5-Pro \citep{google2023gemini}, Gemini-3-Pro, Claude 4.0 Sonnet \citep{anthropic2024claude3}, GPT-5.2 \citep{openai2024gpt4}, and GPT-5-mini \citep{openai2024gpt4}, while the open-source models include Qwen3-VL-235B-A22B-Instruct \citep{yang2025qwen3technicalreport}, Qwen3-Omni-30B-A3B-Instruct \citep{xu2025qwen3omnitechnicalreport}, InternVL3.5-241B-A28B \citep{wang2025internvl35advancingopensourcemultimodal}, and VideoLLaMA3-7B \citep{zhang2025videollama3frontiermultimodal}. In this experiment, we consistently use Gemini-2.5-Pro, GPT-5.2, and Claude-4.0-Sonnet for evaluation. Each model generates two search queries, with two URLs extracted for each query. To maintain consistency, all models are accessed via API calls with the temperature set to 0.

% We refer to the video frame sampling settings used in VLMEvalKit \citep{duan2024vlmevalkit} and VideoMME \citep{fu2025videommefirstevercomprehensiveevaluation}. Specifically, for closed-source models, 64 frames are uniformly sampled from videos shorter than 3 minutes, 128 frames from videos between 3 and 10 minutes, and 192 frames from videos longer than 10 minutes. For open-source models, 64 frames are uniformly sampled from all videos.

\begin{figure}[t]
 
\centering
\includegraphics[width=\linewidth]{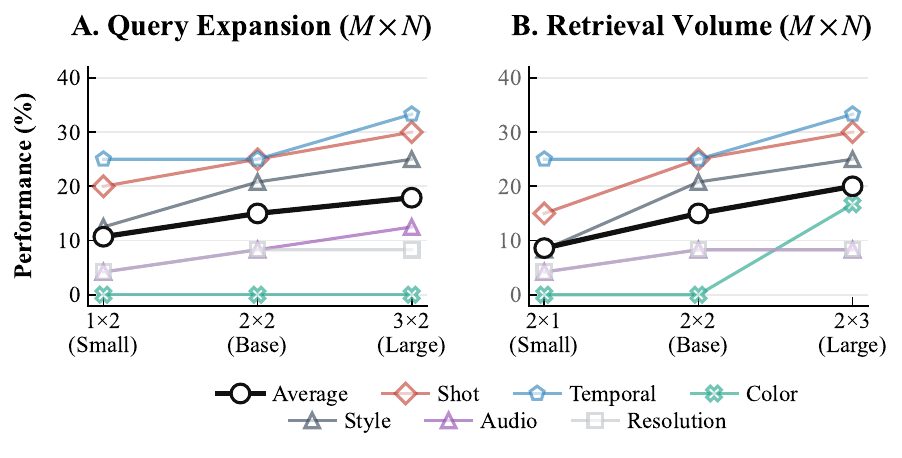}        %这个是在LaTeX文件夹中的相对路径
\caption{Results of further analysis. \textbf{Notation} ($\mathbf{M \times N}$) : $M =$ Search Queries, $N =$ Candidate URLs per Query.}
\label{fig:4}
\vspace{-10pt}
\end{figure}

\begin{figure}[t]
\centering
\includegraphics[width=\linewidth]{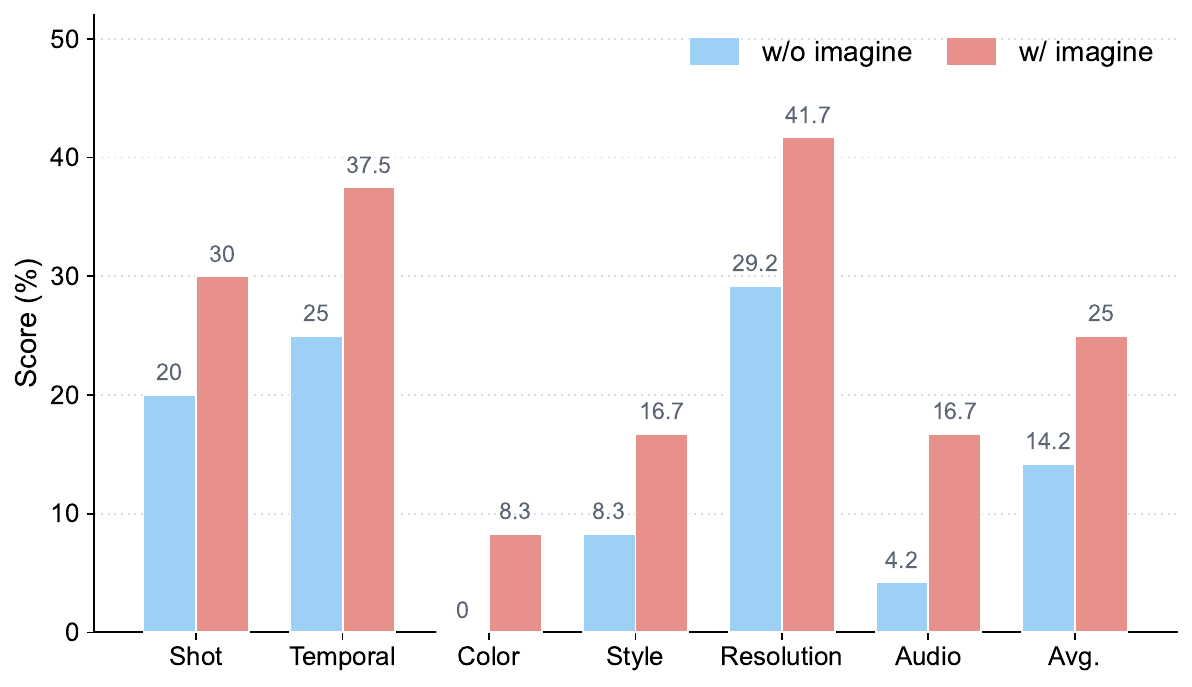}
\caption{Ablation on video imagination.}
\label{tab:imagination_ablation}
\vspace{-10pt}
\end{figure}

\begin{figure}[t]
\centering
\includegraphics[width=\linewidth]{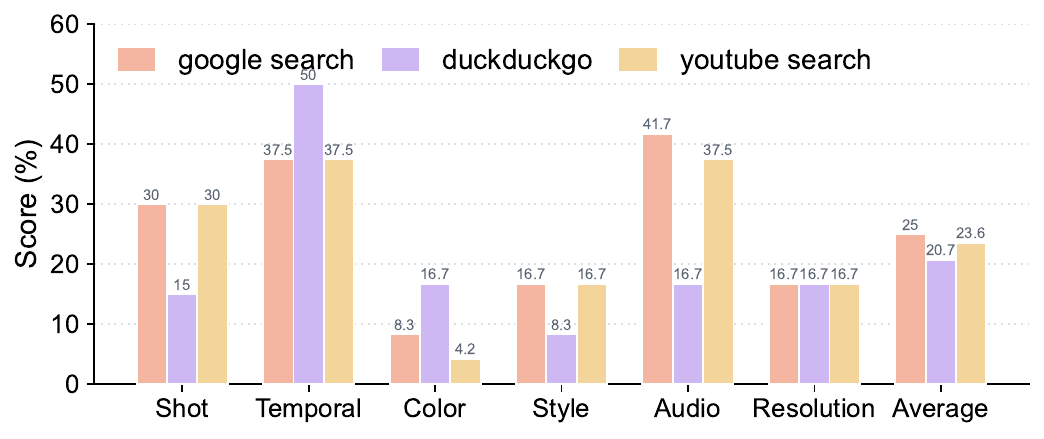}
\caption{Ablation on search engine.}
\label{fig:6}
\vspace{-10pt}
\end{figure}

\subsection{Main Results}

% 表1展示了不同模型在 ShotFinder 上的性能，核心结论如下：

% 人工 vs. 模型对比：人工评估在所有类别中的表现都处于最高水平，平均准确率为 88.5。这突显了基准测试的挑战性，因为即便是表现最好的模型，也明显低于人工评估的水平。人工评估能够全面考虑视频的细节，包括视觉、时间、颜色等各维度的细微变化，而现有模型往往在捕捉这些复杂细节时有所欠缺。

% 闭源模型：在闭源模型中，GPT-5.2 在所有模型中表现最好，平均准确率为 26.9，在大多数类别中表现强劲，尤其是在时序（35.5）和风格（32.5）方面。然而，它在色彩（15.7）上的表现相对较低。Gemini-3-Pro 紧随其后，平均得分为 24.4，在时序（31.0）和音频（30.0）方面表现突出，但在色彩分析（15.7）上落后。GPT-5-mini 和 Claude-4.0-Sonnet 显示出较低的整体表现，表明较小或不够专门化的模型在这种多模态任务中更为吃力。

% 开源模型：Qwen3 系列展示了具有竞争力的表现，其中 Qwen3-VL-235B-A22B 的平均得分为 20.6。尽管在色彩（20.5）上稍微超越 Gemini-3-Pro，但在时序和风格方面未超过闭源模型。Qwen3-Omni-30B-A3B 显示出类似的结果，平均得分为 20.0，其在时序（33.0）和声音（21.0）上表现较强。

% 表现趋势：时序类别在所有类别中表现最好，GPT-5.2 在时序推理方面取得了最高分（35.5）。色彩的得分普遍较低，但闭源和开源模型之间在这些类别中的差距并不大。而风格的得分呈现了比较大的两极分化，强模型能力强大很多。

% 模型扩展性：较大的模型并不一定在所有类别中都能带来更好的表现, Qwen3-Omni-30B-A3B在时序方面超过了Gemini-3-Pro。GPT-5.2 和 Gemini-3-Pro 展示了其较大架构下的稳定表现，但开源模型尽管规模较小，仍然能提供合理的结果。

Table \ref{tab:main_results} presents the performance of different models on ShotFinder, with the key conclusions:

\textbf{Human vs. Model Comparison} Human evaluation performs the best across all categories, and the best-performing models show significant discrepancies compared to human evaluation, highlighting the challenge of the benchmark. Human evaluation can consider the details of video shots, including style, color, and other subtle details, while existing models struggle to capture these details.

\textbf{Closed-source Models} Among the closed-source models, GPT-5.2 performs the best with an average accuracy of 26.9\%, showing strong performance in most categories, especially in \textbf{Temporal} and \textbf{Style}. However, its performance in \textbf{Color} is relatively low. Gemini-3-Pro follows closely with an average accuracy of 24.4\%, excelling in \textbf{Temporal} and \textbf{Audio}, but lagging behind in \textbf{Color}. GPT-5-mini and Claude-4.0-Sonnet show lower overall performance, indicating that smaller or less specialized models struggle more in this benchmark.

\textbf{Open-source Models} The Qwen3 series demonstrates competitive performance, with Qwen3-VL-235B-A22B scoring an average of 20.6. While it slightly outperforms Gemini-3-Pro in \textbf{Color}, it does not surpass closed-source models in \textbf{Temporal} and \textbf{Style}. Qwen3-Omni-30B-A3B shows similar results, with an average accuracy of 20.0\%, performing strongly in \textbf{Temporal} and \textbf{Audio}.

\textbf{Performance Trends} \textbf{Temporal} is the best-performing category across all tasks. \textbf{Color} is generally lower, but the gap between closed-source and open-source models in these categories is not significant. \textbf{Style} shows a larger split, with models like GPT-5.2 and Gemini-3-Pro having significantly stronger capabilities.

\textbf{Model Scalability} Larger models do not necessarily lead to better performance across all categories. Qwen3-Omni-30B-A3B performs close to GPT-5.2 on \textbf{Temporal}. Both GPT-5.2 and Gemini-3-Pro show stable performance with their larger architectures, but smaller open-source models still deliver reasonable results.

\subsection{Further Analysis}
% 在本部分，我们随机抽取了一部分数据进行进一步分析，具体数据信息见图。模型使用的是Gemini 3 Pro。
In this section, we randomly selected a subset of the data for further analysis. Detailed further analysis data information can be found in Appendix \ref{xiaorong}.

\textbf{Effectiveness of the number of search queries.} As shown in Figure \ref{fig:4}A, increasing the number of queries (from $1{\times}2$ to $3{\times}2$) significantly improves the model's performance, especially in \textbf{Shot} and \textbf{Style}. This indicates that increasing the number of queries helps bridge the granularity gap between shot-level descriptions and video-level metadata. Although the performance improvement in \textbf{Color} is relatively small, \textbf{Temporal} and \textbf{Style} show clear benefits with an increase in the number of queries, suggesting that additional queries can better capture complex temporal and style features.

% 如图 \ref{fig:4}A 所示，随着查询数量的增加（从 $1{\times}2$ 到 $3{\times}2$），模型性能显著提升，尤其在 \textit{镜头} 和 \textit{风格} 上，表明增加查询数量可以弥补镜头级描述与视频元数据之间的粒度差距，提升检索召回率。尽管 \textit{色彩} 的性能提升较小，但 \textit{时序} 和 \textit{风格} 在查询数量增加后受益明显，表明额外的查询能够更好地捕捉复杂的时序和风格特征。因此，多查询策略有效地增强了模型对复杂描述条件的处理能力，尤其是在多重约束的检索任务中。

\textbf{Effectiveness of the number of downloaded videos.} As shown in Figure \ref{fig:4}B, increasing the number of videos downloaded per query (from $2{\times}1$ to $2{\times}3$) improves the model's overall performance, particularly in \textbf{Shot} and \textbf{Audio}. This suggests that more candidate videos provide additional matching information, especially for features that are difficult to obtain directly through simple descriptions. However, the improvement in \textbf{Resolution} is relatively slow, indicating that its improvement is related to the inherent quality of the video, rather than the number of retrieved candidates.
% This suggests that increasing the number of retrieved videos has a larger impact on certain features (such as \textbf{Audio} and \textbf{Shot}), but a smaller impact on others (such as \textbf{Resolution}).

% 如图 \ref{fig:4}B 所示，增加每查询下载的视频数量（从 $2{\times}1$ 到 $2{\times}3$）提升了模型的整体性能，尤其是在 \textit{镜头} 和 \textit{声音} 上。这表明更多的候选视频有助于提供更多的匹配信息，尤其对于这些难以通过简单描述直接获取的特征。然而，\textit{分辨率} 的提升较为缓慢，表明它的改进与视频质量本身和编码相关，而非检索候选数量。这表明，增加检索数量对一些特征（如声音和镜头）影响较大，但对其他特征（如分辨率）影响较小。

% \textbf{Effectiveness of the number of frames extracted.} As shown in Figure~\ref{fig:4}C, increasing the number of extracted frames consistently improves overall performance. The gains are more pronounced on \textbf{Shot} and \textbf{Temporal}, indicating that denser temporal sampling helps the model better capture shot boundaries and ordering. In contrast, improvements on \textbf{Color} and \textbf{Style} remain limited, suggesting that increasing frame density alone is insufficient to resolve fine-grained visual attributes.

\textbf{Effectiveness of video imagination.} We compare \textit{NoImagine}, which directly generates search keywords from the input description, with \textit{Imagine}, which first infers the likely full-video context. As shown in Figure~\ref{tab:imagination_ablation}, Imagine improves average accuracy by 10.7 points, with gains across all categories, especially \textbf{Resolution} and \textbf{Audio} (+12.5 each). This improvement suggests that video imagination helps bridge the gap between fine-grained shot descriptions and video-level metadata. The case study is detailed in Appendix \ref{case}.

\textbf{Sensitivity to Search Engine Choice.}
We compare Google Search, DuckDuckGo, and YouTube Search under identical settings. As shown in Figure~\ref{fig:6}, overall performance remains comparable across engines (20.7--25.0 average). Google Search performs best overall, while DuckDuckGo shows stronger results on \textbf{Temporal} and \textbf{Color} but weaker results on \textbf{Shot} and \textbf{Audio}. We therefore use Google Search as the default retriever.

% \textbf{Oracle Retrieval Analysis.} To isolate retrieval errors, we conduct an oracle experiment where the correct video is directly provided to the Localizer. As shown in Table~\ref{tab:oracle}, average accuracy increases from 25.0 to 48.1 (+23.1), confirming retrieval as a major bottleneck. The largest gains appear in \textbf{Color} and \textbf{Style} (+25.0 each), suggesting that such constraints are difficult to satisfy with keyword-based web search. However, oracle performance remains far below human level (88.5), showing that temporal localization itself is still challenging.

\textbf{Reproducibility Analysis.} We run the full pipeline three times under identical settings and report the mean, standard deviation, and variance in Table~\ref{tab:stats_columns}. All metrics show low standard deviation (0--2.67), with an average variance of only 1.14, indicating stable and reproducible results despite the use of closed-source models and external APIs.

\textbf{Prompt Sensitivity Analysis.}
We test three semantically equivalent evaluator prompts with unchanged criteria. As shown in Table~\ref{tab:case_results}, the average variance is only 0.11, with five of six categories showing zero variance and only minor fluctuation in \textbf{Color}. This indicates that the evaluation is robust to prompt wording.

\label{keyframe}
\textbf{Keyframe vs. Shot-level Evaluation.} To assess whether keyframe-based evaluation biases results, we compare it with full-shot human judgment on 50 random instances. As shown in Table~\ref{tab:keyframe_shot}, the two protocols reach 98\% agreement. The few disagreements mainly involve shots with substantial semantic variation, suggesting that keyframes are a practical and reliable proxy for large-scale evaluation.

\textbf{Latency Breakdown} In Table~\ref{tab:keyframe_shot}, we report the average per-sample latency of each pipeline stage with a parallelism level of 5. The full pipeline takes about 6 minutes, dominated by downloading (40.4\%), followed by search (29.3\%) and temporal grounding (21.1\%). This suggests an I/O-bound bottleneck, which could be reduced through higher parallelism, streaming extraction, and caching.

% \label{human}
% \textbf{Model evaluation vs. human evaluation.} As shown in Figure \ref{fig:confusion}, we present the confusion matrices between CLIP evaluation and human evaluation, as well as between MLLM evaluation and human evaluation. From the figure, it is evident that there is a significant discrepancy between CLIP evaluation and human evaluation, while MLLM evaluation aligns closely with human evaluation in most cases. This indicates that, in this experiment, the CLIP evaluation method is not suitable, whereas MLLM's judgments demonstrate strong reliability and can effectively replace human evaluation. We provide a case study in Appendix \ref{case1}.

% 如图所示，我们展示了模型预测与人工评估之间的混淆矩阵。从图中可以看出，模型预测与人工评估在大多数情况下高度一致，表明模型判断有较强的可靠性，能够替代人工评估。

\textbf{Failure cases.} In this study, we analyzed common pipeline errors. Generator errors occur when the MLLM misinterprets input, causing incorrect search expansions and irrelevant video retrieval. Partial search expansions can also omit crucial information, leading to irrelevant or partially relevant videos and inaccurate shot localization. Retriever errors happen when the MLLM generates accurate search expansions, but the retrieval stage either fails to find valid videos or retrieves irrelevant ones, leading to incorrect frame extraction. Localizer errors arise when, despite accurate search expansions and relevant video retrieval, issues with frame sampling lead to incorrect frames being extracted, or when the model mistakenly issues a judgment that contradicts human annotations. A more detailed analysis can be found in Appendix \ref{case2}.

\section{Conclusion}

%在本文中，我们提出了 ShotFinder，一个面向真实视频剪辑场景的开放域视频镜头检索与时间定位基准。ShotFinder 基于镜头描述以及五个具有实际意义的约束维度，对多模态模型进行评估。通过大规模的模型评测，我们发现了现有模型之间以及模型与人类之间显著的能力差距，并揭示了当前多模态大语言模型（MLLMs）在视频镜头检索任务中面临的挑战及其潜在的优化方向。

In this paper, we propose ShotFinder, a benchmark for open-domain video shot retrieval in realistic video editing scenarios and an imagination-driven retrieval method. ShotFinder benchmark evaluates MLLMs based on shot descriptions and five practically meaningful constraint dimensions. Through MLLM evaluations, we identify significant capability gaps both among existing models and between models and humans, and reveal the challenges faced by current MLLMs in video shot retrieval tasks as well as their optimization directions.

\section*{Limitations}

%视频镜头是视频中的一段连续画面，从成本和复杂性的角度考虑，我们用视频镜头的关键画面代替了视频镜头，我们之后会探究更好的处理方法。我们当前只讨论了单因素对视频剪辑的影响，多因素故事性剪辑属于更高阶、更接近真实制作的综合约束问题，我们之后会探究合理的解决方法。Query生成我们选择使用了单轮交互，未来我们将往多轮交互扩展。从搜索引擎获得视频后我们只进行了可用性的过滤，之后我们会尝试根据网页的html信息进行更深度的过滤，抽帧算法我们选用了基于视频时长的自适应抽帧策略，但是没有考虑到视频内容的影响，我们之后会去研究更合适的抽帧算法。

This study has several limitations. (1) In this study, we used key frames as a substitute for the full video shots, considering the cost and complexity of working with complete video clips. Future work will explore better methods for handling entire video shots. (2) Currently, we have focused on the impact of single-factor constraints on video editing. Multi-factor narrative editing, which is a more complex and realistic challenge, requires addressing combined constraints in a holistic manner, and we plan to explore solutions for this in future research. (3) For query generation, we employed single-turn interactions, but we aim to expand to multi-turn interactions in future iterations. After retrieving videos from search engines, we applied basic usability filtering, but we intend to implement more refined filtering based on HTML metadata from webpages for deeper content assessment. (4) Due to cost constraints and task complexity, we only evaluate a limited set of state-of-the-art models with relatively strong capabilities. As more powerful models become available in the future, we plan to expand the range of evaluated models accordingly.

\section*{Acknowledgments}

Thank these students for their contributions during the data quality revision phase: Shaoxiong Cheng, Ruilin Huang, Shuo Li, Yuxi Niu, Xinyuan Zhang, Yueya Xu, Jie Mao, Ruixuan Ji, Yaru Zhao, Mingchen Zhang. Thank these students for their discussions during the idea proposal stage: Yanpei Gong, Yuancheng Liu, Yiming Ding, Kangwei Zeng, Pengfei Yang, Zhongtian Luo, Jiaqi Liu.

This work was jointly supported by National Natural Science Foundation of China (62322607, 62236010 and 62276261), Beijing Natural Science Foundation (L252033).

\section*{Ethical Considerations}
% 使用的YouTube上的一些爬取数据，需要添加ethical considerations

In this study, we only crawl publicly available, non-infringing video content from the YouTube platform and adhere to the platform's terms of service and privacy policy. All data is used solely for academic research, with strict measures to ensure anonymity and privacy protection. The models used in this study are accessed via API calls, ensuring compliance with the terms of service set by the respective API providers. The models themselves were neither distributed nor modified in any way, ensuring that no intellectual property rights were violated. The benchmark is intended solely for academic research purposes. We do not redistribute raw YouTube videos. Any future use of the benchmark should comply with original YouTube Terms of Service and applicable copyright regulations.

\bibliography{custom}

\appendix

\newpage
\EnableTOC
\clearpage
\appendix

\section*{Appendix}
\begingroup
\setcounter{tocdepth}{3}  % 0=只显示section, 1=到subsection, 2=到subsubsection
\tableofcontents
\endgroup

\section{Theme Classification}
\label{theme_classification}

\paragraph{Temporal} To strictly define the ratio of Pre-context, we selected categories such as Knowledge, Fitness, and Food, as these topics inherently have a strong process-oriented nature. As described by \citep{zhou2017automaticlearningproceduresweb} in their research on learning steps through instructional videos, such content relies on initial “ingredients” or “instructions” to build subsequent temporal logic; viewers must understand the prerequisites to grasp the core content. For Post-context categories, we chose Sports, Automotive, and Gaming. Based on \citep{Giancola_2018} research on action recognition in sports videos, the semantic focus of such content often shifts to “event outcomes” (e.g., goals, win/loss status, race rankings), making their temporal grounding dependent on final discrete results rather than initial states. Movies and TV Series are isolated in the Pre-context and Post-context category because they represent complex narrative structures, requiring both setup and resolution, and this bidirectional temporal dependency aligns with the narrative coherence required in narratology. For data distribution, the ratio of Pre-context, Post-context, and both is 75:75:50, with topics in each category distributed equally.

\paragraph{Color} To create clear color boundaries, we applied strategic sampling: we mapped Parenting and Fashion to warm tones, utilizing the family warmth in Parenting content and the warm lighting in Fashion videos, which statistically form a significant contrast with cool-toned backgrounds; Tech and Art were assigned to cool tones because technology and digital art images frequently use cool hues (blue/cyan), high-saturation composite colors, and metallic finishes, enhancing the semantics of “precision,” “technological maturity,” and “futurism”—cool tones create a perception of technology and objectivity visually \citep{coursaris2008empirical}, making them widely used in product illustrations and digital art; Finally, we used Documentary and Travel as neutral categories, retaining natural light distribution from the real world and avoiding overly stylized color grading, serving as the true visual baseline for the dataset. For data distribution, the ratio of warm tones, cool tones, and neutral is 70:70:70, with topics in each category distributed equally.

\paragraph{Style} This choice achieves strict balanced distribution by assigning a single representative category to each quadrant. Variety Shows are the primary Live-action component, characterized by standard studio shooting techniques; Animation is explicitly defined as 2D Animation; Gaming represents 3D Animation/Rendering, as discussed by \citep{richter2016playingdatagroundtruth}, where modern game engines generate realistic images but differ fundamentally in texture mapping and lighting calculations from real cameras, helping the model distinguish between passive CGI and real-time rendering. Lastly, Music was chosen as the representative of Pure Graphics, particularly for lyric videos and dynamic typography, where high-density overlay text and dynamic graphics create unique visual features, ensuring the model can effectively recognize and process non-natural scene content dominated by text elements and user interfaces. For data distribution, the ratio of live-action, 2D animation, 3D animation, and pure graphics is 50:50:50:50, with topics in each category distributed equally.

\paragraph{Audio} To satisfy the 100:50:50 ratio, we prioritized Human Voice, selecting TV Series and Documentaries because these dialogue-driven genres provide the most continuous speech signals. This aligns with \citep{gemmeke2017audio}, whose AudioSet ontology identifies “Human Voice” as the core, essential explicit signal for video semantic understanding. Dance was selected to represent Music because its visual choreography is strictly synchronized with the audio rhythm, offering superior multimodal alignment compared to mixed audio types. Finally, Animals was chosen to represent Ambient Sound, as this content is mainly filmed in natural, outdoor environments, ensuring that the audio track is naturally dominated by biological calls and environmental textures (e.g., wind, water sounds), rather than human noise or dialogue. For data distribution, the ratio of human voice, music sound effects, and ambient sound is 100:50:50, with topics in each category distributed equally.

\paragraph{Resolution} We built a 100:100 (1:1) split between professionally produced content and user-generated content (UGC). Fashion and Movies serve as anchors at 1080p (High-Res), as these genres inherently prioritize extreme visual fidelity and complex color grading to provide an immersive viewing experience, providing the model with clean, artifact-free reference samples. In contrast, Life Vlogs and Fitness are mapped to 720p (Low/Variable Res) to capture “outdoor” authenticity; this aligns with \citep{Hosu_2020}, which emphasizes the necessity of learning from real distortion. These categories introduce typical mobile filming artifacts such as focus errors and compression artifacts, training the model to robustly handle the diverse quality spectrum found in real-world social media streams. For data distribution, the ratio of 1080p and 720p is 100:100, with topics in each category distributed equally.

% \section{Model evaluation vs. human evaluation}
% \label{human}

% As shown in Figure \ref{fig:confusion}, we present the confusion matrices between CLIP evaluation and human evaluation, as well as between MLLM evaluation and human evaluation. From the figure, it is evident that there is a significant discrepancy between CLIP evaluation and human evaluation, while MLLM evaluation aligns closely with human evaluation in most cases. This indicates that, in this experiment, the CLIP evaluation method is not suitable, whereas MLLM's judgments demonstrate strong reliability and can effectively replace human evaluation. We provide a case study in Appendix \ref{case1}. In our experiments, we set the CLIP similarity score threshold to 0.7.

\section{Model Evaluation vs. Human Evaluation}
\label{human}

\begin{figure*}[t]
 
\centering
\includegraphics[width=1\linewidth]{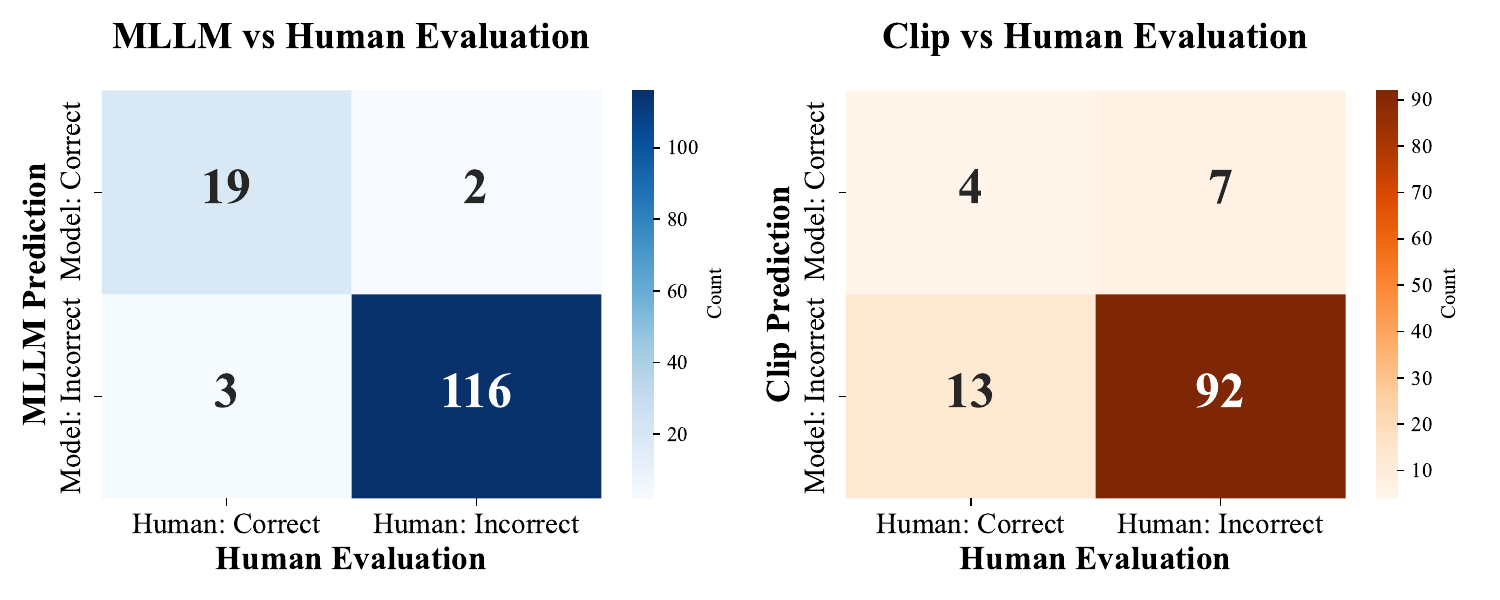}        %这个是在LaTeX文件夹中的相对路径
\caption{Confusion matrices between MLLM voting-based evaluation and human evaluation, and between CLIP evaluation and human evaluation}
\label{fig:confusion}
 
\end{figure*}

As shown in Figure \ref{fig:confusion}, we present the confusion matrices between CLIP evaluation and human evaluation, as well as between MLLM voting-based evaluation and human evaluation. From the figure, it is evident that there is a significant discrepancy between CLIP evaluation and human evaluation, whereas MLLM evaluation aligns closely with human evaluation in most cases. This indicates that, in this experiment, the CLIP evaluation method is not suitable, while the voting-based judgments of MLLMs demonstrate strong reliability and can effectively replace human evaluation. We provide a case study in Appendix \ref{case1}. In our experiments, we set the CLIP similarity score threshold to 0.7.

\section{Description Generation}
\label{data_shengcheng}

\subsection{Data Acquisition}

The benchmark dataset is collected from YouTube, covering six task categories designed to evaluate multimodal understanding capabilities in video editing scenarios. The data acquisition process varies based on task requirements, with different preprocessing and filtering strategies applied to different task types. \vspace{4pt} \\

\paragraph{Task Categories}
The dataset includes six task types: shot detection, temporal grounding, resolution classification, color classification, style classification, and audio description. Each task targets a specific aspect of video editing tasks.\vspace{4pt} \\

\paragraph{Standard Tasks (Direct Collection)}
For shot detection, temporal grounding, and resolution classification tasks, video segments are directly collected through keyword-based YouTube search and undergo standard preprocessing without additional filtering requirements. These tasks focus on structural and technical attributes that can be reliably extracted through automated processing. \vspace{4pt} \\

\paragraph{Filtered Tasks (Dual-Model Strategy)} 
For color classification, style classification, and audio description tasks, video segments require specific characteristics that cannot be directly obtained through metadata or keyword search. Since these attributes are intrinsically linked to video content rather than searchable metadata, we employ automated filtering strategies to ensure data quality. \vspace{4pt} \\

\paragraph{Dual-Model Filtering for Color and Style} 
For color and style classification tasks, we implement a parallel dual-model filtering mechanism using both Qwen-2.5-VL-72B and Gemini-3-Pro to ensure annotation consistency and reduce model bias. Both models independently analyze the same video frames, and they must reach agreement on the classification results meeting the requirements for them to be considered valid. The system automatically retries up to 3 times if either model returns unknown or ambiguous results. Only frames where both models agree on the same category are retained in the dataset. This consensus-based approach significantly improves filtering accuracy and ensures consistent annotation quality. \vspace{4pt} \\

\paragraph{Automated Audio Filtering} 
For the audio description task, we use Gemini-3-Pro with an automated audio quality assessment system to filter video segments. The system analyzes whether the audio meets the requirements of our task, and only segments that meet the criteria are retained for annotation.\vspace{4pt} \\

\paragraph{Preprocessing Pipeline} 
After passing the filtering stage (for color, style, and audio tasks) or direct collection (for shot, temporal, and resolution tasks), all video segments undergo unified preprocessing. For each video, we randomly select a target timestamp and extract a 10-second window centered on that moment. The preprocessing extracts the following multimodal assets: \vspace{2pt} \\

\paragraph{Multimodal Asset Structure}
The extracted assets for each video segment include: 
\begin{itemize}
    \item \textbf{Target Frame:} A single keyframe (JPG format) representing the moment to be described.
    \item \textbf{Preceding Frames:} Three frames sampled at 1-second intervals before the target frame, providing temporal context of what happened before. 
    \item \textbf{Following Frames:} Three frames sampled at 1-second intervals after the target frame, capturing the subsequent narrative development.
    \item \textbf{Audio Segment:} An MP3 file extracted from the 10-second window centered on the target timestamp.
    \item \textbf{Metadata:} A JSON file recording the target timestamp, video duration, resolution, and file paths of all extracted assets.
\end{itemize}

\paragraph{Frame Similarity Filtering}
To ensure temporal diversity and avoid redundant frames, we employ a CLIP-based similarity filtering mechanism. For each 7-frame sequence (target frame + 3 preceding + 3 following), we compute cosine similarity between all 9 pairs of adjacent frames using CLIP embeddings. If all pairwise similarities exceed 0.8, indicating insufficient temporal variation, the entire sequence is discarded and a new target timestamp is randomly selected. This process repeats until a sequence with adequate temporal diversity is obtained. This strategy ensures that the extracted frames capture meaningful temporal changes rather than static scenes.

\subsection{Description Generation Prompts}

\subsubsection{Filtering Prompts}

\tcbset{
    breakable,
    colframe=blue!5!black,
    colback=gray!10!white,
    fonttitle=\bfseries,
    width=\columnwidth
}

\begin{tcolorbox}[
    title=\textbf{Color Filtering Prompt},
    fonttitle=\bfseries
]

{\color{blue}\textbf{Input:}} Target keyframe (JPG image) \vspace{2pt} \\

\textbf{System Prompt:} You are a visual analysis expert specializing in color tone classification. \vspace{2pt} \\

\textbf{Task:} Analyze the provided image and classify its overall color tone into one of three categories. \vspace{2pt} \\

\textbf{Categories:} \vspace{1pt} \\
\textbf{Warm Tone:} The image is dominated by warm colors such as orange, yellow, or red.\vspace{1pt} \\
\textbf{Cold Tone:} The image is dominated by cold colors such as blue, cyan, or purple. \vspace{1pt} \\
\textbf{Neutral Tone:} The image has balanced white balance with no obvious color bias. \vspace{2pt} \\

\textbf{Output:} Return only one word indicating the classification result.

\end{tcolorbox}

\vspace{6pt}

\begin{tcolorbox}[
    title=\textbf{Style Filtering Prompt},
    fonttitle=\bfseries
]

{\color{blue}\textbf{Input:}} Target keyframe (JPG image) \vspace{2pt} \\

\textbf{System Prompt:} You are a visual analysis expert specializing in visual style classification. \vspace{2pt} \\

\textbf{Task:} Analyze the provided image and classify its visual style into one of four categories. \vspace{2pt} \\

\textbf{Categories:} \vspace{1pt} \\
\textbf{Real:} Live-action or cinematic footage, content based on reality. \vspace{1pt} \\
\textbf{2D Animation:} Flat animation including anime, cartoons, or hand-drawn content. \vspace{1pt} \\
\textbf{3D Animation:} 3D animation, game graphics, or rendered scenes. \vspace{1pt} \\
\textbf{Graphic:} Pure graphics such as charts, text, or software interfaces. \vspace{2pt} \\

\textbf{Output:} Return only one option indicating the classification result.

\end{tcolorbox}

\vspace{6pt}

\begin{tcolorbox}[
    title=\textbf{Audio Filtering Prompt},
    fonttitle=\bfseries
]

{\color{blue}\textbf{Input:}} Audio segment (MP3 file) \vspace{2pt} \\

\textbf{System Prompt:} You are an expert in audio description classification. \vspace{2pt} \\

\textbf{Task:} Analyze the provided audio clip and classify its overall audio description into one of the three categories. \vspace{2pt} \\

\textbf{Categories:} \vspace{1pt} \\
\textbf{Human Voice}: The audio clip primarily contains human speech or dialogue. \vspace{1pt} \\
\textbf{Background Music}: The audio clip primarily contains music or background tracks, with no significant human voice. \vspace{1pt} \\
\textbf{Ambient Sound}: The audio clip primarily contains natural or environmental sounds, such as wind, water, animal sounds, etc. \vspace{2pt} \\

\textbf{Output:} Return only one word indicating the classification result.

\end{tcolorbox}

\subsubsection{Frame Description Prompt}

\begin{tcolorbox}[
    title=\textbf{Multimodal Description Generation Prompt},
    fonttitle=\bfseries
]

{\color{blue}\textbf{Input:}} Target keyframe (JPG image), preceding frames (3 images), following frames (3 images), and audio segment (MP3) \vspace{2pt} \\

\textbf{System Prompt:} You are a professional video content analyst. Generate comprehensive multimodal descriptions for video segments. \vspace{2pt} \\

\textbf{Task Requirements:} \vspace{1pt} \\

\textbf{Main Frame Description (240-260 characters):} Provide a detailed description of the key frame, including subjects, actions, props, text overlays, lighting characteristics, and camera language. The description should be rich in visual details and maintain the specified character count. \vspace{2pt} \\

\textbf{Temporal Context:} \vspace{1pt} \\
\textbf{Preceding Seconds:} Describe what happens in the few seconds before the key frame, focusing on actions and scene transitions. \vspace{1pt} \\
\textbf{Following Seconds:} Describe what happens in the few seconds after the key frame, emphasizing narrative continuity. \vspace{2pt} \\

\textbf{Color Analysis:} Analyze the overall color tendency, color intensity, and lighting characteristics of the frame. Describe the dominant color palette and its emotional impact. \vspace{2pt} \\

\textbf{Style Analysis:} Classify the visual style (real, 2D animation, 3D animation, or graphic) and describe its characteristic features, such as rendering techniques, artistic style, or production quality. \vspace{2pt} \\

\textbf{Audio Description:} Analyze the audio track, including speech content, sound events, background music, and ambient audio. Describe the audio atmosphere and its relationship to the visual content. \vspace{2pt} \\

\textbf{Output Format:} Generate descriptions in natural language for each component listed above.

\end{tcolorbox}

\subsubsection{JSON Validation Prompt}

\begin{tcolorbox}[
    title=\textbf{Data Validation and Formatting Prompt},
    fonttitle=\bfseries
]

{\color{blue}\textbf{Input:}} Raw multimodal descriptions (text format) \vspace{2pt} \\

\textbf{System Prompt:} You are a professional data annotator. Organize the input raw video analysis data into strict JSON format. \vspace{2pt} \\

\textbf{Task Requirements:} \vspace{1pt} \\

\textbf{Deduplication and Cleaning:} Remove descriptions of non-content elements such as watermarks, progress bars, and corner icons. \vspace{2pt} \\

\textbf{Privacy Compliance:} Remove specific real-world names and locations (unless they are well-known landmarks), replacing them with generic descriptions. \vspace{2pt} \\

\textbf{Bilingual Output:} All description fields must generate both Chinese (\_ch) and English (\_en) versions. \vspace{2pt} \\

\textbf{Character Count Requirement (Critical):} The segment\_description\_ch must be maintained at approximately 240-260 characters. The detailed content from raw\_main must be fully preserved, with only privacy cleaning applied, without reducing word count. \vspace{2pt} \\

\textbf{Context Description Format (Important):} The context\_description\_ch and context\_description\_en must be arrays containing two elements in the format: [description of preceding seconds, description of following seconds]. \vspace{2pt} \\

\textbf{Format Requirements (Critical):} Output only pure JSON objects without any Markdown markers. Do not add any explanatory text. Ensure all strings are properly escaped. \vspace{2pt} \\

\textbf{Output JSON Schema:} \vspace{1pt} \\
\textbraceleft \vspace{1pt} \\
\hspace{1em}id: youtube\_\textless video\_id\textgreater, \vspace{1pt} \\
\hspace{1em}video\_link: https://..., \vspace{1pt} \\
\hspace{1em}video\_source: YouTube, \vspace{1pt} \\
\hspace{1em}category: ..., \vspace{1pt} \\
\hspace{1em}timestamp: YYYY-MM-DD HH:MM:SS, \vspace{1pt} \\
\hspace{1em}resolution: 1080P/720P, \vspace{1pt} \\
\hspace{1em}segment\_description\_ch: ..., \vspace{1pt} \\
\hspace{1em}segment\_description\_en: ..., \vspace{1pt} \\
\hspace{1em}context\_description\_ch: [before..., after...], \vspace{1pt} \\
\hspace{1em}context\_description\_en: [before..., after...], \vspace{1pt} \\
\hspace{1em}color\_description\_ch: ..., \vspace{1pt} \\
\hspace{1em}color\_description\_en: ..., \vspace{1pt} \\
\hspace{1em}style\_description\_ch: ..., \vspace{1pt} \\
\hspace{1em}style\_description\_en: ..., \vspace{1pt} \\
\hspace{1em}audio\_description\_ch: ..., \vspace{1pt} \\
\hspace{1em}audio\_description\_en: ... \vspace{1pt} \\
\textbraceright

\end{tcolorbox}

\section{Evaluation}
\label{pingce}

\tcbset{
    breakable,
    colframe=blue!5!black,
    colback=gray!10!white,
    fonttitle=\bfseries,
    width=\columnwidth % 使tcolorbox适应单栏宽度
}

\begin{tcolorbox}[
    title=\textbf{Evaluation Prompt},
    fonttitle=\bfseries
]

\textbf{Role:} You are a Visual Frame Grounding Specialist. Your task is to identify the single most relevant frame that best matches the user's description. \vspace{2pt} \\

\textbf{Context:} \vspace{1pt} \
The input consists of multiple images extracted from a video. \
The images are provided in strict chronological order, from the beginning to the end of the video. \
Each image is a representative frame of one video shot, where the frame is sampled approximately from the temporal middle of that shot. \
The first image has frame\_id = 0, the second image has frame\_id = 1, and so on. \
In addition to images, you are also given the full audio track of the video. \
The audio is not segmented per frame or shot and should be used as global temporal and semantic context, not as a direct frame-to-audio alignment. \vspace{2pt} \\

\textbf{Shot Metadata:} \vspace{1pt} \
For each frame, the corresponding shot's temporal boundaries are provided as start\_time and end\_time. \
Each frame\_id identifies a representative frame of one shot, and the corresponding start\_time and end\_time specify the temporal extent of that shot. \vspace{2pt} \\

\textbf{Shot Metadata format:} \

frame\_id = 0, start\_time = <seconds>, end\_time = <seconds> \

frame\_id = 1, start\_time = <seconds>, end\_time = <seconds> \

... \vspace{2pt} \\

\textbf{Target Description (The event to find):} \textquotedbl\$en\_memory\_data\$\textquotedbl \vspace{2pt} \\

\textbf{Instructions:} \vspace{1pt} \

\textbf{1. Joint Reasoning with Vision, Audio, and Shot Metadata:} \
Use the text description, visual content of all frames, the audio track, and the Shot Metadata together. \
Use audio cues (e.g., speech, sound events, changes in intensity) to infer when the described event occurs. \
Use the start\_time and end\_time of each shot to determine which shot covers or is closest to the inferred event time. \
Use visual evidence from the representative frames to determine which frame best represents the described event. \vspace{2pt} \\

\textbf{2. Scan All Frames:} Analyze all provided images in chronological order. \
Consider both the visual-semantic similarity between the target description and each frame and the temporal consistency between the inferred event time and the corresponding shot's start\_time and end\_time. \vspace{2pt} \\

\textbf{3. Select the Best Match:} Identify the single frame whose visual content is most relevant to the target description. \
If the target event occurs within a shot's temporal interval, prefer the representative frame of that shot. \
If multiple shots partially match, choose the frame from the shot that provides the strongest combination of visual relevance, semantic relevance, and temporal proximity to the inferred event. \
If the event occurs near a boundary between two shots, compare the neighboring shots and select the most relevant frame. \
If the audio provides weak or ambiguous temporal information, select the most semantically and visually related frame. \vspace{2pt} \\

\textbf{4. Important Temporal Reasoning Rule:} \
Do not assume that frame\_id corresponds to a uniformly sampled timestamp. \
Each frame\_id identifies the representative frame of a particular shot. \
Use the provided start\_time and end\_time to reason about the temporal location of each shot. \
The approximate representative time of a shot may be considered as 
\textbf{t\textsubscript{i} = (start\_time\textsubscript{i} + end\_time\textsubscript{i}) / 2}. \
However, the actual visual evidence should take precedence over this approximate representative time when selecting the best match. \vspace{2pt} \\

\textbf{5. Mandatory Selection Rule:} \
You \textbf{must return exactly one valid frame\_id}. The frame\_id \textbf{must be an integer between 0 and \$NUM\_FRAMES\$ - 1}. Do \textbf{not} return \textquotedbl N/A\textquotedbl, null, or multiple values. \vspace{2pt} \\

\textbf{Output Format (JSON Only):} \vspace{1pt} \\
\textless tool\_call\textgreater\textbraceleft\textquotedbl frame\_id:\textless integer\textgreater\textbraceright\textless/tool\_call\textgreater

\end{tcolorbox}

\section{Imagination}
\label{imagination}

\tcbset{
    breakable,
    colframe=blue!5!black,
    colback=gray!10!white,
    fonttitle=\bfseries,
    width=\columnwidth % 使tcolorbox适应单栏宽度
}

\begin{tcolorbox}[
    title=\textbf{Imagination Prompt},
    fonttitle=\bfseries
]

You are an autonomous \textquotedbl Video Hunter Agent\textquotedbl. Your ultimate goal is to precisely locate the original video link based on the user's vague memory (including visual, auditory, and timeline clues). You must provide \textbf{2} keywords most likely to hit the target container. \vspace{2pt} \\

\textbf{Tool Available:} \vspace{1pt} \\
\texttt{search\_videos}: Simultaneously performs search on YouTube. \\
Parameters: \texttt{\textbraceleft\textquotedbl query\textquotedbl: \textquotedbl search keywords\textquotedbl\textbraceright} \\
Note: This tool automatically handles the \textquotedbl youtube\textquotedbl\ suffix, so you only need to provide the core keywords. \vspace{2pt} \\

\textbf{Output Format:} \vspace{1pt} \\
\textless think\textgreater...your deep reasoning process...\textless/think\textgreater \\
\textless tool\_call\textgreater\textbraceleft\textquotedbl name\textquotedbl: \textquotedbl search\_videos\textquotedbl, \textquotedbl arguments\textquotedbl: \textbraceleft\textquotedbl query\textquotedbl: [\textquotedbl your search term1\textquotedbl, \textquotedbl your search term2\textquotedbl]\textbraceright\textbraceright\textless/tool\_call\textgreater \vspace{2pt} \\

\textbf{Core Constraints \& Strategy:} \vspace{2pt} \\

\textbf{1. Reasoning First:} Before taking any action, you must think inside \textless think\textgreater...\textless/think\textgreater. You must not directly call tools or output answers without prior reasoning. \vspace{2pt} \\

\textbf{2. Keyword Generation Strategy (Abductive Reasoning):} You must use \textquotedbl Abductive Reasoning\textquotedbl\ to bridge the gap between the user's \textquotedbl clip description\textquotedbl\ and the likely \textquotedbl full video title.\textquotedbl \vspace{2pt} \\

\textbf{Deconstruction:} Extract visual anchors (OCR, specific objects), actions, and atmosphere. \vspace{2pt} \\

\textbf{Divergent Hypotheses (\textquotedbl Container\textquotedbl\ Logic):} You must propose different paths for what the full video actually is: \vspace{2pt} \\

\textbf{Hypothesis A (Literal):} The video is explicitly about the action described (e.g., a \textquotedbl How-to\textquotedbl\ or generic vlog). \vspace{2pt} \\

\textbf{Hypothesis B (Metaphorical/B-roll):} The clip is a visual metaphor for an abstract concept (e.g., \textquotedbl cutting bread with a spoon\textquotedbl\ $\rightarrow$ Search keywords: \textquotedbl inefficiency\textquotedbl, \textquotedbl wrong tools\textquotedbl, \textquotedbl productivity\textquotedbl). \vspace{2pt} \\

\textbf{Hypothesis C (Specific Source):} The clip is from a specific movie, meme, game glitch, or news event (e.g., \textquotedbl bucket on shopkeeper head\textquotedbl\ $\rightarrow$ Search keywords: \textquotedbl Skyrim bucket thief\textquotedbl). \vspace{2pt} \\

\textbf{Selection:} Select the most probable hypothesis to construct a \textbf{single best English search phrase}. \vspace{2pt} \\

\textbf{Information Source Isolation:} Reference ONLY English corpus. \vspace{2pt} \\

\textbf{3. Search Strategy:} You MUST perform search rounds using the best English keywords derived from your strongest hypothesis. \vspace{2pt} \\

\textbf{4. Execution Workflow:} Analyze the description using Abductive Reasoning \& construct the best English keyword $\rightarrow$ call \texttt{search\_videos} \vspace{2pt} \\

\textbf{Example:} \vspace{1pt} \\
\textbf{User:} \textquotedbl Split-screen dance video, Y2K fashion, leg warmers, electronic music\textquotedbl \vspace{1pt} \\
\textbf{Agent:} \textless think\textgreater Y2K fashion + split-screen editing + dance = likely TikTok trend or K-pop cover. Keywords: style + format.\textless/think\textgreater\textless tool\_call\textgreater\textbraceleft\textquotedbl name\textquotedbl: \textquotedbl search\_videos\textquotedbl, \textquotedbl arguments\textquotedbl: \textbraceleft\textquotedbl query\textquotedbl: [\textquotedbl Y2K dance split screen leg warmers\textquotedbl, \textquotedbl split screen clone dance phonk\textquotedbl]\textbraceright\textbraceright\textless/tool\_call\textgreater

\end{tcolorbox}

\section{Further Analysis}
\label{xiaorong}

%消融实验数据分布与主实验相同，对于镜头描述，20个主题每个主题随机抽取一个数据；对于时序描述，共24个数据，知识：健身：美食：体育：汽车：游戏：电影：电视剧 = 3:3:3:3:3:3:3:3，其中知识、健身、美食取 Pre-coontext；体育、汽车、游戏取 Post-context；电影、电视剧取 Both;对于色彩描述，共24个数据，育儿：时尚：科技：艺术：纪录片：旅行 = 4:4:4:4:4:4，其中育儿、时尚取暖色调；科技、艺术取冷色调；纪录片、旅行取中性；对于风格描述，共24个数据，综艺节目：动画：游戏：音乐 = 6:6:6:6，其中综艺节目取实拍；动画取平面动画；游戏取立体动画；音乐取纯图形；对于声音描述，电视剧：纪录片：舞蹈：动物 = 6:6:6:6，其中电视剧、纪录片取人声；舞蹈取音乐；动物取环境音；对于分辨率，共24个数据，时尚：电影：生活Vlog：健身 = 6:6:6:6，其中时尚、电影取1080P；生活vlog、健身取720P

\begin{figure*}[t]
 
\centering
\includegraphics[width=1\linewidth]{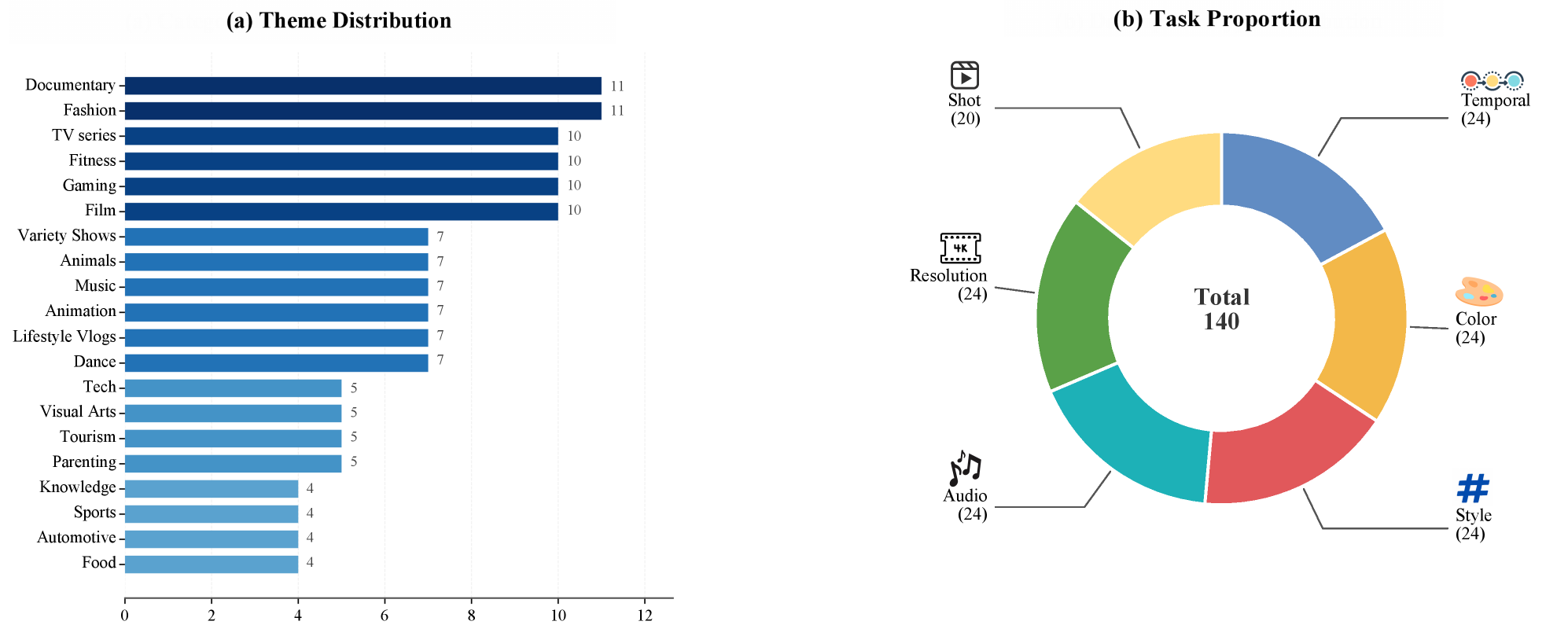}        %这个是在LaTeX文件夹中的相对路径
\caption{Theme distribution and task proportion of the dataset for further analysis.}
\label{shujv}
 
\end{figure*}

The data distribution in the ablation experiments is the same as in the main experiment, and the data distribution is shown in Figure \ref{shujv}. The model used for generating results is Gemini 3 Pro.

For \textbf{Shot}, one data sample is randomly selected from each of the 20 topics; for Temporal, 24 data samples are randomly selected with the following distribution: Knowledge, Fitness, Food, Sports, Cars, Games, Movies, and TV Shows, with 3 samples each. Specifically, Knowledge, Fitness, and Food use Pre-context; Sports, Cars, and Games use Post-context; Movies and TV Shows use both Pre-context and Post-context. For \textbf{Color}, 24 data samples are randomly selected, with the following distribution: Parenting, Fashion, Technology, Art, Documentaries, and Travel, with 4 samples each. Specifically, Parenting and Fashion use warm tones; Technology and Art use cool tones; Documentaries and Travel use neutral tones. For \textbf{Style}, 24 data samples are randomly selected, with the following distribution: Variety Shows, Animation, Games, and Music, with 6 samples each. Specifically, Variety Shows use live-action; Animation uses 2D animation; Games use 3D animation; Music uses pure graphics. For \textbf{Audio}, 24 data samples are randomly selected, with the following distribution: TV Shows, Documentaries, Dance, and Animals, with 6 samples each. Specifically, TV Shows and Documentaries use human voices; Dance uses music; Animals use ambient sounds. For \textbf{Resolution}, 24 data samples are randomly selected, with the following distribution: Fashion, Movies, Lifestyle Vlogs, and Fitness, with 6 samples each. Specifically, Fashion and Movies use 1080P; Lifestyle Vlogs and Fitness use 720P.

\label{sec:appendix}
\input{appendix}

\section{Annotation Guidelines}

The goal of this annotation task is to manually verify and refine model-generated shot descriptions to ensure that they accurately and faithfully reflect the corresponding video shots. Annotators are primarily responsible for identifying discrepancies between the generated descriptions and the actual video content, and correcting them when necessary.

1. Video–Description Alignment

Before making any edits, please carefully watch the corresponding video shot in full. Pay close attention to the main subjects, actions, scenes, and event progression, as well as relevant attributes such as color characteristics, visual style, audio content, and visual quality. Then, review the model-generated shot description and constraint descriptions to form an overall understanding of their alignment with the video.

2. Discrepancy Identification

Please assess whether the generated description contains any discrepancies with respect to the video content, considering the following aspects:

\begin{itemize}
\item Semantic consistency: Whether the description correctly captures the actions, events, and visual content shown in the video, without missing key elements or introducing nonexistent ones.

\item Factual accuracy: Whether entities such as people, objects, scenes, and temporal order match the video.

\item Constraint consistency: Whether constraints (e.g., temporal sequence, color, style, audio, and resolution) are consistent with what is actually observable in the video.
\end{itemize}

Any clear inconsistency in these aspects should be treated as a discrepancy.

3. Revision Principles

If discrepancies are found, please revise the description based strictly on the video content. Revisions should follow these principles:

\begin{itemize}
\item Accuracy first: All statements in the description must be directly supported by the video.

\item Minimal necessary changes: Make only the modifications required to correct errors, and avoid unnecessary rewriting.

\item Objectivity: Do not add speculative or inferred information that is not explicitly visible or audible in the video.

\item Clarity and precision: When correcting errors, improve clarity and precision where appropriate.
\end{itemize}

4. Handling Correct Descriptions

If the generated description is fully consistent with the video and contains no factual or constraint-related errors, it may be retained without modification.

5. Annotation Objective

By following this verification and refinement process, the final descriptions should accurately correspond to their video shots, providing a high-quality, low-noise, and reproducible dataset for benchmark evaluation and subsequent research.

\begin{figure*}[t]
 
\centering
\includegraphics[width=1\linewidth]{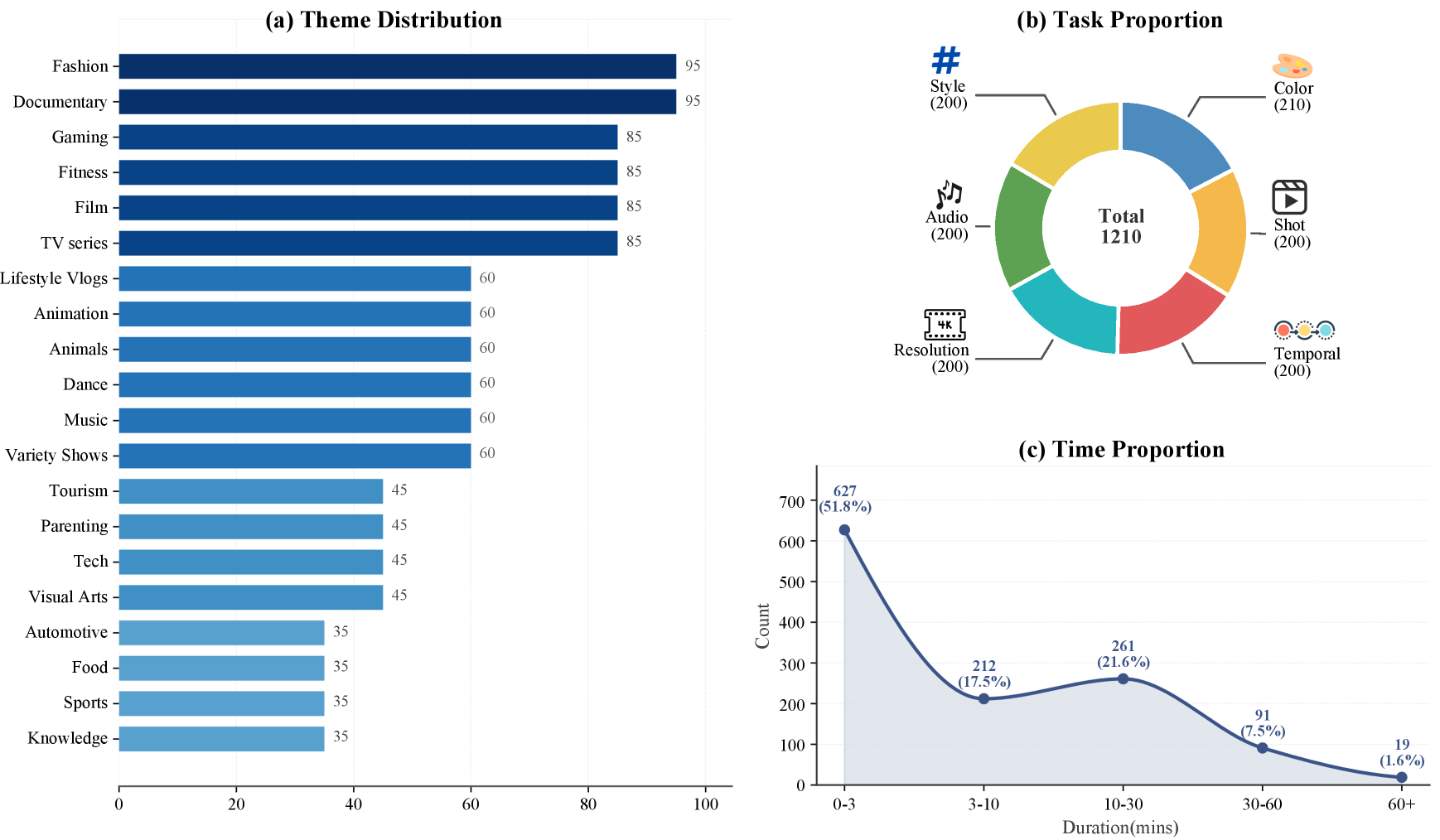}        %这个是在LaTeX文件夹中的相对路径
\caption{Theme distribution, task proportion, and time proportion of the dataset.}
\label{data}
 
\end{figure*}

% rebuttal 补充的实验的表格都放在下面了

% 第三个实验

% 我们认同，仅报告端到端结果确实不利于定位性能瓶颈。因此，我们对整个 pipeline 进行了组件级拆解评估，在更受控的设定下分别分析了检索质量与定位质量。具体而言，我们将系统划分为三个阶段：生成搜索词、候选视频检索以及镜头 grounding，并分别统计各阶段的独立表现。

% \begin{table*}[t]
% \centering
% {
% \begin{tabular}{lccccccc}
% \toprule
% \textbf{Stage} & \textbf{Shot} & \textbf{Temporal} & \textbf{Color} & \textbf{Style} & \textbf{Audio} & \textbf{Resolution} & \textbf{Average} \\
% \midrule
% query generation       & 100   & 100   & 100   & 100   & 100   & 100   & 100 \\
% candidate video retrieval     & 38.4  & 48.0  & 10.7  & 21.3  & 53.3  & 21.3  & 32.0 \\
% shot grounding        & 30.0  & 37.5  & 8.3   & 16.7  & 41.7  & 16.7  & 25.0 \\
% \bottomrule
% \end{tabular}}
% \caption{Results at different stages of the pipeline.}
% \label{tab:stage_results}
% \end{table*}

% 不同的prompt

\begin{table*}[t]
\centering
\begin{tabular}{lccccccc}
\toprule
\textbf{Case} & \textbf{Shot} & \textbf{Temporal} & \textbf{Color} & \textbf{Style} & \textbf{Audio} & \textbf{Resolution} & \textbf{Average} \\
\midrule
original & 30   & 37.5  & 8.3   & 16.7  & 41.7  & 16.7  & 25 \\
case1    & 30   & 37.5  & 4.2   & 16.7  & 41.7  & 16.7  & 24.3 \\
case2    & 30   & 37.5  & 8.3   & 16.7  & 41.7  & 16.7  & 25 \\
Mean     & 30   & 37.5  & 6.9   & 16.7  & 41.7  & 16.7  & 24.8 \\
Variance & 0    & 0     & 3.9   & 0     & 0     & 0     & 0.1 \\
\bottomrule
\end{tabular}
\caption{Experimental results for different cases, including mean and variance.}
\label{tab:case_results}
\end{table*}

% 第四个实验

% 仅在oracle设置（给定正确视频）下单独评估定位器，虽然可以分析模型的上限性能，但无法全面反映系统在真实场景中的整体表现。因此，我们在当前实验设置中更强调端到端性能，以验证方法在实际应用中的有效性。当然，我们认可在oracle设置下补充定位性能评估有助于更全面地分析模型能力。在下表S4我们给出了实验结果。

% \begin{table*}[t]
% \centering
% {
% \begin{tabular}{lccccccc}
% \toprule
% \textbf{Retriever} & \textbf{Shot} & \textbf{Temporal} & \textbf{Color} & \textbf{Style} & \textbf{Audio} & \textbf{Resolution} & \textbf{Average} \\
% \midrule
% oracle           & 90   & 91.7  & 83.3  & 75   & 87.5  & 79.2  & 84.3 \\
% google search    & 30   & 37.5  & 8.3   & 16.7 & 41.7  & 16.7  & 25 \\
% \bottomrule
% \end{tabular}}
% \caption{Experimental results for selected retrievers.}
% \label{tab:oracle}
% \end{table*}

% 第六个实验

% 感谢审稿人的重要意见。我们理解对于依赖闭源模型与外部工具的工作，确实可能引发可重复性方面的担忧。因此，我们从两个方面补充了实验与说明，以验证方法的稳定性与可复现性。
% 首先，我们在相同设置下独立运行了三次完整实验，并对主要评估指标计算了均值（Mean）、标准差（Std）和方差（Var），以衡量结果波动情况。统计结果如表S8所示：

% 这个能不能加到主实验中呢？

\begin{table*}[h]
\centering
\begin{tabular}{lccc}
\toprule
\textbf{Column} & \textbf{Mean} & \textbf{Std} & \textbf{Var} \\
\midrule
Avg        & 24.94 & 1.07 & 1.14 \\
Shot       & 30.42 & 1.38 & 1.91 \\
Temporal   & 37.85 & 2.67 & 7.11 \\
Color      & 7.29  & 1.80 & 3.25 \\
Style      & 16.67 & 0 & 0 \\
Audio      & 41.32 & 1.15 & 1.33 \\
Resolution & 17.01 & 1.15 & 1.33 \\
\bottomrule
\end{tabular}
\caption{Mean, standard deviation, and variance for each column, with values rounded to two decimal places.}
\label{tab:stats_columns}
\end{table*}

% 第七个实验
% 首先，从任务属性来看，ShotFinder 的目标是在开放互联网环境中检索并定位符合文本描述的镜头片段，这一应用场景更接近于内容创作辅助或素材检索，而非强实时交互系统（例如在线对话或自动驾驶决策）。因此，系统的核心优化目标是检索准确性与语义匹配质量，而非毫秒级响应时间。在实际视频编辑或内容制作流程中，数分钟级别的处理时间是可接受的。其次，我们在修订稿中补充了各阶段的耗时统计实验。结果为并行进程数设置为 5 时的平均耗时，具体如表S9所示：

% 我觉得可以放到正文

\begin{table*}[h]
\centering
\begin{tabular}{lcc}
\toprule
\textbf{Stage} & \textbf{Time} & \textbf{Proportion} \\
\midrule
Search (incl. Video Imagination) & 104.60 s & 29.33\% \\
Download                         & 144.20 s & 40.44\% \\
Frame Extraction                 & 35.10 s  & 9.84\%  \\
Grounding                        & 75.30 s  & 21.12\% \\
VLM Test                         & 17.40 s  & 4.88\%  \\
\textbf{Total}                   & 356.60 s (5.94 min) & 100\% \\
\bottomrule
\end{tabular}
\caption{Time consumption and proportion of each stage in the pipeline.}
\label{tab:time_proportion}
\end{table*}

% 第八个实验

% 我们理解，仅使用关键帧作为评估依据，可能会引发对视频级语义完整性的担忧。因此，我们专门补充了一个对照实验，对 50 条样本分别采用“关键帧评估”和“完整镜头评估”两种方式进行人工对比，以验证关键帧评估是否会引入系统性偏差。
% 在该实验中，我们以镜头级人工标注作为参考标准，并统计关键帧评估结果与镜头级评估结果之间的一致性。对应的混淆矩阵如下（第一项表示关键帧判断结果，第二项表示镜头级判断结果）：

% We profile the average per-sample latency of each pipeline stage in Table~\ref{tab:keyframe_shot}. The full pipeline takes about 6 minutes per sample, with Download as the main bottleneck (40.4\%), followed by Search (29.3\%) and Grounding (21.1\%). Frame Extraction and Evaluation are relatively lightweight. This indicates that latency is mainly I/O-bound due to video downloading, suggesting optimizations such as parallel downloading, streaming-based frame extraction, and caching.

\begin{table*}[h]
\centering
\begin{tabular}{lcc}
\toprule
 & \textbf{Shot: True} & \textbf{Shot: False} \\
\midrule
\textbf{Keyframe: True}  & 13 & 1  \\
\textbf{Keyframe: False} & 0  & 36 \\
\bottomrule
\end{tabular}
\caption{Confusion matrix for Shot detection based on Keyframes.}
\label{tab:keyframe_shot}
\end{table*}

\end{document}

%% file: appendix.tex
\section{Case study}
\label{app:case_study}
% 定义Case Study框样式

\subsection{Case study for verification.}
\label{case1}

\input{case_studies/case_4_a}

\label{case_4_a}

\input{case_studies/case_4_b}
\label{case_4_b}

\input{case_studies/case_4_c}
\label{case_4_c}

\input{case_studies/case_4_d}
\label{case_4_d}

\subsection{Case study for errors occured in pipeline}
\label{case2}

\subsubsection{Generator errors}
In this error category, the MLLM misinterprets the input fragment, which results in the generation of incorrect search expansions. This initial error leads the web search interface to retrieve an irrelevant candidate video set, further propagating the inaccuracy to the shot localization stage. Consequently, the entire pipeline operates on irrelevant videos, ultimately yielding an erroneous final output.
\input{case_studies/case5_query}

\label{case5_query}

In this type of error, the MLLM generates only partially precise search expansions, which cause the omission of crucial information from the input. This information loss leads the retrieval interface to return a candidate video set that is either partially relevant or entirely unrelated. Consequently, the system fails to perform accurate shot localization, making it impossible to extract the correct frames for final judgment.
\input{case_studies/case5_query2}

\label{case5_query2}

\subsubsection{Retriever errors}
In this type of error, the MLLM generates precise search expansions, the video retrieval stage fails to construct a valid candidate video set from the web. Due to the lack of accessible video links that match the search expansions on the search engine, the system ultimately returns a null output.
\input{case_studies/case5_noframes}

\label{case5_noresults}

In this type of error, thought the MLLM generates precise search expansions and the retrieval stage returns some results. However, the returned video results are all irrelevant. Consequently, the shot localization stage returns irrelevant frames.
\input{case_studies/case5_videowrong}

\label{case5_videowrong}

\subsubsection{Localizer errors}
In this type of error, the MLLM successfully generates precise search expansions and retrieves relevant videos. However, the video frame sampling process is flawed, causing irrelevant frames to be extracted, which leads to a 'False' judgement.
\input{case_studies/case5_sample_frames}

\label{case5_sample_frames}

In this error category, although the MLLM generates precise search expansions, a discrepancy occurs between the retrieved content and the model's judgment. Specifically, when the video retrieval stage returns an irrelevant candidate video set, the subsequent shot localization captures incorrect frames. However, the MLLM erroneously issues a 'True' judgment based on these irrelevant frames, contradicting human ground truth. Conversely, the error also manifests when the model produces a 'False' judgment despite the retrieval of correct videos and the extraction of accurate frames.
\input{case_studies/case5_verification}

\label{case5_verification_I}
\input{case_studies/case5_verification2}

\label{case5_verification_II}

\subsection{Case Study for video imagination}
\label{case}
We further illustrate the importance of video imagination versus nothink in constructing search queries through two concrete case studies.

\input{case_studies/case6}

\label{video1}
\input{case_studies/case7}

\label{video2}

Across these cases, video imagination demonstrates its ability to transform low-level shot details into high-level, semantically meaningful search queries, substantially improving retrieval success compared to naive, surface-level keyword concatenation.

%% file: case_studies/case_4_a.tex
\tcbset{
    breakable,
    colframe=blue!5!black,
    colback=gray!10!white,
    fonttitle=\bfseries,
    width=\columnwidth % 使tcolorbox适应单栏宽度
}

\begin{tcolorbox}[
    title=\textbf{Example of case A \\(MLLM True \& Human True)},
    fonttitle=\bfseries
]

\textbf{Selected frame:} \vspace{2pt} \\
\includegraphics[width=\columnwidth]
{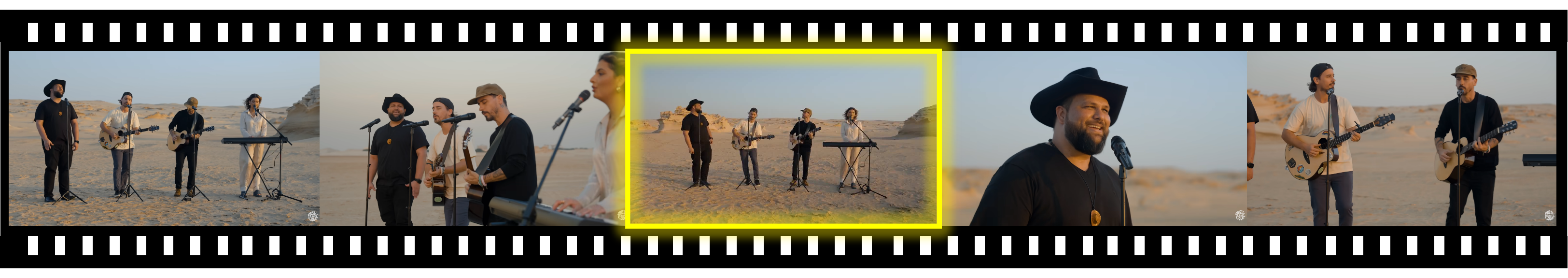}

\textcolor{red}{\textbf{Ground truth text:}} \vspace{2pt}\\
I need a video clip, \textcolor{red}{two male musicians stand side by side on an open expanse of sand: one plays an acoustic guitar while the other cradles his instrument, eyes closed in impassioned vocal performance.} Their postures are natural and fully immersed, with microphone stands positioned before them to form a performance array. The foreground \textcolor{red}{features the subjects and their instruments; the midground reveals a flat desert landscape; and the background presents softly blurred mountain silhouettes.} A horizontal composition enhances spatial depth.
The overall tone is natural, neutral-leaning warm, with a transparent and clean atmosphere. Colors are soft with low saturation and smooth transitions. The lighting is bright, characterized by soft natural diffused light without strong contrast or harsh shadows, resulting in a moderately low overall contrast.\\

\textcolor{red}{\textbf{Ground truth frame:}}\vspace{1pt}\\
\begin{center}
\includegraphics[height=1.5cm, keepaspectratio]{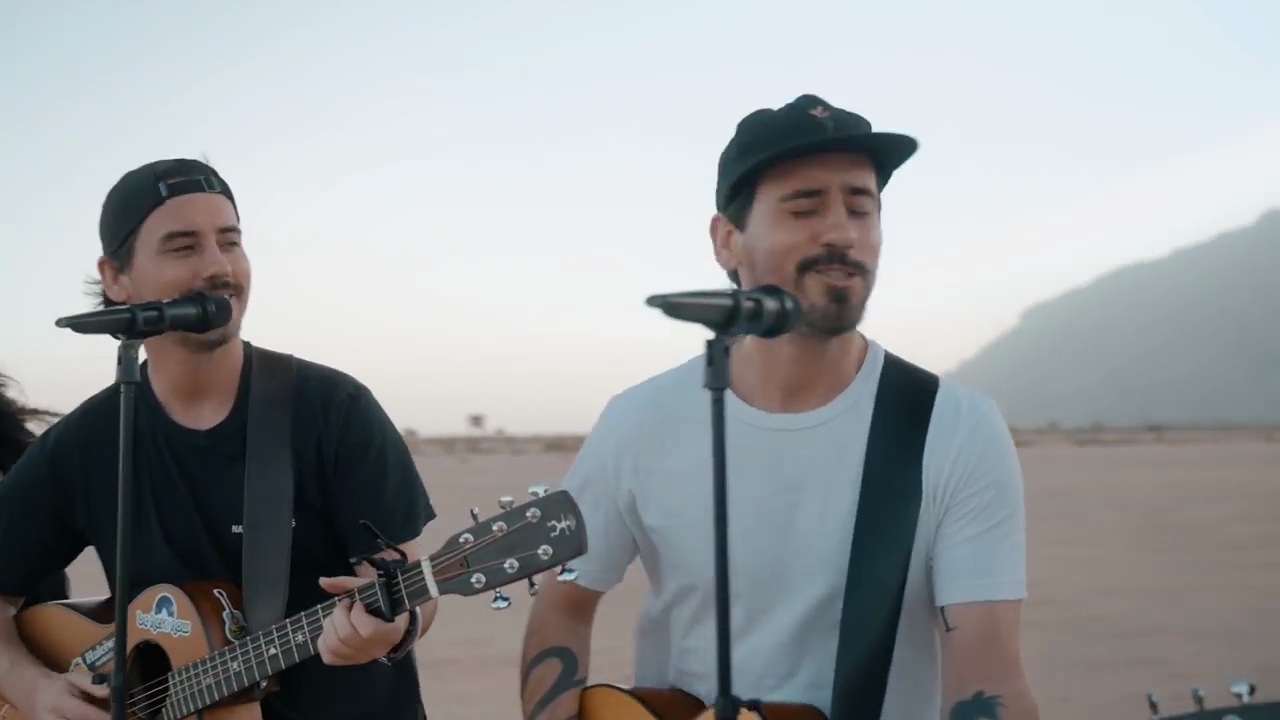}
\vspace{5pt}
\end{center}

\textbf{Model's prediction:}  \textcolor{PineGreen}{\textbf{True}}\\

\textbf{Human's prediction:}  \textcolor{PineGreen}{\textbf{True}}\\

\end{tcolorbox}

%% file: case_studies/case_4_b.tex
\tcbset{
    breakable,
    colframe=blue!5!black,
    colback=gray!10!white,
    fonttitle=\bfseries,
    width=\columnwidth % 使tcolorbox适应单栏宽度
}

\begin{tcolorbox}[
    title=\textbf{Example of case B \\(MLLM True \& Human False)},
    fonttitle=\bfseries
]

\textbf{Selected frame:} \vspace{2pt} \\
\includegraphics[width=\columnwidth]
{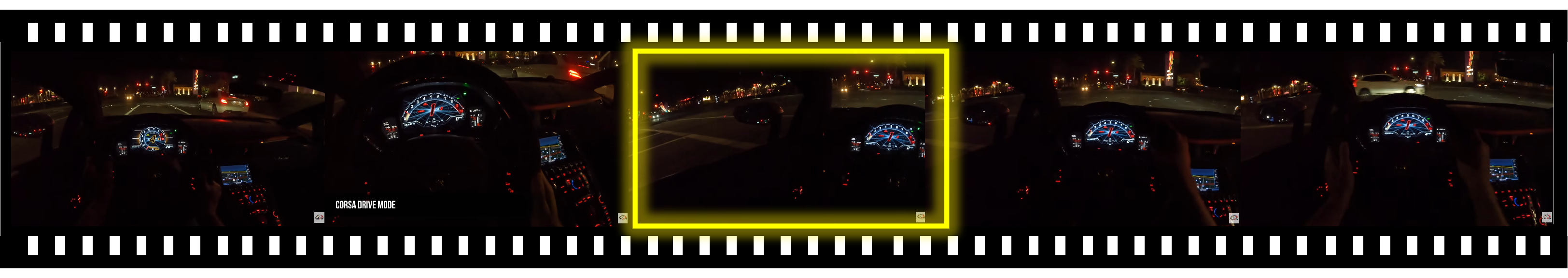}

\textcolor{red}{\textbf{Ground truth text:}} \vspace{2pt}\\
I am looking for a clip, this is a keyframe from \textcolor{red}{a nighttime highway racing perspective, captured with an immersive first-person wide-angle lens.} The foreground is dominated by the hood of a teal-modified sports car, with sharp black air vents and the right-side mirror clearly visible, enhancing the sense of driver immersion. In the midground, a white Lamborghini convertible supercar races ahead on the left, its low-slung body and prominent rear wing standing out dramatically against the night sky. \textcolor{red}{Floating text at the top displays the white English title 'Smoked a Lambo going home' and the central phrase 'wait for it...' builds suspense and foreshadows an impending overtaking climax.} Environmental lighting comes primarily from distant streetlights and digital billboards, creating high-contrast highlights against the dark sky.\\

\textcolor{red}{\textbf{Ground truth frame:}}\vspace{1pt}\\
\begin{center}
\includegraphics[height=2.2cm, keepaspectratio]{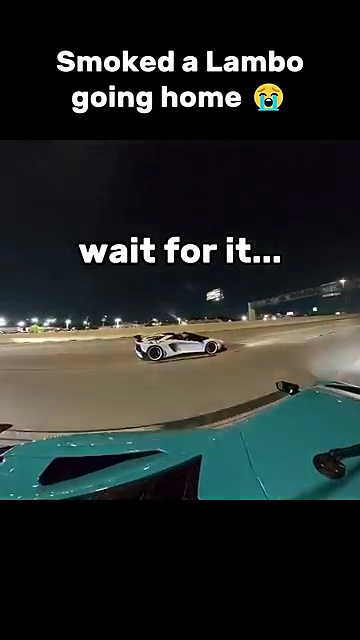}
\vspace{5pt}
\end{center}

\textbf{Model's prediction:}  \textcolor{PineGreen}{\textbf{True}}\\

\textbf{Human's prediction:}  \textcolor{red}{\textbf{False}}\\

\end{tcolorbox}

%% file: case_studies/case_4_c.tex
\tcbset{
    breakable,
    colframe=blue!5!black,
    colback=gray!10!white,
    fonttitle=\bfseries,
    width=\columnwidth % 使tcolorbox适应单栏宽度
}

\begin{tcolorbox}[
    title=\textbf{Example of case C \\(MLLM False \& Human True)},
    fonttitle=\bfseries
]

\textbf{Selected frame:} \vspace{2pt} \\
\includegraphics[width=\columnwidth]
{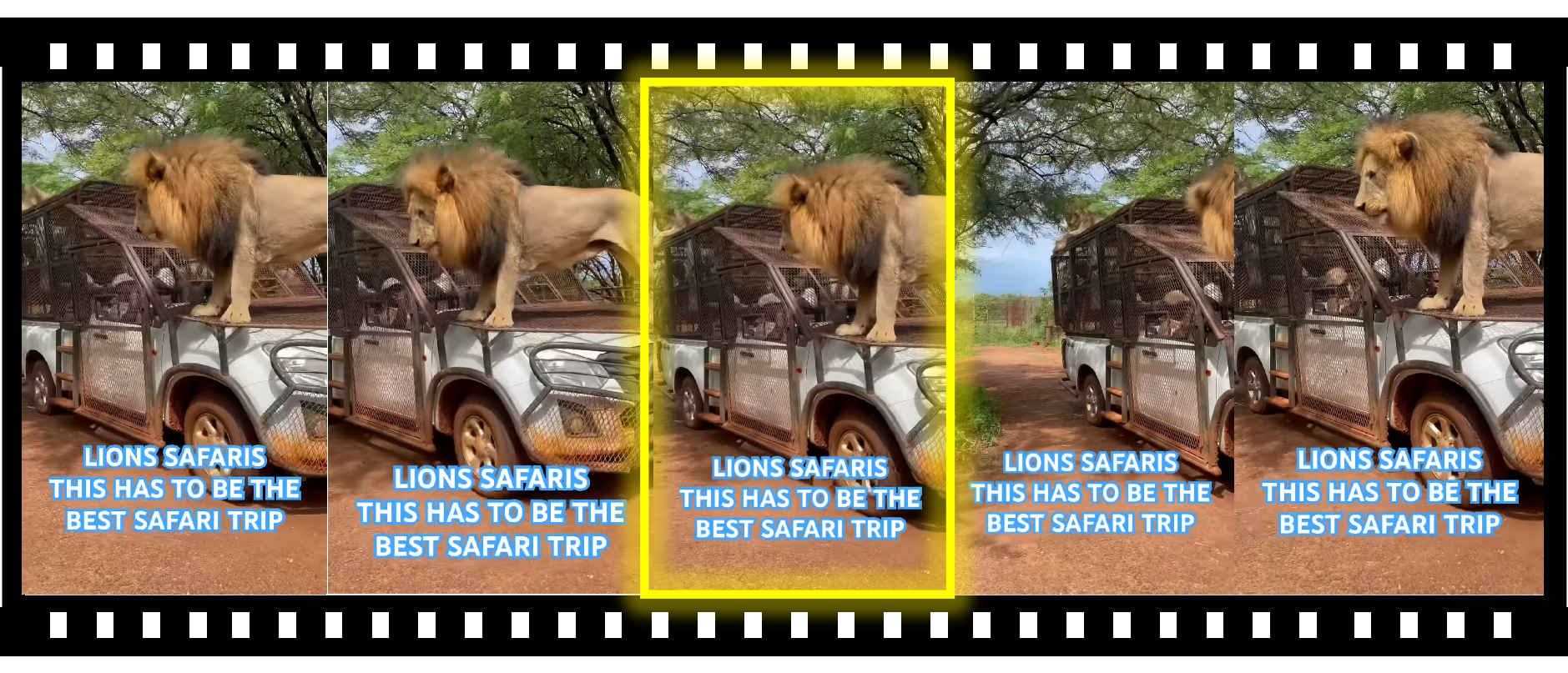}

\textcolor{red}{\textbf{Ground truth text:}} \vspace{2pt}\\
I want to find a video segment, \textcolor{red}{the main subject of this frame is a large, majestic adult male lion standing inside a metal transport cage mounted on the back of a truck. It has a thick, flowing mane, transitioning from golden yellow on the head to deep black around the neck, exuding strength and dignity.} The lion stands tall, alert and solemn, gazing intently toward the upper right of the frame. The cage, featuring a black metal frame and dense silver wire mesh, is positioned above a lowered metal ramp and a heavy-duty white bumper with circular holes, creating a strong industrial and oppressive feel. Bright natural sunlight streams from the upper right, clearly highlighting the texture of the lion's facial muscles and mane.\\

\textcolor{red}{\textbf{Ground truth frame:}}\vspace{1pt}\\
\begin{center}
\includegraphics[height=2.2cm, keepaspectratio]{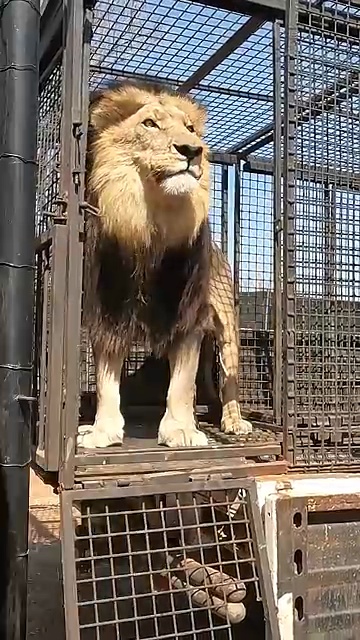}
\vspace{5pt}
\end{center}

\textbf{Model's prediction:}  \textcolor{red}{\textbf{False}}\\

\textbf{Human's prediction:}  \textcolor{PineGreen}{\textbf{True}}\\

\end{tcolorbox}

%% file: case_studies/case_4_d.tex
\tcbset{
    breakable,
    colframe=blue!5!black,
    colback=gray!10!white,
    fonttitle=\bfseries,
    width=\columnwidth % 使tcolorbox适应单栏宽度
}

\begin{tcolorbox}[
    title=\textbf{Example of case D \\(MLLM False \& Human False)},
    fonttitle=\bfseries
]

\textbf{Selected frames:} \vspace{2pt} \\
\includegraphics[width=\columnwidth]
{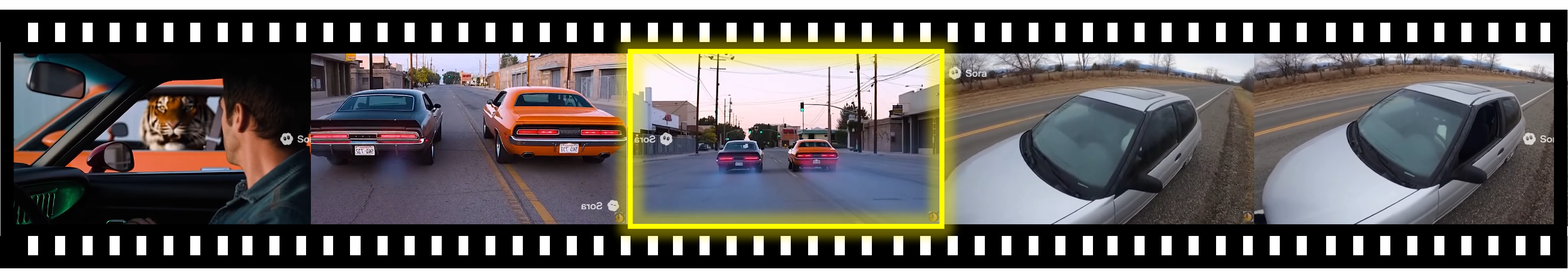}

\includegraphics[width=\columnwidth]
{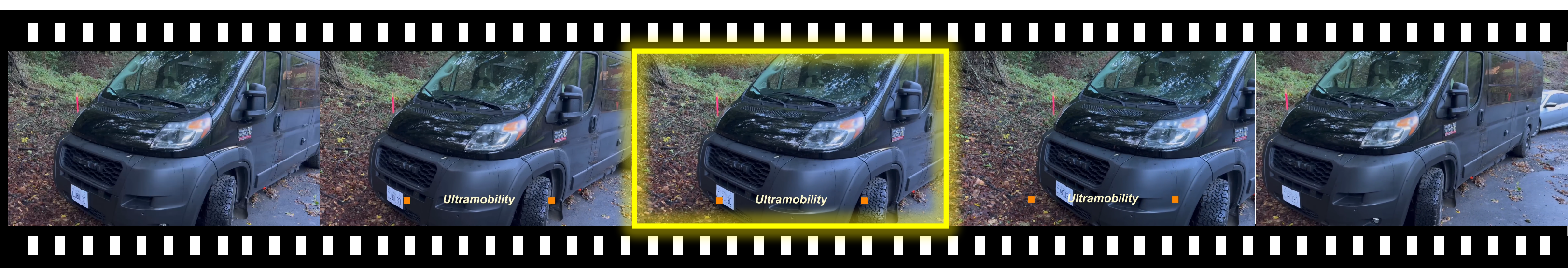}

\includegraphics[width=\columnwidth]
{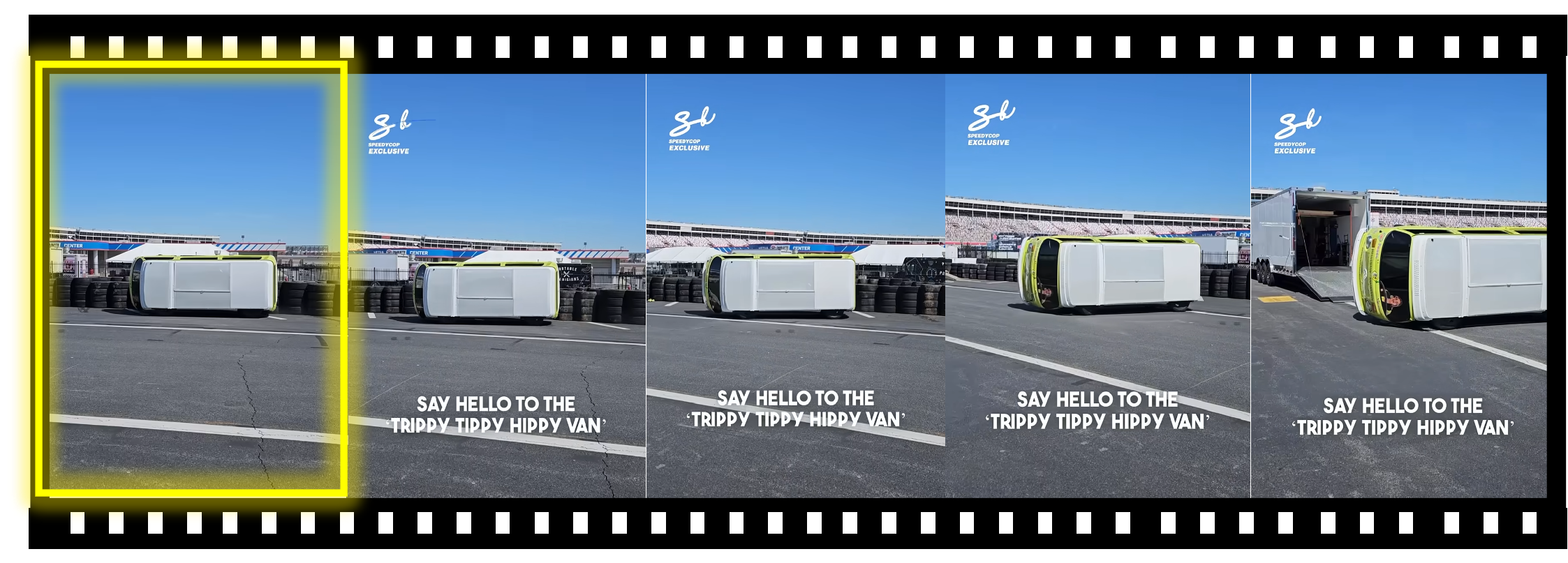}

\includegraphics[width=\columnwidth]
{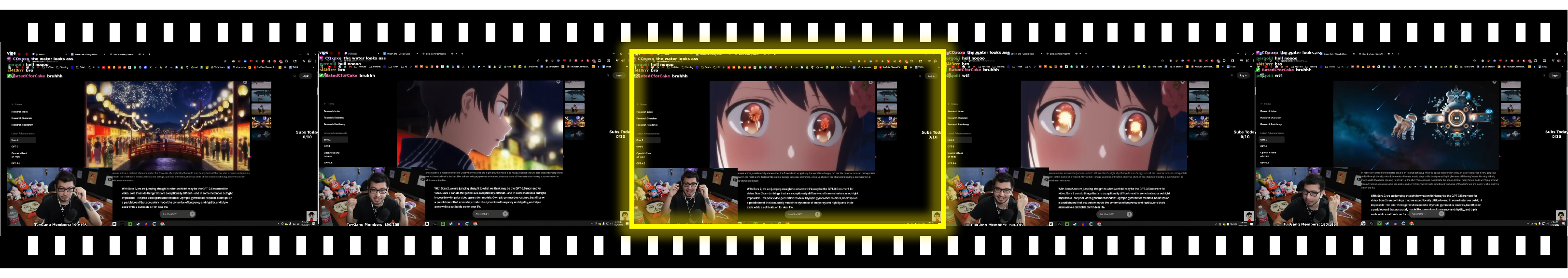}

\textcolor{red}{\textbf{Ground truth text:}} \vspace{2pt}\\
I am searching for a clip, \textcolor{red}{a black van drives centrally toward the camera, its headlights illuminated. The vehicle hugs the road with a low-slung profile, while the street recedes into a backdrop of urban buildings.} The foreground features \textcolor{red}{asphalt pavement with clear lane markings;} the midground is \textcolor{red}{dominated by the vehicle itself;} and the background comprises \textcolor{red}{multi-story buildings and street trees, resulting in a balanced, symmetrical composition.}The overall tone is neutral to warm and natural, dominated by the inherent colors of the realistic scene. Color saturation is moderate to soft, providing a plain and natural visual experience. The image is bright with moderate contrast between light and shadow, featuring both highlights from direct sunlight and clear shadows cast by objects, indicating distinct light direction and dimensionality.\\

\textcolor{red}{\textbf{Ground truth frame:}}\vspace{1pt}\\
\begin{center}
\includegraphics[height=2.2cm, keepaspectratio]{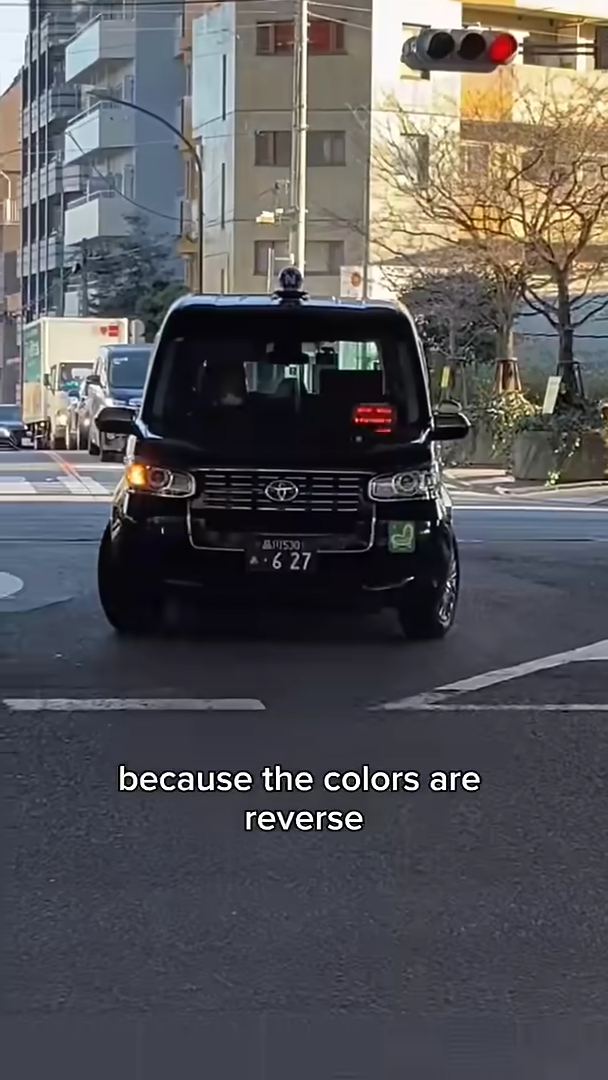}
\vspace{5pt}
\end{center}

\textbf{Model's prediction:}  \textcolor{red}{\textbf{False}}\\

\textbf{Human's prediction:}  \textcolor{red}{\textbf{False}}\\

\end{tcolorbox}

%% file: case_studies/case5_query.tex
\tcbset{
    breakable,
    colframe=blue!5!black,
    colback=gray!10!white,
    fonttitle=\bfseries,
    width=\columnwidth % 使tcolorbox适应单栏宽度
}

\begin{tcolorbox}[
    title=\textbf{Generate wrong queries},
    fonttitle=\bfseries
]

\textbf{Input fragment:} \vspace{2pt} \\
Please find the link to this video: I need a video clip, three figures clad in dark gear move swiftly across the snowfield in a tight triangular formation, maintaining low postures, synchronized movements, and consistent spacing between them, casting sharp, elongated diagonal shadows. In the foreground, textured undulations of ice create intricate surface detail; the midground features the trio as the primary subjects; and the background reveals towering snow-capped peaks and glacial escarpments, establishing a composition that is expansive with pronounced depth.\\

\textbf{MLLM output context:} \vspace{1pt} \\
\textless think\textgreater The description is highly detailed and specific ("three figures clad in dark gear," "triangular formation," "textured undulations of ice"), \textcolor{red}{which strongly suggests it is a prompt or description of an AI-generated video, likely from OpenAI's Sora showcase. The phrasing matches the style of prompts used to demonstrate Sora's capabilities.}\textless/think\textgreater 

\textless tool\_call\textgreater\{\textquotedbl name\textquotedbl: \textquotedbl search\_videos\textquotedbl, \textquotedbl arguments\textquotedbl: \{\textquotedbl query\textquotedbl: [\textquotedbl \textcolor{red}{Sora AI} three figures dark gear snowfield\textquotedbl, \textquotedbl three figures \textcolor{red}{snow triangular formation cinematic}\textquotedbl]\}\}\textless/tool\_call\textgreater \\

% \textcolor{red}{\textbf{The generated queries did not reflect humans and snowfield. Filtered frames are about Sora AI and snow triangular without people.}} \\

\textbf{Selected frames:}\vspace{1pt}\\
\includegraphics[width=\columnwidth]
{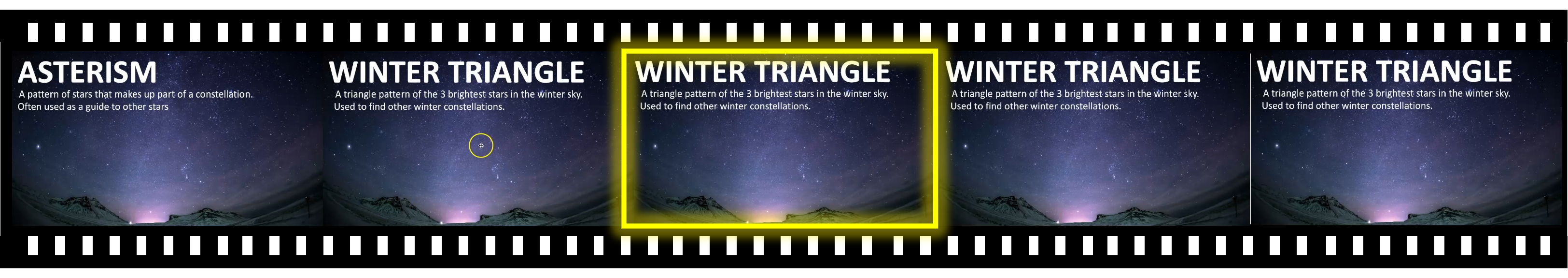}

\includegraphics[width=\columnwidth]
{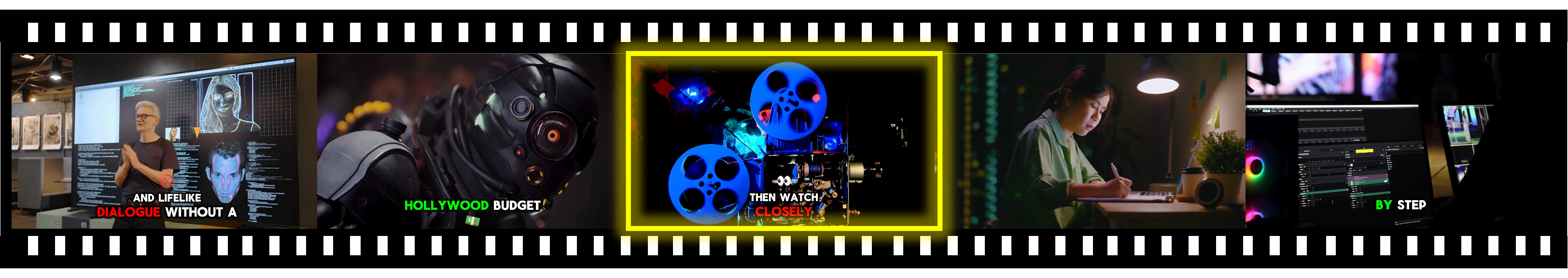}

\includegraphics[width=\columnwidth]
{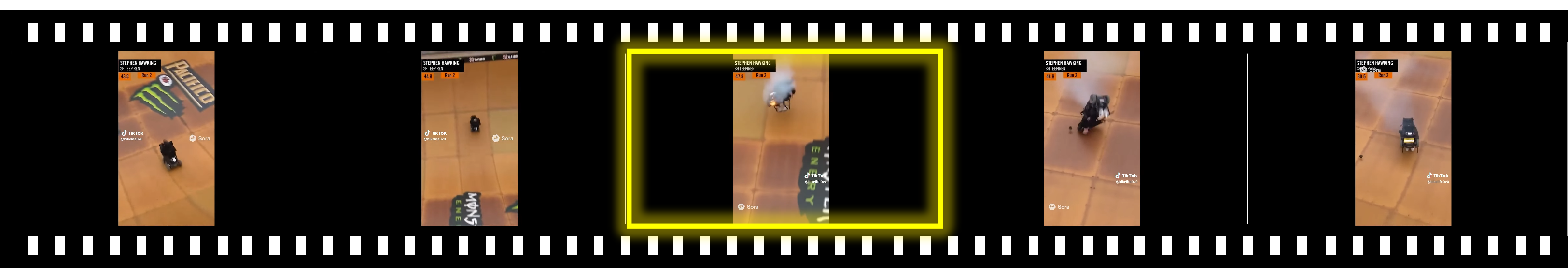}

\includegraphics[width=\columnwidth]
{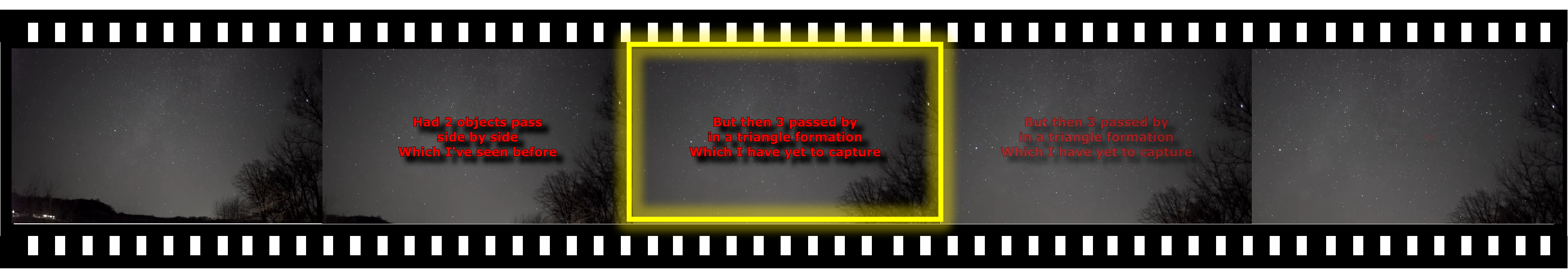}

\textbf{Ground truth text:}\vspace{1pt}\\
I need a video clip, \textcolor{red}{three figures clad in dark gear move swiftly across the snowfield in a tight triangular formation, maintaining low postures, synchronized movements, and consistent spacing between them, casting sharp, elongated diagonal shadows.} In the foreground, \textcolor{red}{textured undulations of ice create intricate surface detail;} the midground features the trio as the primary subjects; and the background \textcolor{red}{reveals towering snow-capped peaks and glacial escarpments, establishing a composition that is expansive with pronounced depth. (1080P)}

\textbf{Ground truth frame:}\vspace{1pt}\\
\begin{center}
\includegraphics[height=1.5cm, keepaspectratio]{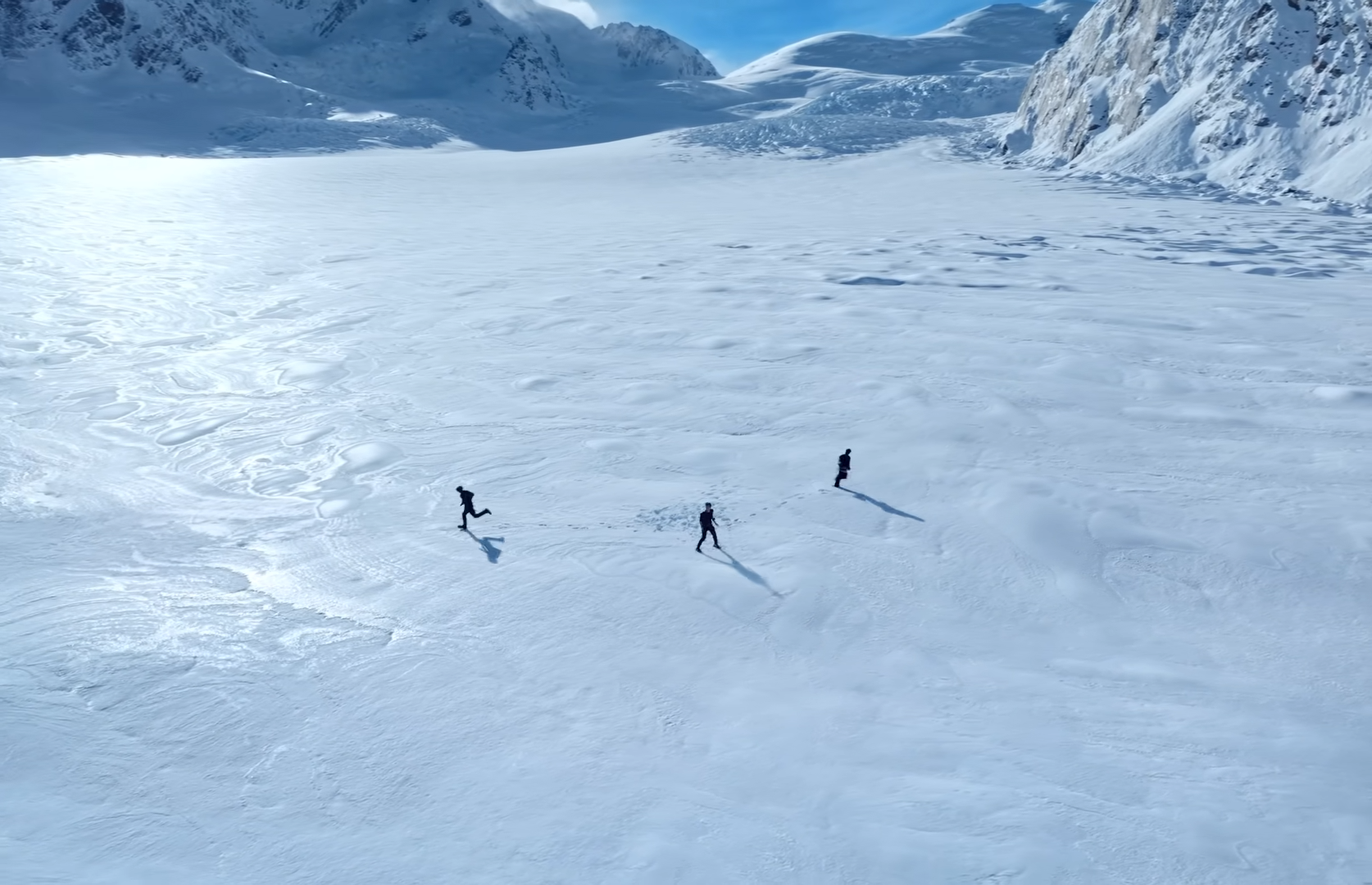}
\vspace{5pt}
\end{center}

\textbf{Model's prediction:}  \textcolor{red}{\textbf{False}}\\

\end{tcolorbox}

%% file: case_studies/case5_query2.tex
\tcbset{
    breakable,
    colframe=blue!5!black,
    colback=gray!10!white,
    fonttitle=\bfseries,
    width=\columnwidth % 使tcolorbox适应单栏宽度
}

\begin{tcolorbox}[
    title=\textbf{Generate partially correct queries},
    fonttitle=\bfseries
]

\textbf{Input fragment:} \vspace{2pt} \\
Please find the link to this video: I need a video clip, a blonde woman stands centered on stage, wearing a tailored grayish-brown dress, her hands gently clasped at her chest in a poised and dignified posture, as if delivering a speech or expressing gratitude. The background features a deep blue gradient backdrop with naturally cascading folds that create soft vertical depth; the foreground subject is sharply rendered and prominently highlighted, while the midground remains free of distracting elements, resulting in a balanced and symmetrical composition.\\

\textbf{MLLM output context:} \vspace{1pt} \\
\textless think\textgreater The detailed, descriptive language ("sharply rendered", "midground remains free", "deep blue gradient backdrop with naturally cascading folds") strongly suggests this is a prompt or description of a specific AI-generated video demo, likely from OpenAI's Sora or similar high-end video generation models. The visual specifics match a known Sora example.\textless/think\textgreater 

\textless tool\_call\textgreater\{\textquotedbl name\textquotedbl: \textquotedbl search\_videos\textquotedbl, \textquotedbl arguments\textquotedbl: \{\textquotedbl query\textquotedbl: [\textquotedbl \textcolor{red}{Sora AI blonde woman speech blue background} \textquotedbl, \textquotedbl\textcolor{PineGreen}{blonde woman grayish brown dress deep blue backdrop} \textit{(missing:handclaspe)}\textquotedbl]\}\}\textless/tool\_call\textgreater \\

% \textcolor{red}{\textbf{The generated queries did not reflect humans and snowfield. Filtered frames are about Sora AI and snow triangular without people.}} \\

\textbf{Selected frames:}\vspace{1pt}\\
\includegraphics[width=\columnwidth]
{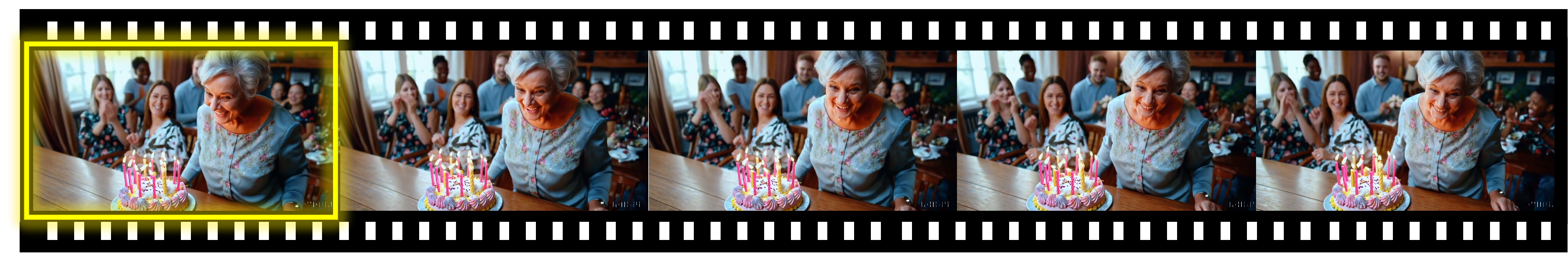}

\includegraphics[width=\columnwidth]
{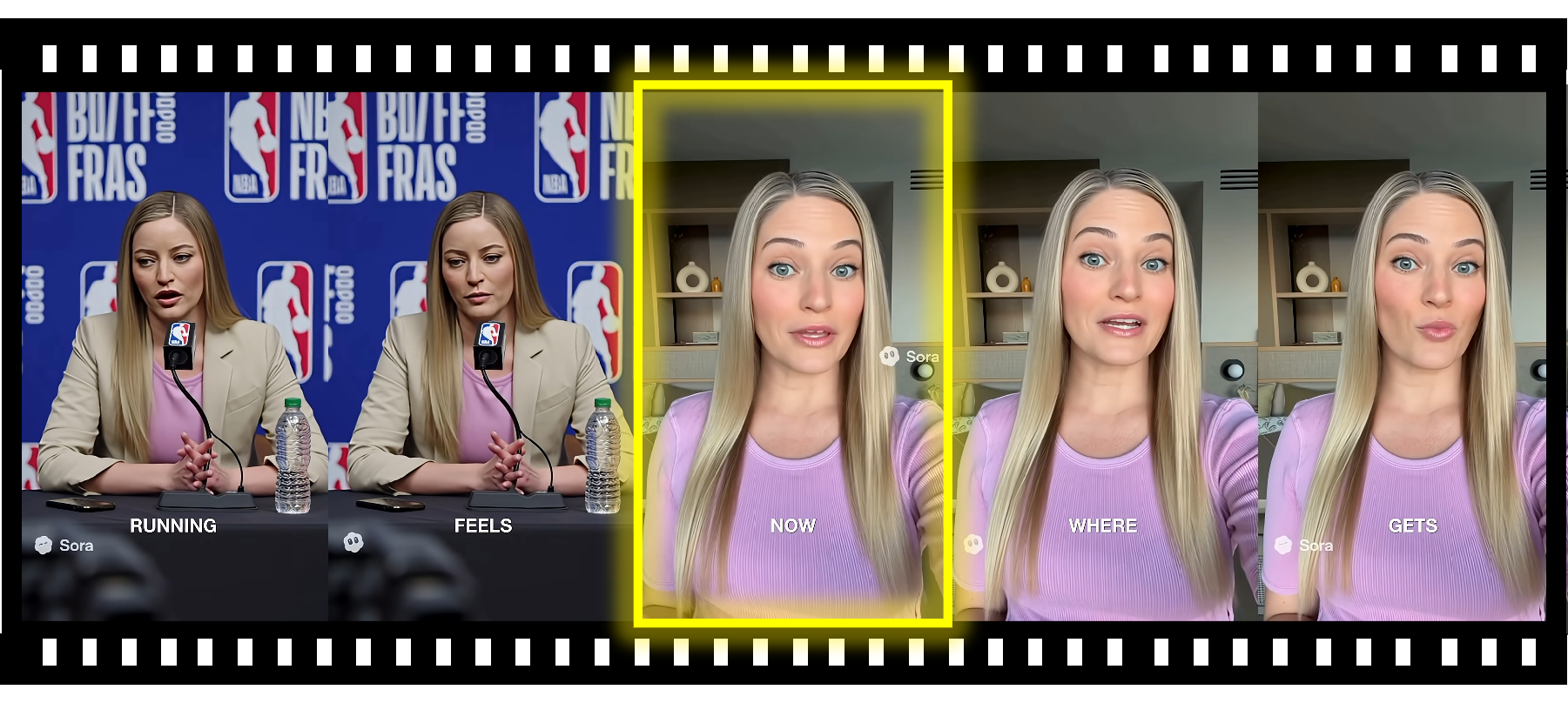}

\includegraphics[width=\columnwidth]
{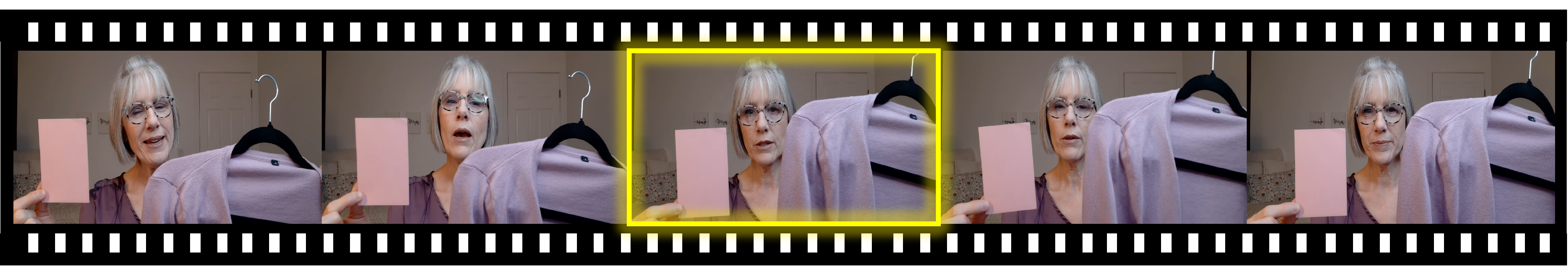}

\includegraphics[width=\columnwidth]
{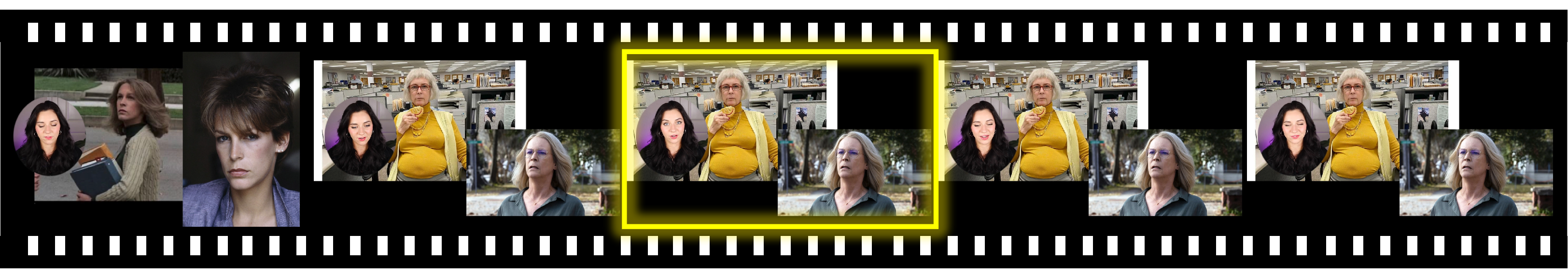}

\textbf{Ground truth text:}\vspace{1pt}\\
I need a video clip, \textcolor{red}{a blonde woman stands centered on stage, wearing a tailored grayish-brown dress, her hands gently clasped at her chest in a poised and dignified posture}, as if delivering a speech or expressing gratitude. The background features \textcolor{red}{a deep blue gradient backdrop with naturally cascading folds that create soft vertical depth}; the foreground subject is sharply rendered and prominently highlighted, while the midground remains free of distracting elements, resulting in a balanced and symmetrical composition.(1080P)

\textbf{Ground truth frame:}\vspace{1pt}\\
\begin{center}
\includegraphics[height=1.5cm, keepaspectratio]{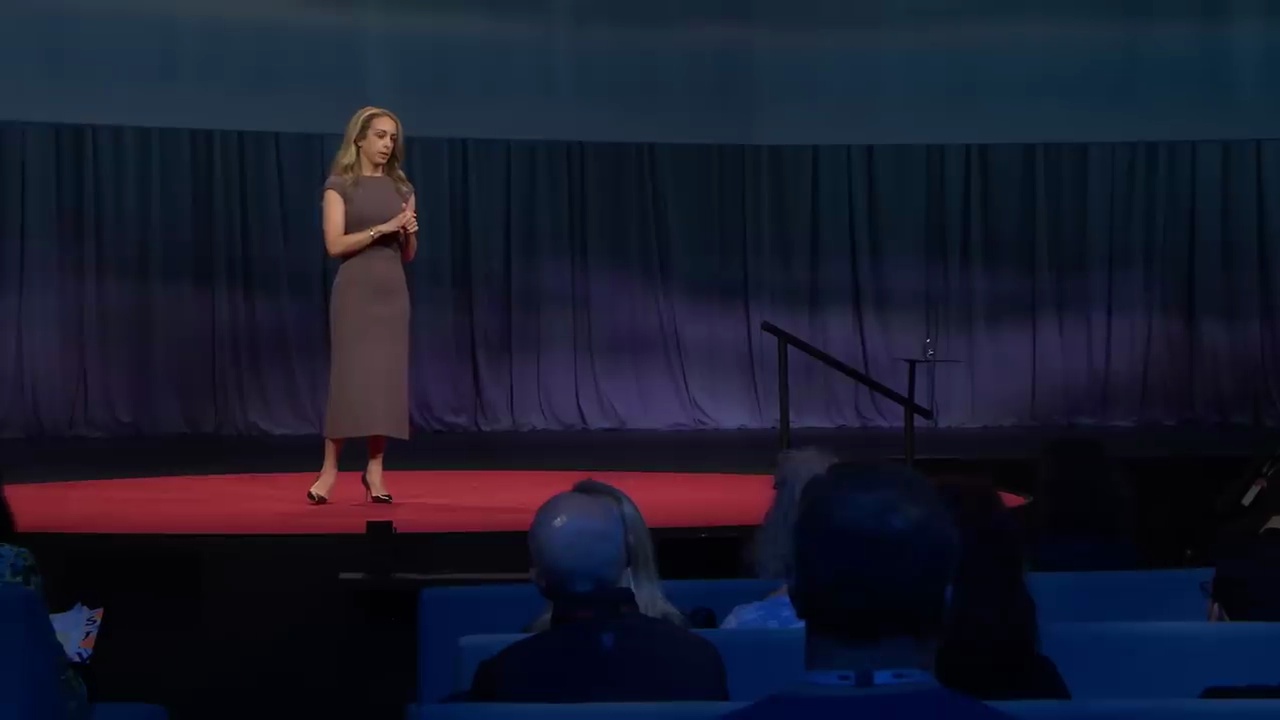}
\vspace{5pt}
\end{center}

\textbf{Model's prediction:}  \textcolor{red}{\textbf{False}}\\

\end{tcolorbox}

%% file: case_studies/case5_noframes.tex
\tcbset{
    breakable,
    colframe=blue!5!black,
    colback=gray!10!white,
    fonttitle=\bfseries,
    width=\columnwidth % 使tcolorbox适应单栏宽度
}

\begin{tcolorbox}[
    title=\textbf{No searched videos},
    fonttitle=\bfseries
]

\textbf{Input fragment:} \vspace{2pt} \\
Please find the link to this video: I need a piece of stock footage, two drummers stand side by side at the center of the stage, clad in black sleeveless tops, long trousers, and brown boots, simultaneously striking wooden war drums suspended at their waists with drumsticks. Their movements are perfectly synchronized and charged with tension. The foreground features a dark, reflective stage surface; the midground frames the performers; and the background consists of a metal mesh panel overlaid with a dot-matrix LED wall, creating a strong sense of depth. The shot is a static, eye-level medium frame that fully captures the performers’ upper bodies and their drums, emphasizing rhythmic dynamism. Overhead top lighting interplays with side-mounted green lights, sharply delineating the performers’ silhouettes, while deep shadows in the background enhance the subjects’ presence and evoke an intense yet enigmatic atmosphere. Real-shot. The visual style is typical of stage performance photography, emphasizing the capture of dynamic moments and realistic states. The composition presents obvious symmetry, with a realistic visual language that focuses on showcasing the on-site atmosphere and tension of movement.\\

\textbf{MLLM output context:} \vspace{1pt} \\
\textless think\textgreater The user describes a specific stock footage clip featuring two synchronized drummers with war drums on a stage with green lighting and a metal mesh background. Key visual elements include the symmetry, the specific outfit (black sleeveless), and the lighting conditions. I will construct queries focusing on the drummers, the type of drum, and the lighting/stage atmosphere. \textless/think\textgreater 

\textless tool\_call\textgreater\{\textquotedbl name\textquotedbl: \textquotedbl search\_videos\textquotedbl, \textquotedbl arguments\textquotedbl: \{\textquotedbl query\textquotedbl: [\textquotedbl two drummers synchronized war drums green lights stage \textquotedbl, \textquotedbl stock footage two drummers black sleeveless wooden drums\textquotedbl]\}\}\textless/tool\_call\textgreater \\

% \textcolor{red}{\textbf{The generated queries did not reflect humans and snowfield. Filtered frames are about Sora AI and snow triangular without people.}} \\

\textbf{Selected frames:}\vspace{1pt}\\
\textcolor{red}{\textbf{No selected videos.}}

\textbf{Ground truth text:}\vspace{1pt}\\
I need a piece of stock footage, two drummers stand side by side at the center of the stage, clad in black sleeveless tops, long trousers, and brown boots, simultaneously striking wooden war drums suspended at their waists with drumsticks. Their movements are perfectly synchronized and charged with tension. The foreground features a dark, reflective stage surface; the midground frames the performers; and the background consists of a metal mesh panel overlaid with a dot-matrix LED wall, creating a strong sense of depth. The shot is a static, eye-level medium frame that fully captures the performers’ upper bodies and their drums, emphasizing rhythmic dynamism. Overhead top lighting interplays with side-mounted green lights, sharply delineating the performers’ silhouettes, while deep shadows in the background enhance the subjects’ presence and evoke an intense yet enigmatic atmosphere.Real-shot. The visual style is typical of stage performance photography, emphasizing the capture of dynamic moments and realistic states. The composition presents obvious symmetry, with a realistic visual language that focuses on showcasing the on-site atmosphere and tension of movement.

\textbf{Ground truth frame:}\vspace{1pt}\\
\begin{center}
\includegraphics[height=1.5cm, keepaspectratio]{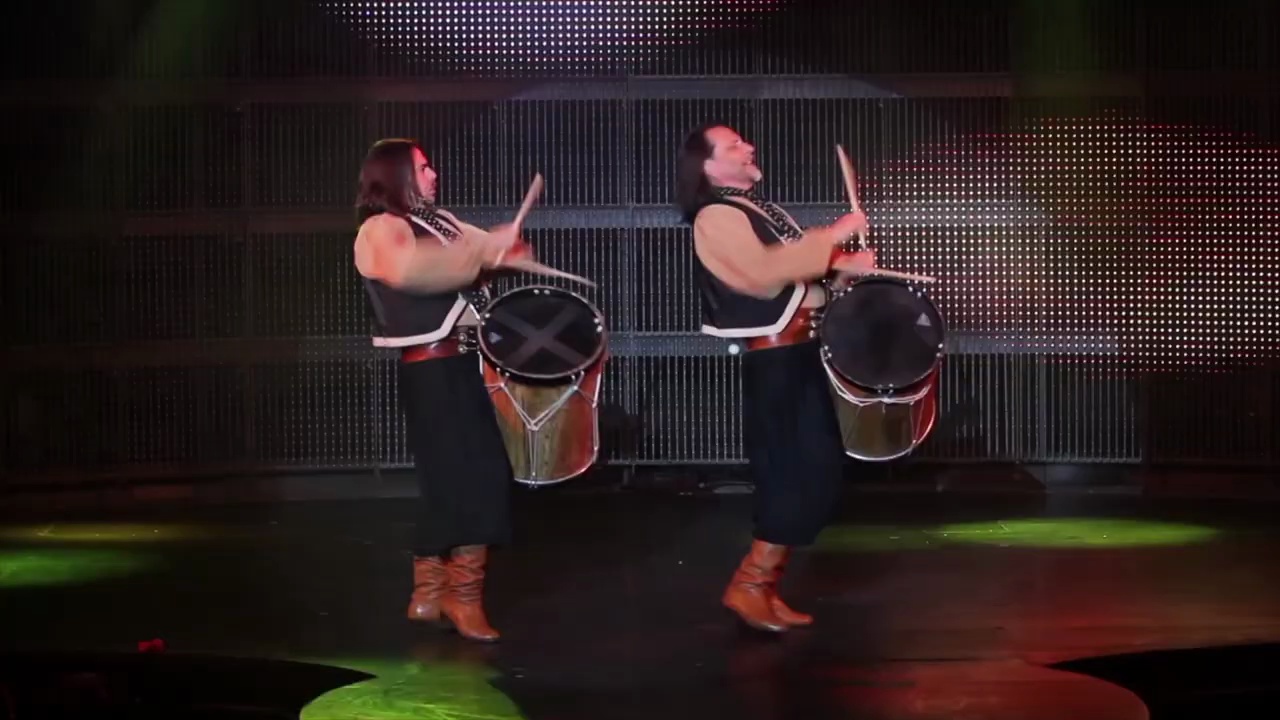}
\vspace{5pt}
\end{center}

\textbf{Model's prediction:}  \textcolor{red}{\textbf{False}}\\

\end{tcolorbox}

%% file: case_studies/case5_videowrong.tex
\tcbset{
    breakable,
    colframe=blue!5!black,
    colback=gray!10!white,
    fonttitle=\bfseries,
    width=\columnwidth % 使tcolorbox适应单栏宽度
}

\begin{tcolorbox}[
    title=\textbf{Searched videos are all wrong},
    fonttitle=\bfseries
]

\textbf{Input fragment:} \vspace{2pt} \\
Please find the link to this video: I need a piece of stock footage, the primary subject is a standing male figure wearing an olive-green jacket, holding a luminous device in his right hand and raising his left hand in a natural, focused explanatory gesture. In the foreground lies an orange tabletop with black equipment; the midground features the main subject; and the background displays large electronic instruments alongside a blurred topographic map, establishing a clear sense of depth and layered composition. The shot is a static eye-level medium frame, tightly framing the upper body of the subject with no discernible camera movement, ensuring a stable presentation of the demonstration scene.\\

\textbf{MLLM output context:} \vspace{1pt} \\
\textless think\textgreater The user describes a specific stock footage clip featuring a man in an olive-green jacket holding a luminous device with an orange table and electronic equipment. The setting suggests a tech, scientific, or military context. Keywords should focus on the visual elements and the specific action.\textless/think\textgreater 

\textless tool\_call\textgreater\{\textquotedbl name\textquotedbl: \textquotedbl search\_videos\textquotedbl, \textquotedbl arguments\textquotedbl: \{\textquotedbl query\textquotedbl: [\textquotedbl \textcolor{PineGreen}{stock footage man olive green jacket luminous device orange table}\textquotedbl, \textquotedbl \textcolor{PineGreen}{man holding luminous device explanatory gesture electronic instruments stock video}\textquotedbl]\}\}\textless/tool\_call\textgreater \\

\textbf{Selected frames from the filtered videos:}\vspace{1pt}\\
\includegraphics[width=\columnwidth]
{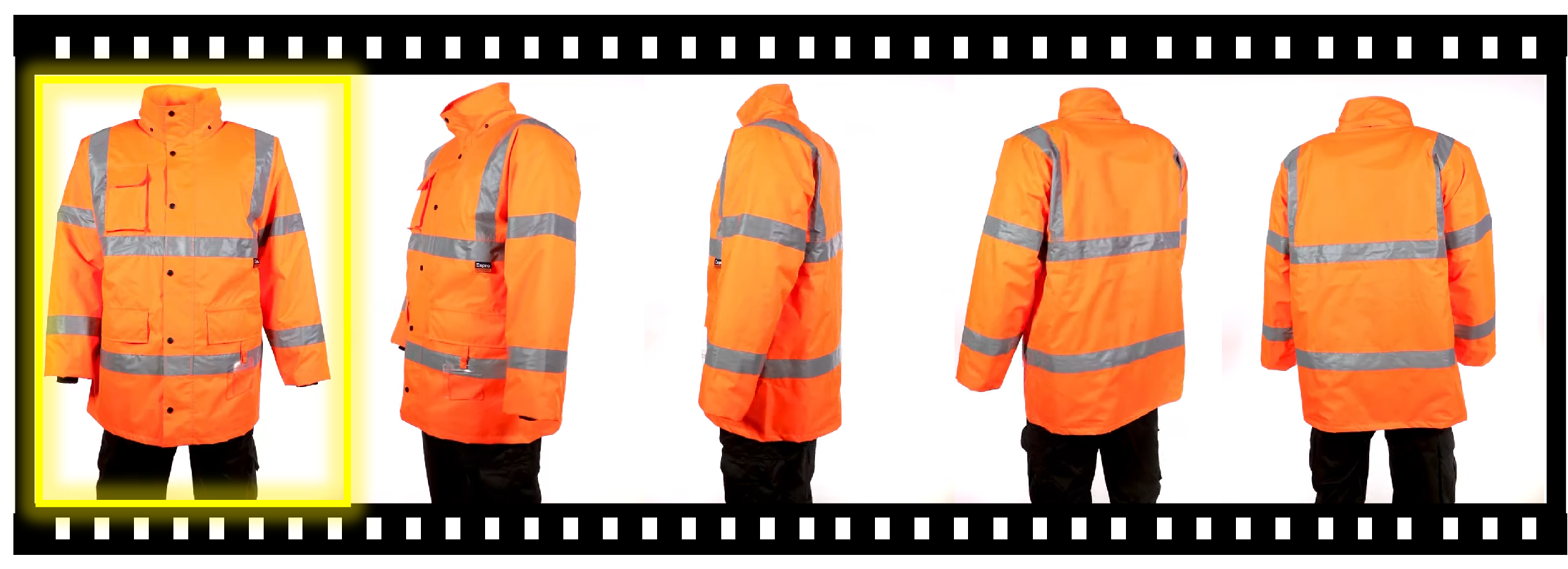}

\includegraphics[width=\columnwidth]
{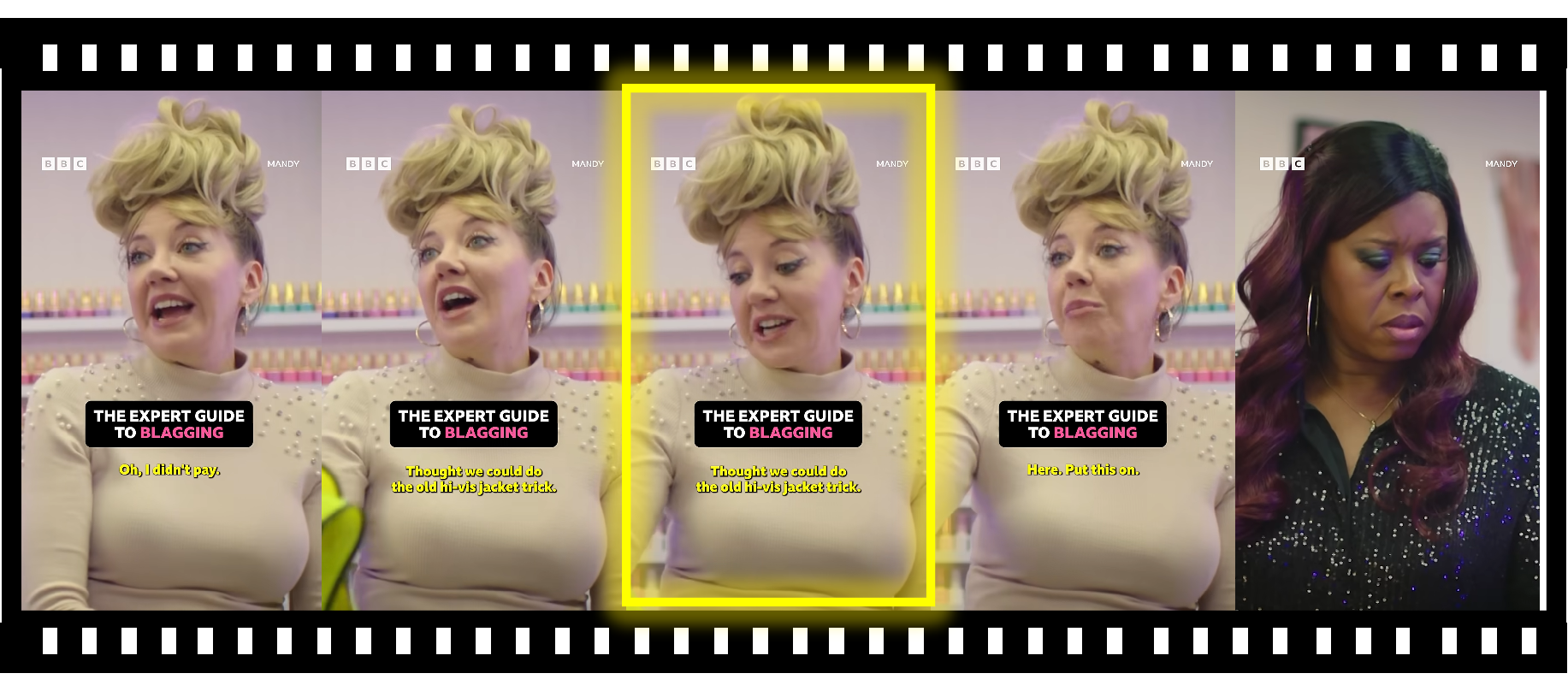}

\includegraphics[width=\columnwidth]
{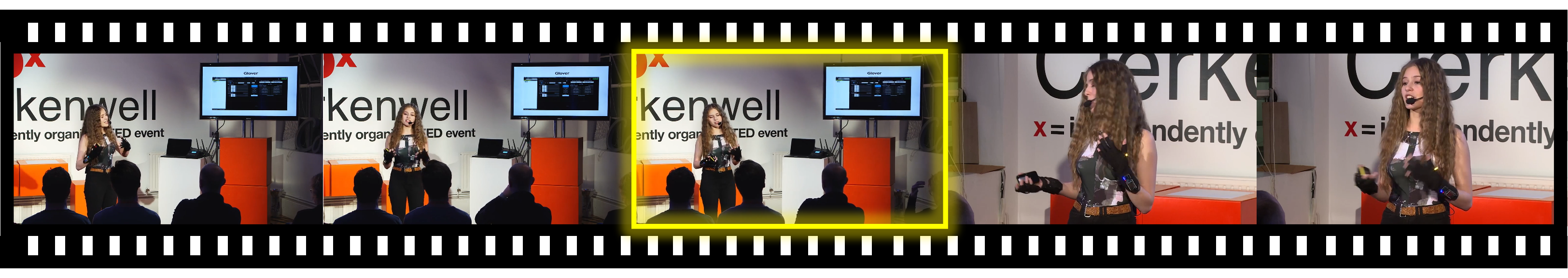}

\includegraphics[width=\columnwidth]
{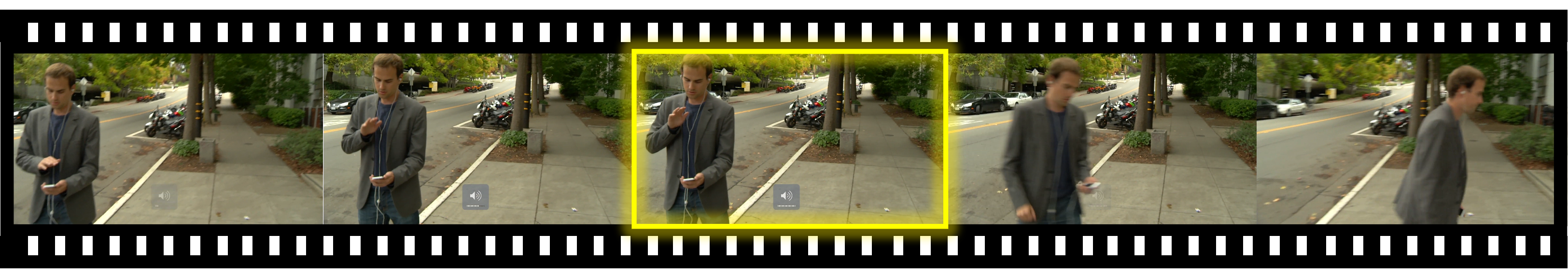}

\textbf{Selected frame from the dropped video (we choose the most similar frame):}\vspace{1pt}\\
\includegraphics[width=\columnwidth]
{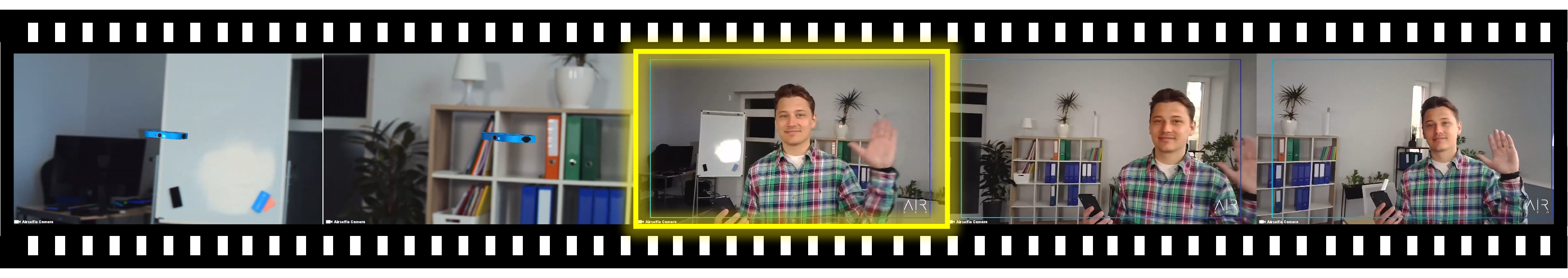}\\

% \textcolor{red}{\textbf{The generated queries are correct, but search wrong videos.}} \\ 

\textbf{Ground truth text:}\vspace{1pt}\\
I need a piece of stock footage, the primary subject is \textcolor{red}{a standing male figure wearing an olive-green jacket, holding a luminous device in his right hand and raising his left hand in a natural, focused explanatory gesture.} In the foreground lies an \textcolor{red}{orange tabletop with black equipment;} the midground features the main subject; and the background displays \textcolor{red}{large electronic instruments alongside a blurred topographic map, establishing a clear sense of depth and layered composition.} The shot is a static eye-level medium frame, tightly framing the upper body of the subject with no discernible camera movement, ensuring a stable presentation of the demonstration scene. (1080P)

\textbf{Ground truth frame:}\vspace{1pt}\\
\begin{center}
\includegraphics[height=2.0cm, keepaspectratio]{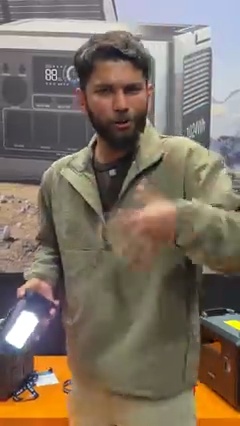}
\vspace{5pt}
\end{center}

\textbf{Model's prediction:}  \textcolor{red}{\textbf{False}}\\

\end{tcolorbox}

%% file: case_studies/case5_sample_frames.tex
\tcbset{
    breakable,
    colframe=blue!5!black,
    colback=gray!10!white,
    fonttitle=\bfseries,
    width=\columnwidth % 使tcolorbox适应单栏宽度
}

\begin{tcolorbox}[
    title=\textbf{Sample wrong frames},
    fonttitle=\bfseries
]

\textbf{Input fragment:} \vspace{2pt} \\
Please find the link to this video: I need to get a clip, this keyframe captures a breathtaking moment of a massive orca leaping vertically out of an artificial pool. The orca's distinct black-and-white streamlined body is almost fully airborne, with its dorsal fin upright, showcasing immense power. As it surges upward, it generates a burst of churning white water, with droplets glistening in the sunlight. To the right, a trainer in a blue-and-black wetsuit stands at the pool's edge, facing the orca with hands raised as if applauding or giving commands, creating clear interaction. The background features a painted ocean mural, metal railings, and distant greenery, with a prominent orange traffic cone by the poolside. I vaguely recall the audio being lively and chaotic, as if I were in some sort of obstacle course game. From what I remember, there was a rapid and cheerful soundtrack, interspersed with exaggerated sound effects of objects crashing violently. I think there were also some indistinct shouts, creating a tense yet comical and joyful atmosphere that left a lasting impression on me.\\

\textbf{MLLM output context:} \vspace{1pt} \\
\textless think\textgreater The user describes a specific visual of an orca jumping near a trainer with an orange cone, combined with audio resembling a chaotic obstacle course game (likely *Wipeout* or *Fall Guys*). This suggests the user saw a meme edit. I will search for the visual elements to find the source clip and the meme context.\textless/think\textgreater 

\textless tool\_call\textgreater\{\textquotedbl name\textquotedbl: \textquotedbl search\_videos\textquotedbl, \textquotedbl arguments\textquotedbl: \{\textquotedbl query\textquotedbl: [\textquotedbl \textcolor{PineGreen}{orca jump trainer orange cone}\textquotedbl, \textquotedbl \textcolor{PineGreen}{orca jump wipeout meme}\textquotedbl]\}\}\textless/tool\_call\textgreater \\

\textbf{Selected frame:}\vspace{1pt}\\
\includegraphics[width=\columnwidth]
{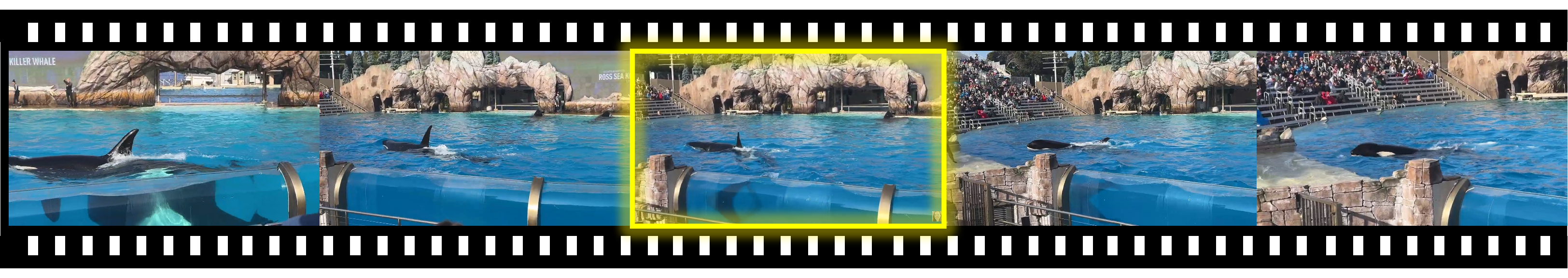}

\textbf{True frame in the same video:}\vspace{1pt}\\
\includegraphics[width=\columnwidth]
{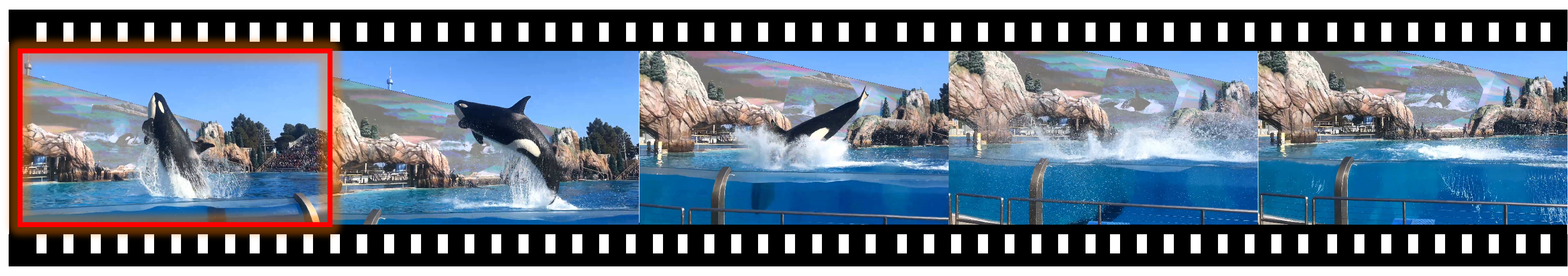}

\textbf{Ground truth text:}\vspace{1pt}\\
I need to get a clip, this keyframe captures a breathtaking moment of \textcolor{red}{a massive orca leaping vertically out of an artificial pool. The orca's distinct black-and-white streamlined body is almost fully airborne, with its dorsal fin upright, showcasing immense power.} As it surges upward, it generates a burst of churning white water, with droplets glistening in the sunlight. To the right, a trainer in a blue-and-black wetsuit stands at the pool's edge, facing the orca with hands raised as if applauding or giving commands, creating clear interaction. \textcolor{red}{The background features a painted ocean mural, metal railings, and distant greenery, with a prominent orange traffic cone by the poolside. (1080P)} \\

\textbf{Ground truth frame:}\vspace{1pt}\\
\begin{center}
\includegraphics[height=1.5cm, keepaspectratio]{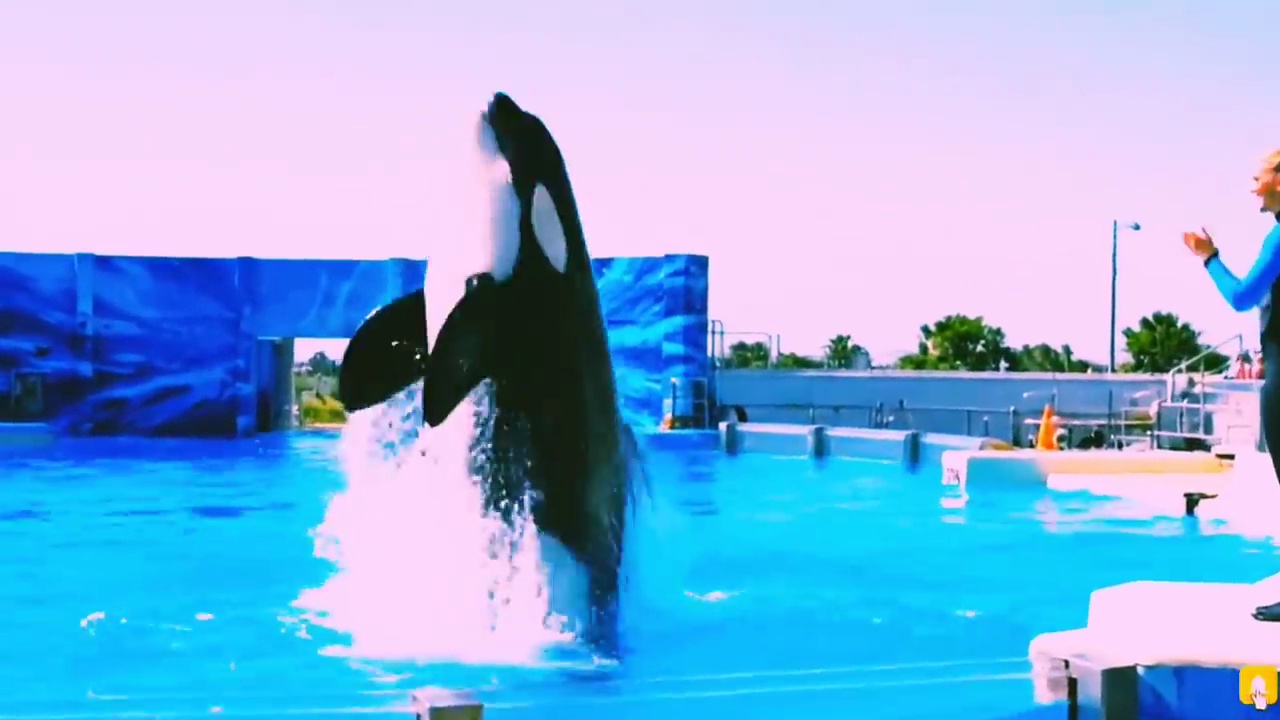}
\vspace{5pt}
\end{center}

\textbf{Model's prediction:}  \textcolor{red}{\textbf{False}}\\

\end{tcolorbox}

%% file: case_studies/case5_verification.tex
\tcbset{
    breakable,
    colframe=blue!5!black,
    colback=gray!10!white,
    fonttitle=\bfseries,
    width=\columnwidth % 使tcolorbox适应单栏宽度
}

\begin{tcolorbox}[
    title=\textbf{Wrong verification (I)},
    fonttitle=\bfseries
]

\textbf{Input fragment:} \vspace{2pt} \\
Please find the link to this video: I am looking for a clip, this is a keyframe from a nighttime highway racing perspective, captured with an immersive first-person wide-angle lens. The foreground is dominated by the hood of a teal-modified sports car, with sharp black air vents and the right-side mirror clearly visible, enhancing the sense of driver immersion. In the midground, a white Lamborghini convertible supercar races ahead on the left, its low-slung body and prominent rear wing standing out dramatically against the night sky. Floating text at the top displays the white English title 'Smoked a Lambo going home' and the central phrase 'wait for it...' builds suspense and foreshadows an impending overtaking climax. Environmental lighting comes primarily from distant streetlights and digital billboards, creating high-contrast highlights against the dark sky.\\

\textbf{MLLM output context:} \vspace{1pt} \\
\textless think\textgreater The user describes a specific POV street racing video. Key identifiers are the text overlays "Smoked a Lambo going home" and "wait for it...", along with a teal sports car chasing a white Lamborghini convertible at night. The text is the strongest search signal.\textless/think\textgreater 

\textless tool\_call\textgreater\{\textquotedbl name\textquotedbl: \textquotedbl search\_videos\textquotedbl, \textquotedbl arguments\textquotedbl: \{\textquotedbl query\textquotedbl: [\textquotedbl \textcolor{PineGreen}{Smoked a Lambo going home wait for it}\textquotedbl, \textquotedbl \textcolor{PineGreen}{teal car racing white lamborghini night pov}\textquotedbl]\}\}\textless/tool\_call\textgreater \\

\textbf{Selected frame:}\vspace{1pt}\\
\includegraphics[width=\columnwidth]
{case_studies/photos/4b.png}

\textbf{Ground truth text:} \vspace{2pt}\\
I am looking for a clip, this is a keyframe from \textcolor{red}{a nighttime highway racing perspective, captured with an immersive first-person wide-angle lens.} The foreground is dominated by the hood of a teal-modified sports car, with sharp black air vents and the right-side mirror clearly visible, enhancing the sense of driver immersion. In the midground, a white Lamborghini convertible supercar races ahead on the left, its low-slung body and prominent rear wing standing out dramatically against the night sky. \textcolor{red}{Floating text at the top displays the white English title 'Smoked a Lambo going home' and the central phrase 'wait for it...' builds suspense and foreshadows an impending overtaking climax.} Environmental lighting comes primarily from distant streetlights and digital billboards, creating high-contrast highlights against the dark sky.\\

\textbf{Ground truth frame:}\vspace{1pt}\\
\begin{center}
\includegraphics[height=2.2cm, keepaspectratio]{case_studies/photos/gt_b.jpg}
\vspace{5pt}
\end{center}

\textbf{Model's prediction:}  \textcolor{PineGreen}{\textbf{True}}\\

\textbf{Human's prediction:}  \textcolor{red}{\textbf{False}}\\

\end{tcolorbox}

%% file: case_studies/case5_verification2.tex
\tcbset{
    breakable,
    colframe=blue!5!black,
    colback=gray!10!white,
    fonttitle=\bfseries,
    width=\columnwidth % 使tcolorbox适应单栏宽度
}

\begin{tcolorbox}[
    title=\textbf{Wrong verification (II)},
    fonttitle=\bfseries
]

\textbf{Input fragment:} \vspace{2pt} \\
Please find the link to this video: I want to find a video segment, the main subject of this frame is a large, majestic adult male lion standing inside a metal transport cage mounted on the back of a truck. It has a thick, flowing mane, transitioning from golden yellow on the head to deep black around the neck, exuding strength and dignity. The lion stands tall, alert and solemn, gazing intently toward the upper right of the frame. The cage, featuring a black metal frame and dense silver wire mesh, is positioned above a lowered metal ramp and a heavy-duty white bumper with circular holes, creating a strong industrial and oppressive feel. Bright natural sunlight streams from the upper right, clearly highlighting the texture of the lion's facial muscles and mane. I vaguely recall the audio of a metal cage being shaken violently, as if a small animal was agitated and restless. From what I remember, there was also a man's voice, casually repeating "Yeah," as if responding to the lively creature, which felt very routine.\\

\textbf{MLLM output context:} \vspace{1pt} \\
\textless think\textgreater The user describes a majestic male lion in a transport cage on a truck with specific lighting. The audio description (violent shaking, small animal, man saying "Yeah") strongly resembles a viral meme involving a raccoon or possum, suggesting the user might be looking for a meme edit or a specific viral clip. I will search for the visual content of the lion transport and the specific audio context.\textless/think\textgreater 

\textless tool\_call\textgreater\{\textquotedbl name\textquotedbl: \textquotedbl search\_videos\textquotedbl, \textquotedbl arguments\textquotedbl: \{\textquotedbl query\textquotedbl: [\textquotedbl \textcolor{PineGreen}{lion in transport cage truck majestic}\textquotedbl, \textquotedbl \textcolor{PineGreen}{lion cage truck release video}\textquotedbl]\}\}\textless/tool\_call\textgreater \\

\textbf{Selected frame:}\vspace{1pt}\\
\includegraphics[width=\columnwidth]
{case_studies/photos/4c.png}

\textbf{Ground truth text:} \vspace{2pt}\\
I want to find a video segment, the main subject of this frame is \textcolor{red}{a large, majestic adult male lion standing inside a metal transport cage mounted on the back of a truck. It has a thick, flowing mane, transitioning from golden yellow on the head to deep black around the neck, exuding strength and dignity.} The lion stands tall, alert and solemn, gazing intently toward the upper right of the frame. The cage, featuring a black metal frame and dense silver wire mesh, is positioned above a lowered metal ramp and a heavy-duty white bumper with circular holes, creating a strong industrial and oppressive feel. Bright natural sunlight streams from the upper right, clearly highlighting the texture of the lion's facial muscles and mane.\\

\textbf{Ground truth frame:}\vspace{1pt}\\
\begin{center}
\includegraphics[height=2.2cm, keepaspectratio]{case_studies/photos/gt_c.jpg}
\vspace{5pt}
\end{center}

\textbf{Model's prediction:}  \textcolor{red}{\textbf{False}}\\

\textbf{Human's prediction:}  \textcolor{PineGreen}{\textbf{True}}\\

\end{tcolorbox}

%% file: case_studies/case6.tex
\tcbset{
    breakable,
    colframe=blue!5!black,
    colback=gray!10!white,
    fonttitle=\bfseries,
    width=\columnwidth % 使tcolorbox适应单栏宽度
}

\begin{tcolorbox}[
    title=\textbf{Case 1: The Game Awards},
    fonttitle=\bfseries
]

For the case, \textbf{nothink} directly enumerates OCR/ASR details observed within the shots, for example:

\begin{itemize}
    \item "The Game Awards Best Esports Game nominees Counter-Strike 2 DOTA 2 League of Legends."
    \item "The Game Awards Best Esports Game female presenter silver gold halter neck gown."
\end{itemize}

This approach directly concatenates localized information such as game names and clothing details into the search queries. However, full video titles rarely include such fine-grained enumerations, making the queries overly specific and reducing retrieval success. In practice, the second set of nothink queries failed to retrieve the target video.

In contrast, \textbf{video imagination} generates queries such as:
\begin{itemize}
    \item "The Game Awards 2023 Best Esports Game."
    \item "The Game Awards 2023 Best Esports Game presenter."
\end{itemize}

Here, the strategy performs shot-to-container abstraction, mapping individual shots to a higher-level semantic description—“the Best Esports Game segment at TGA 2023”—which aligns better with typical video titles and significantly improves retrieval accuracy.

\end{tcolorbox}

%% file: case_studies/case7.tex
\tcbset{
    breakable,
    colframe=blue!5!black,
    colback=gray!10!white,
    fonttitle=\bfseries,
    width=\columnwidth % 使tcolorbox适应单栏宽度
}

\begin{tcolorbox}[
    title=\textbf{Case 2: Alpaca video},
    fonttitle=\bfseries
]

The difference is even more pronounced in the second case. \textbf{Nothink} produces queries like:

\begin{itemize}
    \item "I AM THE RAREST ANIMAL brown alpaca clown emoji hand stroking."
    \item "AURAFORGE alpaca heavy grotesque breaths electronic rhythm."
\end{itemize}

These include highly localized and surface-level visual or auditory details. Although present in the shots, such details are not necessarily reflected in full video titles, leading to retrieval failure or semantic overfitting.

In comparison, \textbf{video imagination} generates queries such as:
\begin{itemize}
    \item "I AM THE RAREST ANIMAL alpaca meme."
    \item "AURAFORGE alpaca video."
\end{itemize}

Here, surface-level details such as “clown emoji hand stroking” are abstracted into higher-level concepts, e.g., “meme.” This abstraction better aligns with the semantic expectations of search engines and facilitates more robust semantic retrieval. Notably, this is not mere information reduction; rather, it uses causal reasoning to map shot-level cues to container-level semantics likely to match full video titles.

\end{tcolorbox}

%% file: custom.bib
@misc{yu2025aligningmultimodalllmhuman,
      title={Aligning Multimodal LLM with Human Preference: A Survey}, 
      author={Tao Yu and Yi-Fan Zhang and Chaoyou Fu and Junkang Wu and Jinda Lu and Kun Wang and Xingyu Lu and Yunhang Shen and Guibin Zhang and Dingjie Song and Yibo Yan and Tianlong Xu and Qingsong Wen and Zhang Zhang and Yan Huang and Liang Wang and Tieniu Tan},
      year={2025},
      eprint={2503.14504},
      archivePrefix={arXiv},
      primaryClass={cs.CV},
      url={https://arxiv.org/abs/2503.14504}, 
}

@misc{zhang2025mmrlhfstepforwardmultimodal,
      title={MM-RLHF: The Next Step Forward in Multimodal LLM Alignment}, 
      author={Yi-Fan Zhang and Tao Yu and Haochen Tian and Chaoyou Fu and Peiyan Li and Jianshu Zeng and Wulin Xie and Yang Shi and Huanyu Zhang and Junkang Wu and Xue Wang and Yibo Hu and Bin Wen and Fan Yang and Zhang Zhang and Tingting Gao and Di Zhang and Liang Wang and Rong Jin and Tieniu Tan},
      year={2025},
      eprint={2502.10391},
      archivePrefix={arXiv},
      primaryClass={cs.CL},
      url={https://arxiv.org/abs/2502.10391}, 
}

@misc{jin2025searchr1trainingllmsreason,
      title={Search-R1: Training LLMs to Reason and Leverage Search Engines with Reinforcement Learning}, 
      author={Bowen Jin and Hansi Zeng and Zhenrui Yue and Jinsung Yoon and Sercan Arik and Dong Wang and Hamed Zamani and Jiawei Han},
      year={2025},
      eprint={2503.09516},
      archivePrefix={arXiv},
      primaryClass={cs.CL},
      url={https://arxiv.org/abs/2503.09516}, 
}

@misc{wu2025webdancer,
      title={WebDancer: Towards Autonomous Information Seeking Agency},
      author={Jialong Wu and Baixuan Li and Runnan Fang and Wenbiao Yin and Liwen Zhang and Zhengwei Tao and Dingchu Zhang and Zekun Xi and Yong Jiang and Pengjun Xie and Fei Huang and Jingren Zhou},
      year={2025},
      eprint={2505.22648},
      archivePrefix={arXiv},
      primaryClass={cs.CL},
      url={https://arxiv.org/abs/2505.22648},
}

@misc{wu2025webwalker,
      title={WebWalker: Benchmarking LLMs in Web Traversal},
      author={Jialong Wu and Wenbiao Yin and Yong Jiang and Zhenglin Wang and Zekun Xi and Runnan Fang and Deyu Zhou and Pengjun Xie and Fei Huang},
      year={2025},
      eprint={2501.07572},
      archivePrefix={arXiv},
      primaryClass={cs.CL},
      url={https://arxiv.org/abs/2501.07572},
}

@misc{tao2025webshaper,
      title={WebShaper: Agentically Data Synthesizing via Information-Seeking Formalization},
      author={Zhengwei Tao and Jialong Wu and Wenbiao Yin and Junkai Zhang and Baixuan Li and Haiyang Shen and Kuan Li and Liwen Zhang and Xinyu Wang and Yong Jiang and Pengjun Xie and Fei Huang and Jingren Zhou},
      year={2025},
      eprint={2507.15061},
      archivePrefix={arXiv},
      primaryClass={cs.CL},
      url={https://arxiv.org/abs/2507.15061},
}

@misc{yu2025browseragentbuildingwebagents,
      title={BrowserAgent: Building Web Agents with Human-Inspired Web Browsing Actions}, 
      author={Tao Yu and Zhengbo Zhang and Zhiheng Lyu and Junhao Gong and Hongzhu Yi and Xinming Wang and Yuxuan Zhou and Jiabing Yang and Ping Nie and Yan Huang and Wenhu Chen},
      year={2025},
      eprint={2510.10666},
      archivePrefix={arXiv},
      primaryClass={cs.CL},
      url={https://arxiv.org/abs/2510.10666}, 
}

@article{geng2025webwatcher,
  title={WebWatcher: Breaking New Frontiers of Vision-Language Deep Research Agent},
  author={Geng, Xinyu and Xia, Peng and Zhang, Zhen and Wang, Xinyu and Wang, Qiuchen and Ding, Ruixue and Wang, Chenxi and Wu, Jialong and Zhao, Yida and Li, Kuan and others},
  journal={arXiv preprint arXiv:2508.05748},
  year={2025}
}

@misc{lu2025skaldlearningbasedshotassembly,
      title={SKALD: Learning-Based Shot Assembly for Coherent Multi-Shot Video Creation}, 
      author={Chen Yi Lu and Md Mehrab Tanjim and Ishita Dasgupta and Somdeb Sarkhel and Gang Wu and Saayan Mitra and Somali Chaterji},
      year={2025},
      eprint={2503.08010},
      archivePrefix={arXiv},
      primaryClass={cs.CV},
      url={https://arxiv.org/abs/2503.08010}, 
}

@inproceedings{10.1145/3591106.3592247,
author = {Yang, Guoxing and Lu, Haoyu and Sun, Zelong and Lu, Zhiwu},
title = {Shot Retrieval and Assembly with Text Script for Video Montage Generation},
year = {2023},
doi = {10.1145/3591106.3592247},
series = {ICMR '23}
}

@misc{pardo2021learningcutwatchingmovies,
      title={Learning to Cut by Watching Movies}, 
      author={Alejandro Pardo and Fabian Caba Heilbron and Juan León Alcázar and Ali Thabet and Bernard Ghanem},
      year={2021},
      eprint={2108.04294},
      archivePrefix={arXiv},
      primaryClass={cs.CV},
      url={https://arxiv.org/abs/2108.04294}, 
}

@article{bonneel2013example,
  title={Example-based video color grading.},
  author={Bonneel, Nicolas and Sunkavalli, Kalyan and Paris, Sylvain and Pfister, Hanspeter},
  journal={ACM Trans. Graph.},
  volume={32},
  number={4},
  pages={39--1},
  year={2013}
}

@misc{tian2018audiovisualeventlocalizationunconstrained,
      title={Audio-Visual Event Localization in Unconstrained Videos}, 
      author={Yapeng Tian and Jing Shi and Bochen Li and Zhiyao Duan and Chenliang Xu},
      year={2018},
      eprint={1803.08842},
      archivePrefix={arXiv},
      primaryClass={cs.CV},
      url={https://arxiv.org/abs/1803.08842}, 
}

@article{Tu_2021,
   title={UGC-VQA: Benchmarking Blind Video Quality Assessment for User Generated Content},
   volume={30},
   ISSN={1941-0042},
   url={http://dx.doi.org/10.1109/TIP.2021.3072221},
   DOI={10.1109/tip.2021.3072221},
   journal={IEEE Transactions on Image Processing},
   publisher={Institute of Electrical and Electronics Engineers (IEEE)},
   author={Tu, Zhengzhong and Wang, Yilin and Birkbeck, Neil and Adsumilli, Balu and Bovik, Alan C.},
   year={2021},
   pages={4449–4464} }

@misc{openai2024gpt4,
  author       = {OpenAI},
  title        = {GPT-4 Technical Report},
  year         = {2024},
  url          = {https://openai.com/gpt-4}
}

@misc{google2023gemini,
  title={Gemini: Google's Multimodal Models},
  author={Google DeepMind},
  year={2023},
  note={https://deepmind.google/technologies/gemini/}
}

@misc{anthropic2024claude3,
  author       = {Anthropic},
  title        = {Claude 3 Models},
  year         = {2024},
  howpublished = {\url{https://www.anthropic.com/index/introducing-claude-3}},
  note         = {Accessed: 2025-09-21}
}

@misc{yang2025qwen3technicalreport,
      title={Qwen3 Technical Report}, 
      author={An Yang and Anfeng Li and Baosong Yang and Beichen Zhang and Binyuan Hui and Bo Zheng and Bowen Yu and Chang Gao and Chengen Huang and Chenxu Lv and Chujie Zheng and Dayiheng Liu and Fan Zhou and Fei Huang and Feng Hu and Hao Ge and Haoran Wei and Huan Lin and Jialong Tang and Jian Yang and Jianhong Tu and Jianwei Zhang and Jianxin Yang and Jiaxi Yang and Jing Zhou and Jingren Zhou and Junyang Lin and Kai Dang and Keqin Bao and Kexin Yang and Le Yu and Lianghao Deng and Mei Li and Mingfeng Xue and Mingze Li and Pei Zhang and Peng Wang and Qin Zhu and Rui Men and Ruize Gao and Shixuan Liu and Shuang Luo and Tianhao Li and Tianyi Tang and Wenbiao Yin and Xingzhang Ren and Xinyu Wang and Xinyu Zhang and Xuancheng Ren and Yang Fan and Yang Su and Yichang Zhang and Yinger Zhang and Yu Wan and Yuqiong Liu and Zekun Wang and Zeyu Cui and Zhenru Zhang and Zhipeng Zhou and Zihan Qiu},
      year={2025},
      eprint={2505.09388},
      archivePrefix={arXiv},
      primaryClass={cs.CL},
      url={https://arxiv.org/abs/2505.09388}, 
}

@misc{lei2021qvhighlightsdetectingmomentshighlights,
      title={QVHighlights: Detecting Moments and Highlights in Videos via Natural Language Queries}, 
      author={Jie Lei and Tamara L. Berg and Mohit Bansal},
      year={2021},
      eprint={2107.09609},
      archivePrefix={arXiv},
      primaryClass={cs.CV},
      url={https://arxiv.org/abs/2107.09609}, 
}

@misc{hou2023coneefficientcoarsetofinealignment,
      title={CONE: An Efficient COarse-to-fiNE Alignment Framework for Long Video Temporal Grounding}, 
      author={Zhijian Hou and Wanjun Zhong and Lei Ji and Difei Gao and Kun Yan and Wing-Kwong Chan and Chong-Wah Ngo and Zheng Shou and Nan Duan},
      year={2023},
      eprint={2209.10918},
      archivePrefix={arXiv},
      primaryClass={cs.CV},
      url={https://arxiv.org/abs/2209.10918}, 
}

@misc{radford2021learningtransferablevisualmodels,
      title={Learning Transferable Visual Models From Natural Language Supervision}, 
      author={Alec Radford and Jong Wook Kim and Chris Hallacy and Aditya Ramesh and Gabriel Goh and Sandhini Agarwal and Girish Sastry and Amanda Askell and Pamela Mishkin and Jack Clark and Gretchen Krueger and Ilya Sutskever},
      year={2021},
      eprint={2103.00020},
      archivePrefix={arXiv},
      primaryClass={cs.CV},
      url={https://arxiv.org/abs/2103.00020}, 
}

@misc{xu2025qwen3omnitechnicalreport,
      title={Qwen3-Omni Technical Report}, 
      author={Jin Xu and Zhifang Guo and Hangrui Hu and Yunfei Chu and Xiong Wang and Jinzheng He and Yuxuan Wang and Xian Shi and Ting He and Xinfa Zhu and Yuanjun Lv and Yongqi Wang and Dake Guo and He Wang and Linhan Ma and Pei Zhang and Xinyu Zhang and Hongkun Hao and Zishan Guo and Baosong Yang and Bin Zhang and Ziyang Ma and Xipin Wei and Shuai Bai and Keqin Chen and Xuejing Liu and Peng Wang and Mingkun Yang and Dayiheng Liu and Xingzhang Ren and Bo Zheng and Rui Men and Fan Zhou and Bowen Yu and Jianxin Yang and Le Yu and Jingren Zhou and Junyang Lin},
      year={2025},
      eprint={2509.17765},
      archivePrefix={arXiv},
      primaryClass={cs.CL},
      url={https://arxiv.org/abs/2509.17765}, 
}

@misc{huang2023vtimellmempowerllmgrasp,
      title={VTimeLLM: Empower LLM to Grasp Video Moments}, 
      author={Bin Huang and Xin Wang and Hong Chen and Zihan Song and Wenwu Zhu},
      year={2023},
      eprint={2311.18445},
      archivePrefix={arXiv},
      primaryClass={cs.CV},
      url={https://arxiv.org/abs/2311.18445}, 
}

@misc{ren2024timechattimesensitivemultimodallarge,
      title={TimeChat: A Time-sensitive Multimodal Large Language Model for Long Video Understanding}, 
      author={Shuhuai Ren and Linli Yao and Shicheng Li and Xu Sun and Lu Hou},
      year={2024},
      eprint={2312.02051},
      archivePrefix={arXiv},
      primaryClass={cs.CV},
      url={https://arxiv.org/abs/2312.02051}, 
}

@misc{wang2025groundedvideollmsharpeningfinegrainedtemporal,
      title={Grounded-VideoLLM: Sharpening Fine-grained Temporal Grounding in Video Large Language Models}, 
      author={Haibo Wang and Zhiyang Xu and Yu Cheng and Shizhe Diao and Yufan Zhou and Yixin Cao and Qifan Wang and Weifeng Ge and Lifu Huang},
      year={2025},
      eprint={2410.03290},
      archivePrefix={arXiv},
      primaryClass={cs.CV},
      url={https://arxiv.org/abs/2410.03290}, 
}

@misc{yu2026seeingbelievingbenchmark,
      title={When Seeing Is Not Believing -- A Benchmark for Search-Grounded Video Misinformation Detection}, 
      author={Tao Yu and Yujia Yang and Shenghua Chai and Zhang Jinshuai and Haopeng Jin and Hao Wang and Minghui Zhang and Zhongtian Luo and Yuchen Long and Xinlong Chen and Jiabing Yang and Zhaolu Kang and Yuxuan Zhou and Zhengyu Man and Xinming Wang and Hongzhu Yi and Zheqi He and Xi Yang and Yan Huang and Liang Wang},
      year={2026},
      eprint={2606.04098},
      archivePrefix={arXiv},
      primaryClass={cs.CV},
      url={https://arxiv.org/abs/2606.04098}, 
}

@misc{yu2026allinoneagentbenchmarkingrolespecialized,
      title={Beyond the All-in-One Agent: Benchmarking Role-Specialized Multi-Agent Collaboration in Enterprise Workflows}, 
      author={Tao Yu and Hao Wang and Changyu Li and Shenghua Chai and Minghui Zhang and Zhongtian Luo and Yuxuan Zhou and Haopeng Jin and Zhaolu Kang and Jiabing Yang and YiFan Zhang and Xinming Wang and Hongzhu Yi and Zheqi He and Jing-Shu Zheng and Xi Yang and Yan Huang and Liang Wang},
      year={2026},
      eprint={2605.08761},
      archivePrefix={arXiv},
      primaryClass={cs.MA},
      url={https://arxiv.org/abs/2605.08761}, 
}

@misc{yu2026paperxunifiedframeworkmultimodal,
      title={PaperX: A Unified Framework for Multimodal Academic Presentation Generation with Scholar DAG}, 
      author={Tao Yu and Minghui Zhang and Zhiqing Cui and Hao Wang and Zhongtian Luo and Shenghua Chai and Junhao Gong and Yuzhao Peng and Yuxuan Zhou and Yujia Yang and Zhenghao Zhang and Haopeng Jin and Xinming Wang and Yufei Xiong and Jiabing Yang and Jiahao Yuan and Hanqing Wang and Hongzhu Yi and Yan Huang and Liang Wang},
      year={2026},
      eprint={2602.03866},
      archivePrefix={arXiv},
      primaryClass={cs.DL},
      url={https://arxiv.org/abs/2602.03866}, 
}

@misc{yu2026closedpoolvideoretrievalbenchmark,
      title={Beyond Closed-Pool Video Retrieval: A Benchmark and Agent Framework for Real-World Video Search and Moment Localization}, 
      author={Tao Yu and Yujia Yang and Haopeng Jin and Junhao Gong and Xinlong Chen and Yuxuan Zhou and Shanbin Zhang and Jiabing Yang and Xinming Wang and Hongzhu Yi and Ping Nie and Kai Zou and Zhang Zhang and Yan Huang and Liang Wang and Yeshani and Ruiwen Tao and Jin Ma and Haijin Liang and Jinwen Luo},
      year={2026},
      eprint={2602.10159},
      archivePrefix={arXiv},
      primaryClass={cs.CV},
      url={https://arxiv.org/abs/2602.10159}, 
}

@misc{yu2026omnideepsearchbenchmarkaudiodrivenomnimodal,
      title={Omni-DeepSearch: A Benchmark for Audio-Driven Omni-Modal Deep Search}, 
      author={Tao Yu and yiming ding and Shenghua Chai and Minghui Zhang and Zhongtian Luo and Xinming Wang and Xinlong Chen and Zhaolu Kang and Junhao Gong and Yuxuan Zhou and Haopeng Jin and Zhiqing Cui and Jiabing Yang and YiFan Zhang and Hongzhu Yi and Zheqi He and Xi Yang and Yan Huang and Liang Wang},
      year={2026},
      eprint={2605.08762},
      archivePrefix={arXiv},
      primaryClass={cs.SD},
      url={https://arxiv.org/abs/2605.08762}, 
}

@misc{yu2026scoperoutercostawareopensetvlm,
      title={SCOPE-Router: Cost-Aware Open-Set VLM Routing for Execution-Oriented Tasks}, 
      author={Tao Yu and Yifei Qu and Zhiqing Cui and Pengfei Zhou and Zhongtian Luo and Yujia Yang and Shenghua Chai and Haopeng Jin and Zhenghao Zhang and Xinming Wang and Hongzhu Yi and Wangbo Zhao and Zhenglin Wan and Yan Huang and Yeshani and Jinwen Luo and Yang You},
      year={2026},
      eprint={2608.12127},
      archivePrefix={arXiv},
      primaryClass={cs.CV},
      url={https://arxiv.org/abs/2608.12127}, 
}

@misc{zhou2017automaticlearningproceduresweb,
      title={Towards Automatic Learning of Procedures from Web Instructional Videos}, 
      author={Luowei Zhou and Chenliang Xu and Jason J. Corso},
      year={2017},
      eprint={1703.09788},
      archivePrefix={arXiv},
      primaryClass={cs.CV},
      url={https://arxiv.org/abs/1703.09788}, 
}

@inproceedings{Giancola_2018,
   title={SoccerNet: A Scalable Dataset for Action Spotting in Soccer Videos},
   url={http://dx.doi.org/10.1109/CVPRW.2018.00223},
   DOI={10.1109/cvprw.2018.00223},
   booktitle={2018 IEEE/CVF Conference on Computer Vision and Pattern Recognition Workshops (CVPRW)},
   publisher={IEEE},
   author={Giancola, Silvio and Amine, Mohieddine and Dghaily, Tarek and Ghanem, Bernard},
   year={2018},
   month=jun, pages={1792–179210} }

@article{coursaris2008empirical,
  title={An empirical investigation of color temperature and gender effects on web aesthetics},
  author={Coursaris, Constantinos K and Swierenga, Sarah J and Watrall, Ethan},
  journal={Journal of usability studies},
  volume={3},
  number={3},
  pages={103--117},
  year={2008},
  publisher={Usability Professionals' Association Bloomingdale, IL}
}

@misc{richter2016playingdatagroundtruth,
      title={Playing for Data: Ground Truth from Computer Games}, 
      author={Stephan R. Richter and Vibhav Vineet and Stefan Roth and Vladlen Koltun},
      year={2016},
      eprint={1608.02192},
      archivePrefix={arXiv},
      primaryClass={cs.CV},
      url={https://arxiv.org/abs/1608.02192}, 
}

@inproceedings{gemmeke2017audio,
  title={Audio set: An ontology and human-labeled dataset for audio events},
  author={Gemmeke, Jort F and Ellis, Daniel PW and Freedman, Dylan and Jansen, Aren and Lawrence, Wade and Moore, R Channing and Plakal, Manoj and Ritter, Marvin},
  booktitle={2017 IEEE international conference on acoustics, speech and signal processing (ICASSP)},
  pages={776--780},
  year={2017},
  organization={IEEE}
}

@article{Hosu_2020,
   title={KonIQ-10k: An Ecologically Valid Database for Deep Learning of Blind Image Quality Assessment},
   volume={29},
   ISSN={1941-0042},
   url={http://dx.doi.org/10.1109/TIP.2020.2967829},
   DOI={10.1109/tip.2020.2967829},
   journal={IEEE Transactions on Image Processing},
   publisher={Institute of Electrical and Electronics Engineers (IEEE)},
   author={Hosu, Vlad and Lin, Hanhe and Sziranyi, Tamas and Saupe, Dietmar},
   year={2020},
   pages={4041–4056} }

@misc{soucek2020transnetv2effectivedeep,
      title={TransNet V2: An effective deep network architecture for fast shot transition detection}, 
      author={Tomáš Souček and Jakub Lokoč},
      year={2020},
      eprint={2008.04838},
      archivePrefix={arXiv},
      primaryClass={cs.CV},
      url={https://arxiv.org/abs/2008.04838}, 
}

@misc{wang2025internvl35advancingopensourcemultimodal,
      title={InternVL3.5: Advancing Open-Source Multimodal Models in Versatility, Reasoning, and Efficiency}, 
      author={Weiyun Wang and Zhangwei Gao and Lixin Gu and Hengjun Pu and Long Cui and Xingguang Wei and Zhaoyang Liu and Linglin Jing and Shenglong Ye and Jie Shao and Zhaokai Wang and Zhe Chen and Hongjie Zhang and Ganlin Yang and Haomin Wang and Qi Wei and Jinhui Yin and Wenhao Li and Erfei Cui and Guanzhou Chen and Zichen Ding and Changyao Tian and Zhenyu Wu and Jingjing Xie and Zehao Li and Bowen Yang and Yuchen Duan and Xuehui Wang and Zhi Hou and Haoran Hao and Tianyi Zhang and Songze Li and Xiangyu Zhao and Haodong Duan and Nianchen Deng and Bin Fu and Yinan He and Yi Wang and Conghui He and Botian Shi and Junjun He and Yingtong Xiong and Han Lv and Lijun Wu and Wenqi Shao and Kaipeng Zhang and Huipeng Deng and Biqing Qi and Jiaye Ge and Qipeng Guo and Wenwei Zhang and Songyang Zhang and Maosong Cao and Junyao Lin and Kexian Tang and Jianfei Gao and Haian Huang and Yuzhe Gu and Chengqi Lyu and Huanze Tang and Rui Wang and Haijun Lv and Wanli Ouyang and Limin Wang and Min Dou and Xizhou Zhu and Tong Lu and Dahua Lin and Jifeng Dai and Weijie Su and Bowen Zhou and Kai Chen and Yu Qiao and Wenhai Wang and Gen Luo},
      year={2025},
      eprint={2508.18265},
      archivePrefix={arXiv},
      primaryClass={cs.CV},
      url={https://arxiv.org/abs/2508.18265}, 
}

@misc{zhang2025videollama3frontiermultimodal,
      title={VideoLLaMA 3: Frontier Multimodal Foundation Models for Image and Video Understanding}, 
      author={Boqiang Zhang and Kehan Li and Zesen Cheng and Zhiqiang Hu and Yuqian Yuan and Guanzheng Chen and Sicong Leng and Yuming Jiang and Hang Zhang and Xin Li and Peng Jin and Wenqi Zhang and Fan Wang and Lidong Bing and Deli Zhao},
      year={2025},
      eprint={2501.13106},
      archivePrefix={arXiv},
      primaryClass={cs.CV},
      url={https://arxiv.org/abs/2501.13106}, 
}

@article{zhou2025does,
  title={Why does weak-OOD help? A Further Step Towards Understanding Jailbreaking VLMs},
  author={Zhou, Yuxuan and Peng, Yuzhao and Bai, Yang and Gao, Kuofeng and Zhang, Yihao and Zhang, Yechao and Chen, Xun and Yu, Tao and Dai, Tao and Xia, Shu-Tao},
  journal={arXiv preprint arXiv:2511.08367},
  year={2025}
}

@article{zhou2025jpro,
  title={Jpro: Automated multimodal jailbreaking via multi-agent collaboration framework},
  author={Zhou, Yuxuan and Bai, Yang and Gao, Kuofeng and Dai, Tao and Xia, Shu-Tao},
  journal={arXiv preprint arXiv:2511.07315},
  year={2025}
}

@misc{chou2026improving,
      title={Improving Deepfake Detection with Reinforcement Learning-Based Adaptive Data Augmentation}, 
      author={Yuxuan Zhou and Tao Yu and Wen Huang and Yuheng Zhang and Tao Dai and Shu-Tao Xia},
      year={2025},
      eprint={2511.07051},
      archivePrefix={arXiv},
      primaryClass={cs.CV},
      url={https://arxiv.org/abs/2511.07051}, 
}

@article{wang2025hitchhiker,
  title={The hitchhiker’s guide to autonomous research: A survey of scientific agents},
  author={Wang, Xinming and Xu, Jian and Feng, Aslan H and Chen, Yi and Guo, Haiyang and Zhu, Fei and Shao, Yuanqi and Ren, Minsi and Yi, Hongzhu and Lian, Sheng and others},
  journal={TechRxiv.August 07, 2025. DOI:10.36227/techrxiv175459840.02185500/V1},
  year={2025},
  publisher={Authorea}
}

@article{wang2026beyond,
  title={Beyond Correctness: Benchmarking and Aligning Response Behaviors in Hybrid-Thinking MLLMs},
  author={Wang, Xinming and Wang, Weinong and Yang, Hongming and Lin, Yansong and Ruan, Zheng and Peng, Shangpin and Peng, Qiming and Qiao, Nan and Lu, Fengyuan and Ma, Guoqing and others},
  journal={arXiv preprint arXiv:2608.12781},
  year={2026}
}
